\documentclass[10pt,journal,compsoc]{IEEEtran}
\ifCLASSOPTIONcompsoc
  \usepackage[nocompress]{cite}
\else
  \usepackage{cite}
\fi
\ifCLASSINFOpdf
\else
\fi
\usepackage{booktabs}
\usepackage{array}
\usepackage{makecell}
\usepackage{multirow}
\usepackage{pdflscape}
\usepackage{varwidth}
\usepackage{graphicx}
\usepackage{color}
\usepackage{subfigure}
\usepackage[graphicx]{realboxes}
\usepackage{diagbox}
\usepackage{amsfonts,amssymb} 
 \usepackage{cite}
\usepackage[numbers,sort&compress]{natbib}
\usepackage{amsmath}
\usepackage{tablefootnote}
\usepackage[ruled]{algorithm2e}

\usepackage{pifont}
\usepackage{booktabs}
\usepackage{threeparttable}
\makeatletter
\renewcommand{\maketag@@@}[1]{\hbox{\m@th\normalsize\normalfont#1}}%
\makeatother
\usepackage{graphicx}

\makeatletter
\DeclareRobustCommand{\cev}[1]{%
	{\mathpalette\do@cev{#1}}%
}
\newcommand{\do@cev}[2]{%
	\vbox{\offinterlineskip
		\sbox\z@{$\m@th#1 x$}%
		\ialign{##\cr
			\hidewidth\reflectbox{$\m@th#1\vec{}\mkern4mu$}\hidewidth\cr
			\noalign{\kern-\ht\z@}
			$\m@th#1#2$\cr
		}%
	}%
}
\makeatother
\begin{document}
%\title{Time Series Forecasting via Semi-Asymmetric Convolutional Architecture with Global Atrous Sliding Window}
\title{Time Series Forecasting via Semi-Asymmetric Convolutional Architecture with Global Atrous Sliding Window}

\author{Yuanpeng He
        % <-this % stops a space
%\IEEEcompsocitemizethanks{\IEEEcompsocthanksitem Yuanpeng He and Wenpin Jiao are with Key Laboratory of High Confidence Software Technologies, Peking
%	University, Peking, 100871, China; School of Computer Science, Peking University, Peking, 100871, China.\\
%	E-mail: heyuanpengpku@gmail.com, hyppyh010403@email.swu.edu.cn and jwp@pku.edu.cn
%	\IEEEcompsocthanksitem Yuntong Hu is with Hanhong College, Southwest University, Chongqing, 400715, China
%	\IEEEcompsocthanksitem Youjiang Shao is with College of Computer Science, Sichuan University, Chengdu, 610065, China
%	\IEEEcompsocthanksitem Haoyun Dong is with School of Information Science and Engineering, East China University of Science and Technology, Shanghai, 200237, China}% <-this % stops an unwanted space
\IEEEcompsocitemizethanks{\IEEEcompsocthanksitem Yuanpeng He is with Key Laboratory of High Confidence Software Technologies, Peking
	University, Peking, 100871, China; School of Computer Science, Peking University, Peking, 100871, China.\\
	E-mail: heyuanpengpku@gmail.com
}% <-this % stops an unwanted space
\thanks{Manuscript received; revised}}


\markboth{Journal of \LaTeX\ Class Files,~Vol.~14, No.~8, August~2015}%
{Shell \MakeLowercase{\textit{et al.}}: Bare Demo of IEEEtran.cls for Computer Society Journals}



\IEEEtitleabstractindextext{%
\begin{abstract}
The proposed method in this paper is designed to address the problem of time series forecasting. Although some exquisitely designed models achieve excellent prediction performances, how to extract more useful information and make accurate predictions is still an open issue. Most of modern models only focus on a short range of information, which are fatal for problems such as time series forecasting which needs to capture long-term information characteristics. As a result, the main concern of this work is to further mine relationship between local and global information contained in time series to produce more precise predictions. In this paper, to satisfactorily realize the purpose, we make three main contributions that are experimentally verified to have performance advantages. Firstly, original time series is transformed into difference sequence which serves as input to the proposed model. And secondly, we introduce the global atrous sliding window into the forecasting model which references the concept of fuzzy time series to associate relevant global information with temporal data within a time period and utilizes central-bidirectional atrous algorithm to capture underlying-related features to ensure validity and consistency of captured data. Thirdly, a variation of widely-used asymmetric convolution which is called semi-asymmetric convolution is devised to more flexibly extract relationships in adjacent elements and corresponding associated global features with adjustable ranges of convolution on vertical and horizontal directions. The proposed model in this paper achieves state-of-the-art on most of time series datasets provided compared with competitive modern models.
\end{abstract}

}

\maketitle

\IEEEdisplaynontitleabstractindextext

% \fi
\IEEEpeerreviewmaketitle

\IEEEraisesectionheading{\section{Introduction}\label{sec:introduction}}

\IEEEPARstart{T}{ime} series is a sequence taken at successive equally spaced points in time which is also known as dynamic series. Precise prediction of time series has close connections to human society, for instance, it may help people format schedules and company make adjustments on investment strategy. Moreover, foreseeing future behaviour based on analysis of known historical data is of great importance in lots of fields such as epidemic \cite{DBLP:journals/tim/SharmaKMR21}, medical treatment \cite{DBLP:journals/tnn/TanYMYYWY21}, finance \cite{DBLP:journals/pr/ChengYXL22,DBLP:journals/tfs/LiuXLC20} and industrial Internet \cite{DBLP:journals/tii/UchitelevaPLHS22}. Time series forecasting therefore attracts attention from researchers around the world. Nevertheless, how to fully utilize observation to generate accurate and reasonable predictions is still an unsolved problem.

To realize accurate prediction of future, researchers develop various kinds of solutions.  RNN model has been favored by researchers since it was proposed. Because of its recurrent architecture design, RNN models can effectively model long-term dependencies \cite{DBLP:journals/tits/HeLZT22}, therefore achieve an effective understanding of temporal data \cite{DBLP:journals/eswa/LiuGYC20,DBLP:conf/nips/DennisAMSSSJ19} as well. However, RNN may encounter memory overflow due to continuous storage of previous states and gradient vanishing problems. To further make up for shortcomings of RNN, an improved solution based on it is proposed which is called LSTM \cite{DBLP:journals/tim/GuoWLP23}. Its core concepts are the memory cell states that allow information to be passed on backwards, and the gate structures that allow certain information to be added and removed. Coincidentally, researchers also find that temporal task can also benefit form LSTM's characteristics \cite{DBLP:journals/tfs/TangYPYWL22,DBLP:journals/tnn/BandaraBH21}. In a quite long period of time, the model based on RNN has played an important role in development of time series forecasting. Recently, transformer-based models \cite{zhou2022fedformer,DBLP:conf/iclr/LiuYLLLLD22,zhou2021informer,DBLP:conf/iccv/ChenPFL21} have been proposed enormously, which applied self-attention mechanism to distill useful semantic information in time series. However, there exists a doubt that transformer-like structure is not suitable for the task of time series forecasting. Under certain circumstances, the performance of the models can not even match ingeniously designed linear model \cite{DBLP:journals/corr/abs-2205-13504}, which has shaken the position of transformer-based models in time series forecasting. At present, the controversy still continues. Besides, there also exist lots of meaningful works trying to satisfy demand of time series forecasting from other multiple aspects as well \cite{DBLP:journals/pami/GuenT23}.

Moreover, CNN-based models are also widely utilized for prediction of temporal data. They are mainly divided into two categories, one is the variation of causal and dilated convolution \cite{DBLP:journals/tkdd/LiLCZZL22}, the other is algorithms using graph convolutional neural network \cite{DBLP:conf/iclr/ChenSCG22} to solve corresponding problems. Generally, it can be concluded that transformer and CNN models, the two well-established solutions in the field of computer vision, also achieve excellent performance in tasks of time series forecasting. Back to CNN-based models, there have been many new CNN models in recent years, for instance, temporal convolutional network (TCN) \cite{DBLP:journals/corr/abs-1803-01271}, convolutionally low-rank model \cite{DBLP:journals/tit/Liu22} and non-pooling CNN \cite{DBLP:journals/tnn/LiuJW20}. Among them, TCN attracts the most attention which is capable of large-scale parallel processing and managing a series of sequences of arbitrary length and uniformly output sequences with the same length. Specifically, the casual and dilated convolution introduced by it enable CNN forecasting model to possess a larger receptive field to better acquire information in a longer range under strict time restrictions. Moreover, other effective models \cite{DBLP:conf/ijcnn/LiWLLT22,DBLP:journals/tetci/FergusCMRLP21,DBLP:conf/acml/WangLHZ19} also improve performance by enlarging ranges of data selection and more ingenious and flexible extraction of relationships of adjacent and non-adjacent elements because of similar considerations about demand of time series forecasting mentioned above. Nevertheless, it is noting that all of the models still receive data in a relatively restricted way without considering global data features. 

To address the issue, we design a kind of data reconstruction method referencing solution based on partitioned universe of discourse \cite{DBLP:journals/fss/Chen96a} partly which transforms original temporal data into difference sequence \cite{DBLP:journals/tcyb/MaCTN22,DBLP:journals/tii/ZhangTXYCG22,DBLP:journals/complexity/GaoFZYSNH21} ensuring that the model is more likely dealing with steady-state sequences and associates relative positional information of data captured in the view of whole observation time series to reduce difficulty of model learning to some extent. More than that, we choose to replace all of elements by the last one only keeping their positional information as subsidiaries to maximize timeliness of data without losing too much semantic information of temporal data. Besides, the relationship among converted information in different subsections and time series are probably separate \cite{DBLP:journals/tai/WangZZWSLK22}, so there is a need to devise a convolution strategy with different directions and shapes to further mine underlying information. Due to particularity of time series, traditional squared convolution is not capable to manage complex extraction of relationship of elements in temporal data, a variation of asymmetric convolution \cite{DBLP:journals/corr/JinDC14,DBLP:conf/mmasia/LoHCL19} which is called semi-asymmetric convolution is designed accordingly. The semi-asymmetric convolution is divided into horizontal and vertical filters, and they probably possess different length to retrieve interaction information at a more fine-grained level in selected fragments from difference series. The advantage of this improvement is that it is able to effectively obtain temporal features using adjustable scales \cite{DBLP:conf/kdd/YeL0SLF022,DBLP:journals/tvcg/MeschenmoserBSW21}, and speeding up the training and inference process of the proposed model \cite{DBLP:conf/bmvc/JaderbergVZ14} at the same time. In general, the major contributions of this work are summarized as follows:
\begin{enumerate}
	\item The input to proposed model is difference sequences transformed by original observation time series to enable model to learn more easily
	\item A new kind of method of data reconstruction is designed to endow each elements with their corresponding relative positional information
	\item A novel convolutional architecture called semi-asymmetric convolution with flexible scales is designed to acquire information at different levels.
\end{enumerate}

The rest of this paper is organized as follows. In the second section, some related concepts of the proposed model are introduced. And the details of the proposed model are presented in the third section. Besides, the fifth section provides experimental results and corresponding discussions with respect to models. In the last section, conclusions and outlook of future work are given.

\section{Preliminary}
In this section, related concepts about the proposed model are briefly introduced.
\subsection{Difference of First Order}
A first order difference is the difference between two consecutive adjacent terms in a discrete function. Assume there exists a function $y=f(x)$, $y$ is defined only on the non-negative integer value of $x$ and when the independent variable $x$ is iterated through the non-negative integers in turn, namely $x = 0,1,2,...$, the corresponding values of function can be defined as:
\begin{equation}
	f(0), f(1), f(2),...
\end{equation}
it can be abbreviated as:
\begin{equation}
	y_{0},y_{1},y_{2},...
\end{equation}
when the independent value changes from $x$ to $x+1$, the variation of $y=f(x)$ can be defined as:
\begin{equation}
	\Delta y_{x} = f(x+1) - f(x), (x=1,2,3,...)
\end{equation}
it's called the first difference of the function $y(x)$ at point $x$ which is usually denoted as:
\begin{equation}
	\Delta y_{x} = y_{x+1} - y_{x}, (x=1,2,3,...)
\end{equation}

\subsection{Asymmetric Convolution Architecture}
CNN has embraced a quick development recently, it is widely applied in different fields, such as time series and computer vision \cite{DBLP:journals/tkde/LiuZCWWWZ22, DBLP:conf/ijcai/Oh22,DBLP:conf/cvpr/Ding0MHD021} due to its stable and excellent performance. For an operation of convolving, assume an input $\varsigma \in \mathbb{R}^{H \times W}$ and filter  $\mathbb{C}$, the process of generating output $\lambda \in \mathbb{R}^{H' \times W'}$ can be defined as:
\begin{equation}
		  \lambda = \mathbb{C} * \varsigma , \quad \varsigma \in \mathbb{R}^{H \times W}, \lambda \in \mathbb{R}^{H' \times W'},\mathbb{C} \in \mathbb{R}^{d\times d}
\end{equation}
where $*$ is the 2D convolution operator. Moreover, asymmetric convolution \cite{articleDenton, DBLP:conf/bmvc/JaderbergVZ14} is considered as an economical choice to approximate an existing square-kernel convolutional layer for obtaining acceleration and compression. Specifically, the original filter can be decomposed into horizontal and vertical filters, $\mathbb{C}_h,\mathbb{C}_v$, respectively, which can be defined as:
\begin{equation}
	 \mathbb{C} * \varsigma = \mathbb{C}_v * (\mathbb{C}_h * \varsigma ), \mathbb{C}_v \in \mathbb{R}^{d \times 1}, \mathbb{C}_h \in \mathbb{R}^{1 \times d}
\end{equation}
compared with the original convolution utilizing $d\times d$ kernel size, the time complexity changes from $\mathcal{O}((d^{2}H'W')$ to $\mathcal{O}(2dH'W')$. Due to efficiency of the asymmetric architecture, it is widely applied in convolutional neural network design \cite{DBLP:conf/iccv/DingGDH19,DBLP:journals/corr/PaszkeCKC16} and gains performance improvement generally.

\subsection{Atrous Algorithm}
The atrous algorithm is proposed in \cite{DBLP:journals/corr/ChenPKMY14,DBLP:journals/pami/ChenPKMY18} which is also known as dilated convolution. Assume there exists a one-dimensional input $\alpha[s]$, the corresponding output $\beta[s]$ of dilated convolution via a filter $\omega[e]$ with length $E$ can be defined as:
\begin{equation}
	\beta[s] = \sum_{e=1}^{E}\alpha[s+r \cdot e]\omega[e]
\end{equation}
where rate parameter $r$ is corresponding to the stride and standard convolution is a special case for $r=1$. Generally, atrous algorithm is designed to avoid precision loss brought by reduction of feature map on account of multiple convolutional and pooling layers on vision tasks and is broadly utilized in many other import fields, such as audio processing \cite{DBLP:conf/ssw/OordDZSVGKSK16} and time series forecasting \cite{DBLP:journals/corr/abs-1803-01271}. 

\subsection{Naive Forecasting}
The naive forecasting is the simplest prediction method in the field of time series which regards the most recent observation value as the prediction of future. Assume there exists a time series $\mathbb{T}$ with a length $n$ which can be defined as:
\begin{equation}
	\mathbb{T} = \{(t_1,\zeta_1),(t_2,\zeta_2),...,(t_{n-1},\zeta_{n-1}),(t_n,\zeta_n)\}
\end{equation}
where $\zeta_i, i \in [1,n]$ represents observation value at time point $i$. For example, if there is a need to predict $\zeta_{n+q}$ which is unknown, the value of $\zeta_{n}$ can be referenced as the prediction value of $\zeta_{n+q}$ directly. Assume the prediction value of $\zeta_{n+q}$ is $\hat{\zeta}_{n+q}$, then the method of forecasting can be defined as:
\begin{equation}
	\hat{\zeta}_{n+q}= \zeta_{n}
\end{equation}
under various circumstances, naive forecasting is an effective solution to tell the future trend of time series like stock price prediction. Nevertheless, the method introduced is not a satisfying solution in time series forecasting, but it can provide a benchmark for other prediction methods.

\subsection{Fuzzy Time Series}
One of concepts of fuzzy time series is introduced by Chen \cite{DBLP:journals/fss/Chen96a} which is developed based on theories proposed in \cite{articlesong,DBLP:journals/iandc/Zadeh65,articlesong2}. The method of fuzzy time series could extract information effectively by utilizing overall characteristics of time series data and provide stable performance. Specifically, given a time series $\mathbb{T}$, and let $\eta_{min}$ and $\eta_{max}$ be minimum and maximum value in $\mathbb{T}$, the four steps to generate fuzzy time series can be presented as:

Step 1: Select two proper positive numbers $\eta_1$ and $\eta_2$,  a universe of discourse $\mathbb{U}$ can be defined as $[\eta_{min}-\eta_1, \eta_{max}-\eta_2]$

Step 2: Partition $\mathbb{U}$ into segments with equal length $\{u_1,u_2,...u_m\}$ which is called fuzzy intervals

Step 3: Let $\mathbb{Z}_1, \mathbb{Z}_2,..., \mathbb{Z}_k$ be fuzzy sets and they are defined on universe of discourse $\mathbb{U}$ as:
%\begin{equation}
%	\begin{aligned}
%		\mathbb{Z}_1 = z_{11}/u_1 + z_{12}/u_2+...+z_{1m}/u_{m},\\
%		\mathbb{Z}_2 = z_{21}/u_1 + z_{22}/u_2+...+z_{2m}/u_{m},\\
%		\vdots \qquad \qquad \qquad \quad \quad\\
%		\mathbb{Z}_k = z_{k1}/u_1 + z_{k2}/u_2+...+z_{km}/u_{m},
%	\end{aligned}
%\end{equation}

\begin{equation}
	\left\{
	\begin{array}{l}
		\mathbb{Z}_1 = z_{11}/u_1 + z_{12}/u_2+...+z_{1m}/u_{m},\\
		\mathbb{Z}_2 = z_{21}/u_1 + z_{22}/u_2+...+z_{2m}/u_{m},\\
		\vdots \qquad \qquad \qquad \quad \quad\\
		\mathbb{Z}_k = z_{k1}/u_1 + z_{k2}/u_2+...+z_{km}/u_{m}
	\end{array}
	\right.
\end{equation}
where $z_{ij} \in [0,1], i \in [1,k], j \in [1,m]$ and the value of $z_{ij}$ represents the degree of membership of $u_j$ in fuzzy set $\mathbb{Z}_i$.

Step 4: The derived fuzzy logical relationships which possess identical initial states are divided into the same group. Then, the matches between actual values in time series and groups of fuzzy logical relationships can be acquired. 

After these four steps, the original data is transformed into fuzzy time series.

\section{Architecture of the Proposed Forecasting Model}
In this section, the proposed forecasting model based on relevant concepts mentioned above is introduced.
\begin{figure*}
	\centering %表示居中
	\includegraphics[width = 8.8cm,angle=90]{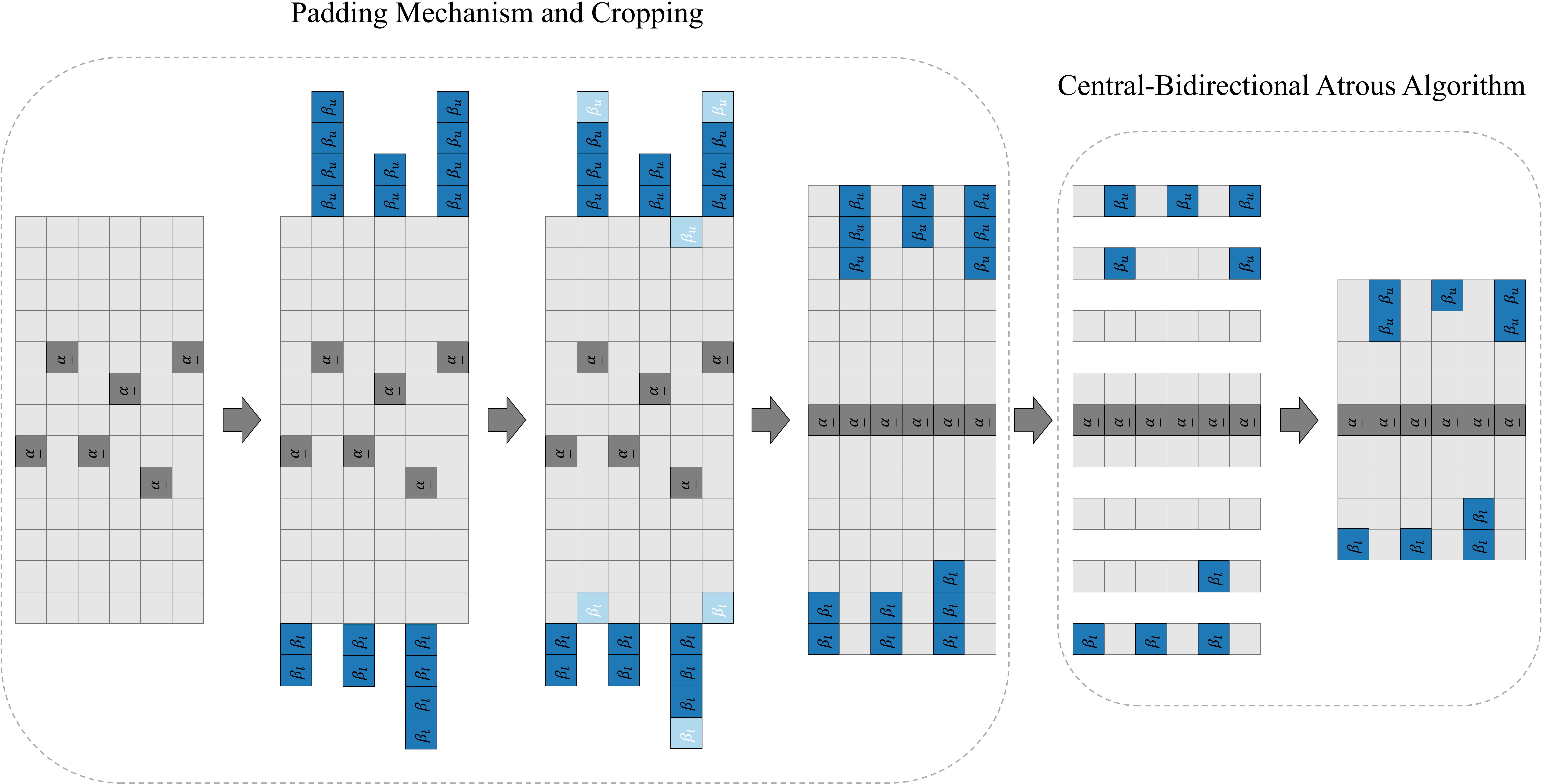}
	\caption{Padding Mechanism, Cropping and Central-Bidirectional Atrous Algorithm with Dilation Factor $d=1$}
	\label{3}
\end{figure*}
\subsection{Difference Layer: Convert Time Series into First Order Difference Sequence}
First, a time series $\mathbb{T}$ is transformed into its first order difference sequence $\mathbb{T}_{\Delta}$. The process can be given as:
\begin{equation}
	\left\{
	\begin{array}{l}
		\mathbb{T} = \{(t_1,\zeta_1),(t_2,\zeta_2),...,(t_{n-1},\zeta_{n-1}),(t_n,\zeta_n)\}\\
		\qquad \qquad \qquad\qquad \qquad\Downarrow  \\
		\mathbb{T}_{\Delta}^{-} = \{(t_{2,1},\zeta_2-\zeta_1),...,(t_{n,n-1},\zeta_n-\zeta_{n-1}) \}\  \\
		\qquad \qquad \quad\Downarrow  \\
		\mathbb{T}_{\Delta} = \{\alpha_1,\alpha_2,...,\alpha_{h}\}\\
	\end{array}
    \right.
\end{equation}
where $h = n-1$ and $\alpha_i, i \in [1,h]$ only contains observed value without timestamp. Obviously, it can be obtained that the length of $\mathbb{T}_{\Delta}$ is $n-1$. In the next step, the input is $\mathbb{T}_{\Delta}$ instead of $\mathbb{T}$.
\subsection{Division layer: Divide Converted Time Series into Sub-Series Based on Sliding Window}
Second, series $\mathbb{T}_{\Delta}$ is divided by sliding window whose size is $\mathbb{W}$ into sub-series, the segmented data fragments are:
\begin{equation}
	\mathbb{T}_{\Delta}^{Seg} = \{ \varUpsilon_1,\varUpsilon_2,...,\varUpsilon_c\}, c = h - \mathbb{W} + 1
\end{equation}
where $\varUpsilon_{j} = \{\alpha_{j}, \alpha_{j+1},...,\alpha_{j+\mathbb{W}-1}\}, j \in [1,c]$.
\subsection{Encoder of segmented sequences}
\subsubsection{Relative Positional Encoding: Reconstruct Sub-Series with View on Global Observation Temporal Data}
Third, in the concept of fuzzy time series proposed in \cite{DBLP:journals/fss/Chen96a}, the two numbers $\eta_1$ and $\eta_2$ are selected intuitively, which may lead to non-reproducibility of experiment results on various datasets. As a result, $\eta_1$ and $\eta_2$ are uniformly set as standard deviation of corresponding first order difference sequence, which can be given as:
\begin{equation}
	\varphi = \eta_1 = \eta_2 = \sigma(\mathbb{T}_{\Delta})
\end{equation}
where $\sigma$ represents standard deviation. Then, universe of discourse $\mathbb{U}$ of $\mathbb{T}_{\Delta}$ can be calculated as:
\begin{equation}
	\mathbb{U}_{\mathbb{T}_{\Delta}} = [\alpha_{min}-\varphi, \alpha_{max}+\varphi] = [\beta_{l},\beta_{u}]
\end{equation}
where $\alpha_{min}$ and $\alpha_{max}$ represent minimum and maximum element contained in $\mathbb{T}_{\Delta}$ and $\beta_{l}$ and $\beta_{u}$ denote lower and upper bound of $\mathbb{U}_{\mathbb{T}_{\Delta}}$. And the number of intervals, $\mathbb{N}$, can be confirmed as:
\begin{equation}
	\mathbb{N} = log_{2}^{h} - 1
\end{equation}
the partitioned universe of discourse can be given as:
\begin{equation}
	\mathbb{U}_{\mathbb{T}_{\Delta}}  = [\beta_{l}, \beta_{l}+\xi,...,\beta_{l}+\kappa\times\xi,...,\beta_{u}-\xi,\beta_{u}]
\end{equation}
where $\xi = (\beta_{u} - \beta_{l}) / \mathbb{N}$ and $\kappa \in [1,\mathbb{N}]$. Then, integrate each element contained in $\varUpsilon_{j}$ into the partitioned universe of discourse $\mathbb{U}_{\mathbb{T}_{\Delta}}$ to create new sequences based on data fragments captured by sliding window:
\begin{equation}\tiny
	\varUpsilon_{j}^{\mathbb{U}_{\mathbb{T}_{\Delta}}}  = \begin{bmatrix}
		\beta_{l}& ... & \beta_{l} +\kappa' \times \xi& \alpha_{j}& \beta_{l} +(\kappa'+1) \times \xi & ... &  \beta_{u} \\
		\beta_{l}& ... & \beta_{l} +\kappa'' \times \xi& \alpha_{j+1}& \beta_{l} -(\kappa''+1) \times \xi & ... &  \beta_{u}\\
		\vdots&\vdots&\vdots&\vdots&\vdots&\vdots&\vdots\\
		\beta_{l}& ... & \beta_{l} +\kappa''' \times \xi& \alpha_{j+\mathbb{W}-1}& \beta_{l} -(\kappa'''+1) \times \xi & ... &  \beta_{u}
	\end{bmatrix}
\end{equation}
the position of each integrated element is uniquely identified. Then, all the data from sliding window in the reconstructed series is replaced with the last element in original subset divided only keeping position information of former elements:
\begin{equation}\scriptsize
	\ell_{j}  = \begin{bmatrix}
		\beta_{l}& ... & \beta_{l} +\kappa' \times \xi& \alpha_{j+\mathbb{W}-1}& \beta_{l} +(\kappa'+1) \times \xi & ... &  \beta_{u} \\
		\beta_{l}& ... & \beta_{l} +\kappa'' \times \xi&\alpha_{j+\mathbb{W}-1}& \beta_{l} +(\kappa''+1) \times \xi & ... &  \beta_{u}\\
		\vdots&\vdots&\vdots&\vdots&\vdots&\vdots&\vdots\\
		\beta_{l}& ... & \beta_{l} +\kappa''' \times \xi& \alpha_{j+\mathbb{W}-1}& \beta_{l} +(\kappa'''+1) \times \xi & ... &  \beta_{u}
	\end{bmatrix}
\end{equation}
the simplified form of it can be given as:
\begin{equation}
		\ell_{j}  = \begin{bmatrix}
		\beta_{l}& ... & x_1& \underline{\alpha}& \mathring{x}_1 & ... &  \beta_{u} \\
		\beta_{l}& ... & x_2& \underline{\alpha}& \mathring{x}_2 & ... &  \beta_{u} \\
		\vdots&\vdots&\vdots&\vdots&\vdots&\vdots&\vdots\\
		\beta_{l}& ... & x_p&\underline{\alpha}&\mathring{x}_p& ... &  \beta_{u} \\
	\end{bmatrix}
\end{equation}
where $p \in [1, \mathbb{W}]$ and the final input to the proposed network is:
\begin{equation}
	\mathbb{T}_{Input} = \{\ell_{1},\ell_{2},...,\ell_{c}\}
\end{equation}
%\begin{algorithm}[]  %其中这里面不能有H不然会报错，不过不影响结果
%	\caption{Data Reconstruction}%算法名字
%	\LinesNumbered %要求显示行号
%	\KwIn{The original time series data $\mathbb{T} = \{(t_1,\zeta_1),(t_2,\zeta_2),...,(t_{n-1},\zeta_{n-1}),(t_n,\zeta_n)\}$}%输入参数
%	\KwOut{The final input to the proposed network $\mathbb{T}_{Input} = \{\ell_{1},\ell_{2},...,\ell_{c}\}$}%输出
%	\tcc{generate differential sequence using fuzzy intervals} %\;用于换行
%	\For{$\tilde{n} \in [2,n]$}{
%		$\alpha_{\tilde{n}-1} = \zeta_{\tilde{n}} - \zeta_{\tilde{n}-1}$
%	}
%	\tcc{$\mathbb{T}_{\Delta} = \{\alpha_1,\alpha_2,...,\alpha_{h}\}$}
%	\For{$\tilde{c} \in [1,c]$}{
%		only if\;
%		\If{condition}{
%			1\;
%		}
%	}
%	\For{$\tilde{n} \in [2,n]$}{
%		only if\;
%		\If{condition}{
%			1\;
%		}
%	}
%	\While{not at end of this document}{
%		if and else\;
%		\eIf{condition}{
%			1\;
%		}{
%			2\;
%		}
%	}
%	\ForEach{condition}{
%		\If{condition}{
%			1\;
%		}
%	}
%\end{algorithm}

\subsubsection{Padding Mechanism and Cropping}
Forth, one side of each row of input data is filled separately so that the length of data on both sides of the last element in original subset is the same. Assume data of row $p$ in $\ell_{\hbar}$ is vector $\vec{\varLambda_p}$, the process of padding can be given as:
\begin{equation}\small
	\vec{\breve{\varLambda}}_p =  \left\{
	\begin{array}{l}
		Concat(rep(\beta_{l})_{\mathbb{D}},\vec{\varLambda}_p^{\beta_{l}\Rightarrow x_p}), \vert\vert(\vec{\varLambda}_p^{\beta_{l}\Rightarrow x_p})\vert\vert<\vert\vert(\vec{\varLambda}_p^{\mathring{x}_p \Rightarrow \beta_u})\vert\vert \\
		Concat(\vec{\varLambda}_p^{\mathring{x}_p \Rightarrow\beta_u},rep(\beta_u)_{\mathbb{D}'}),
		\vert\vert(\vec{\varLambda}_p^{\beta_{l}\Rightarrow x_p})\vert\vert>\vert\vert(\vec{\varLambda}_p^{\mathring{x}_p \Rightarrow\beta_u})\vert\vert\\
	\end{array}
	\right.
\end{equation}
where $\vec{\varLambda}_p^{\dot{o}\Rightarrow \ddot{o}}$ represents a segmented vector which ranges from element $\dot{o}$ to $\ddot{o}$, $\vert\vert\vec{\varLambda}_p^{\dot{o}\Rightarrow \ddot{o}}\vert\vert$ is the length of $\vec{\varLambda}_p^{\dot{o}\Rightarrow \ddot{o}}$ and $Concat$ denotes the operation of concatenation of two vectors. Besides, $rep(\dot{o})_{\mathbb{D}}$ means creating a vector containing $\mathbb{D} $ copies of element $\dot{o}$ and $\mathbb{D} = \vert\vert(\vec{\varLambda}_p^{\mathring{x} \Rightarrow \beta_u})\vert\vert - \vert\vert(\vec{\varLambda}_p^{\beta_{l}\Rightarrow x})\vert\vert$ or $\mathbb{D}' = \vert\vert(\vec{\varLambda}_p^{\beta_{l}\Rightarrow x})\vert\vert-\vert\vert(\vec{\varLambda}_p^{\mathring{x} \Rightarrow\beta_u})\vert\vert$. 

Then, each row containing in $\ell_{\hbar}$ is padded ensuring lengths of two sides of the last element in original subset are equal. However, the operation of padding brings a problem that length of each row is not exactly the same which is difficult for neural network to acquire information and capture features. As a result, there is a need to crop redundant elements in each padded data row. Assume length of the shortest padded vector is $\mathbb{S}$ and the operation of cropping $\mathbb{M}$ elements which lie from both ends of the vector $\breve{\varLambda}_p$ to its centre is $Crop_{\mathbb{M}}$, the process of cropping is defined as:
\begin{equation}
	 \underline{\vec{\varLambda}}_{p} = Crop_{\mathbb{M}}(\vec{\breve{\varLambda}}_p)
\end{equation}
where $\mathbb{M} = (\vert\vert\vec{\breve{\varLambda}}_p\vert\vert-\mathbb{S})/2$ and $\underline{\vec{\varLambda}}_{p}$ is the cropped vector.
\subsubsection{Central-Bidirectional Atrous Algorithm}
Fifth, the processed information needs to be further extracted so that subsequent networks can capture more useful information and avoid unnecessary calculations. Because of the unique nature of the reconstructed timing data, the atrous algorithm is modified to obtain data from the centre to both sides of each segment, which reserves the nearest observation value and corresponding position distribution information from the prediction object. Assume the leftmost and rightmost element in $\underline{\vec{\varLambda}}_{p}$ are $\epsilon_{l_p}$ and $\epsilon_{r_p}$, the input which is divided into two parts by the central element to central-bidirectional atrous algorithm (CBAA) is:
\begin{equation}
	\left\{
	\begin{array}{l}
			\ell_{j}^{l} = [\underline{\vec{\varLambda}}_{1_{\mathring{x}_{1} + g \times d}}^{\ \mathring{x}_{1}\Rightarrow\epsilon_{r_1}},\underline{\vec{\varLambda}}_{2_{\mathring{x}_{2} + g \times d}}^{\ \mathring{x}_{2}\Rightarrow\epsilon_{r_2}},...,\underline{\vec{\varLambda}}_{{p-1}_{\mathring{x}_{p-1} + g \times d}}^{\ \mathring{x}_{p-1}\Rightarrow\epsilon_{r_{p-1}}},\underline{\vec{\varLambda}}_{p_{\mathring{x}_{p} + g \times d}}^{\ \mathring{x}_{p}\Rightarrow\epsilon_{r_p}}]\\
			\ell_{j}^{r} = [\cev{\underline{\varLambda}}_{1_{x_1-g\times d}}^{\epsilon_{l_1}\Rightarrow x_1},\cev{\underline{\varLambda}}_{2_{x_2-g\times d}}^{\epsilon_{l_2}\Rightarrow x_2},...,\cev{\underline{\varLambda}}_{{p-1}_{x_{p-1}-g\times d}}^{\epsilon_{l_{p-1}}\Rightarrow x_{p-1}},\cev{\underline{\varLambda}}_{p_{x_p-g\times d}}^{\epsilon_{l_p}\Rightarrow x_p}] \\
	\end{array}
	\right.
\end{equation}
where $\ell_{j}^{l}$ and $\ell_{j}^{r}$ denote left and right part of cropped vector $\underline{\vec{\varLambda}}_{p}$ and the directions of filters on them are opposite. Moreover, assume a filter $f:\{0,...,v-1\}$ and the operation of CBAA, $\mathbb{F}$, starting with elements $\mathring{x}_{p}$ and $x_p$ is defined as:
\begin{equation}\footnotesize
	\mathbb{F}(\underline{\alpha})_j =Concat( \sum_{g=0}^{v-1}f(g)\cdot\ell_{j}^{l},\underline{\alpha}, \sum_{g=0}^{v-1}f(g)\cdot\ell_{j}^{r})
\end{equation}
where $\cdot$ represents the operation of dilated convolution, $d$ is the factor of dilation, $v$ means the filter size, $\mathring{x}_{p} + g \times d$ and $x_p-g\times d$ account for the direction of movement of filters. 

When $d=1$, the form of atrous algorithm degenerates into regular convolution. A larger dilation factor enables the algorithm to capture features at a longer range. In original atrous algorithm, the operation of dilation is utilized to enlarge the receptive field without reduce sizes of feature maps. But in the CBAA, the dilated convolution is mainly used to construct efficient maps with proper sizes containing underlying features of historic information via multiple non-adjacent fuzzy intervals.
\subsection{Semi-Asymmetric Convolutional Architecture}
Sixth, a semi-asymmetric convolutional neural network (SACNN) is designed to aggregate information and produce differential predictions. SACNN is made up of a stack of one module which is called SAC block. The SAC block consists of two parts, the first part is the batchnorm layer $Bn$ which is defined as:
\begin{equation}
	\tilde {C}_j=Bn(\mathbb{F}(\underline{\alpha})_j)=\frac{\mathbb{F}(\underline{\alpha})_j-\bar{\mathbb{F}(\underline{\alpha})_j}}{\sqrt{\sigma(\mathbb{F}(\underline{\alpha})_j)+\epsilon}}*\gamma+\delta
\end{equation}
where $\bar{\mathbb{F}(\underline{\alpha})_j}$ and $\sigma(\mathbb{F}(\underline{\alpha})_j)$ denotes mean and standard-deviation of $\mathbb{F}(\underline{\alpha})_j$, $\gamma$ and $\delta$ are learnable parameter vectors whose size is the number of channel of input. The output $\tilde {A}_j$ is supposed to be sent into the next part, semi-asymmetric convolutional layer $Sa$ which consists of $L$ combinations of $X$ horizontal and vertical filters $\check{f}_V \in \mathbb{R}^{V\times 1}$ and $\check{f}_H \in \mathbb{R}^{1\times H}$	. Assume the input $\tilde {C}_j \in \mathbb{R}^{H'\times V'\times Y}$ with $H'\times V'$ feature map and $Y$ channels, the process of generating output can be defined as:
\begin{equation}
		B_j = Sa(\tilde {C}_j) = [\check{f}_V\diamond(\check{f}_H\diamond\tilde {C}_j)]_{\times L},\  B_j \in \mathbb{R}^{H''\times V'' \times  X}
\end{equation}
where $\diamond$ represents semi-asymmetric convolution and $OUT = X / Y$ is the lifting factor of number of input's to output's channels. When $V=H$, $Sa$ degenerates into the form of regular asymmetric convolution. Before outputting the final values, the information is expected to be sent into two linear layers:
\begin{equation}
	Q_j = SAC(\mathbb{F}(\underline{\alpha})_j) = (B_j A^{T} + b)A'^{T} + b'
\end{equation}
where $A$ and $A'$ are the learnable weights of the module of shape which is transposed to times the original input $B_j$ and $b$ and $b'$ are the biases to be added. Then, $Q_j$ is the prediction which the proposed model produce on the first order difference sequence $\mathbb{T}_{\Delta}$.
%\begin{algorithm}[]  %其中这里面不能有H不然会报错，不过不影响结果
%	\caption{algorithm caption}%算法名字
%	\LinesNumbered %要求显示行号
%	\KwIn{input parameters A, B, C}%输入参数
%	\KwOut{output result}%输出
%	some description\; %\;用于换行
%	\For{condition}{
%		only if\;
%		\If{condition}{
%			1\;
%		}
%	}
%	\While{not at end of this document}{
%		if and else\;
%		\eIf{condition}{
%			1\;
%		}{
%			2\;
%		}
%	}
%	\ForEach{condition}{
%		\If{condition}{
%			1\;
%		}
%	}
%\end{algorithm}
\subsection{Restore Output of Network to Original Position and Make Prediction}
\begin{figure}
	\centering %表示居中
	\includegraphics[width = 8.8cm]{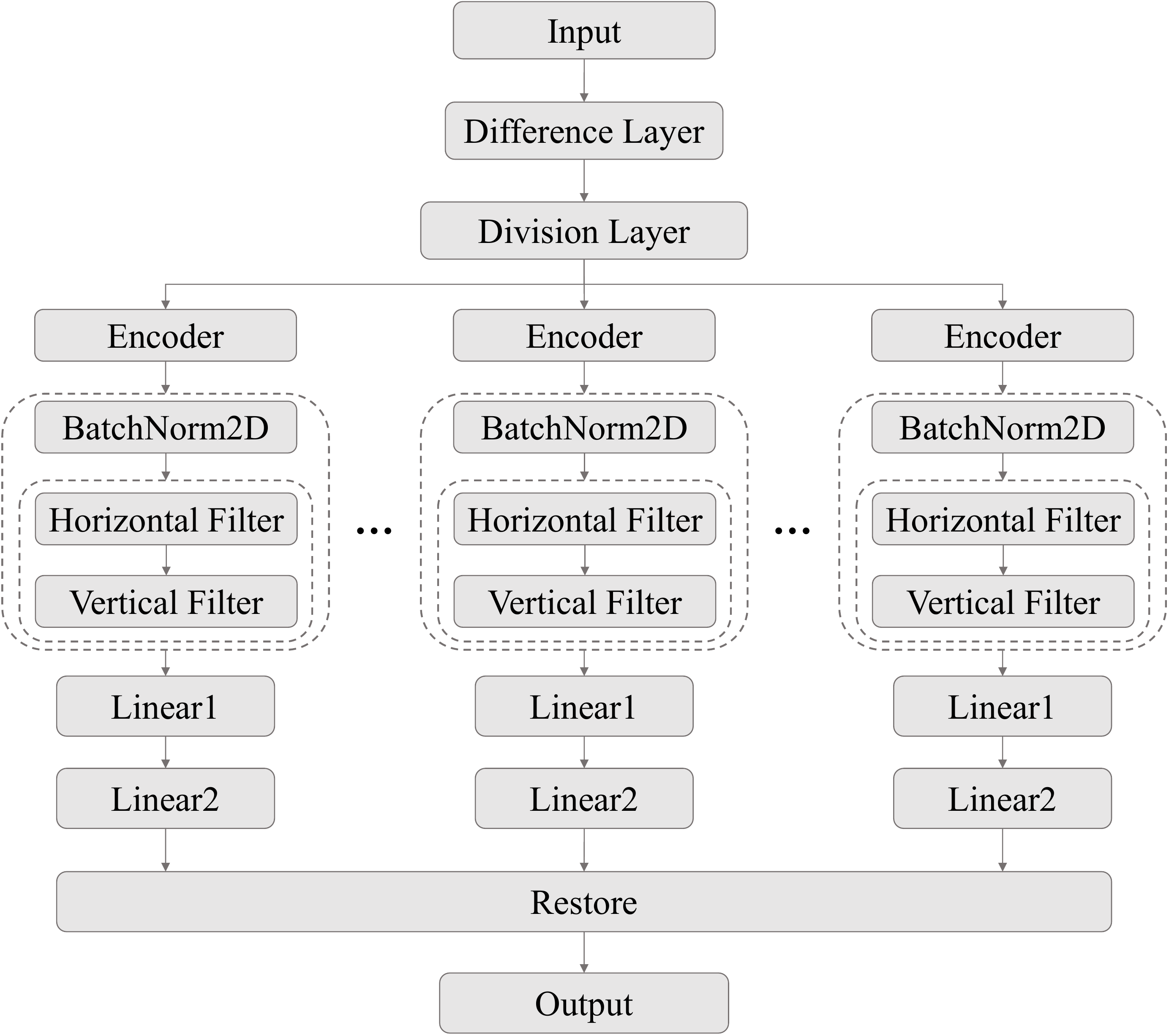}
	\caption{Details of the Proposed Model}
	\label{2}
\end{figure}

Seventh, restore the differential prediction $Q_j$ to the original time series $\mathbb{T}$. The process of a sliding window generating corresponding prediction is given as:
\begin{equation}
	\hat{\zeta}_{j+\mathbb{W}+1} = \zeta_{j+\mathbb{W}} + Q_{j}
\end{equation}
on the training data, the proposed model is expected to approximate trend of changes of time series. For the prediction of value beyond the known time series data, the prediction is made as:
\begin{equation}
	\hat{\zeta}_{n+1} = \zeta_{n} + Q_{n-\mathbb{W}}
\end{equation}
the process of producing prediction of the proposed model is illustrated in Fig.\ref{2}.

%\begin{algorithm}[]  %其中这里面不能有H不然会报错，不过不影响结果
%	\caption{algorithm caption}%算法名字
%	\LinesNumbered %要求显示行号
%	\KwIn{input parameters A, B, C}%输入参数
%	\KwOut{output result}%输出
%	some description\; %\;用于换行
%	\For{condition}{
%		only if\;
%		\If{condition}{
%			1\;
%		}
%	}
%	\While{not at end of this document}{
%		if and else\;
%		\eIf{condition}{
%			1\;
%		}{
%			2\;
%		}
%	}
%	\ForEach{condition}{
%		\If{condition}{
%			1\;
%		}
%	}
%\end{algorithm}

\begin{table*}[htbp] \footnotesize
	\renewcommand{\arraystretch}{0.95}
	\caption{MEAN MAE RESULTS OF UNIVARIATE DATASETS}
	\label{table1}
	\centering
	\begin{threeparttable}
		\setlength{\tabcolsep}{0.6mm}{
			\begin{tabular}{l| c c c c c c c c c c}
				\toprule
				Dataset&Naive & SES & Theta & TBATS & ETS & ARIMA & PR& CatBoost &  FFNN & DeepAR \\
				\midrule
				M1 Yearly & 221512.32&  171353.41 &  152799.26 & 103006.90 & 146110.11 & 145608.87 &   134246.38 &  215904.20 & 136238.80 & 152084.40\\
				M1 Quarterly & 3350.81 &2206.27 & 1981.96 &  2326.46 & 2088.15 & 2191.10 &  1630.38 & 1802.18 & 1617.39 &1951.14\\
				M1 Monthly &2866.26 &2259.04 & 2166.18 & 2237.50 & 1905.28 &  2080.13 &  2088.25 & 2052.32 & 2162.58&1860.81\\
				M3 Yearly &1563.64  &1022.27 & 957.40 & 1192.85 & 1031.40 & 1416.31 &  1018.48 &  1163.36 &  1082.03 & 994.72\\
				M3 Quarterly & 711.65&571.96 & 486.31 & 561.77 & 513.06 &  559.40 &  519.30 & 593.29 &  528.47 & 519.35\\
				M3 Monthly &1002.94 &743.41 & 623.71 & 630.59 &  626.46 & 654.80 &  692.97 & 732.00 & 692.48 &728.81\\
				M3 Other & 452.11&277.83 &  215.35 & 189.42 & 194.98 & 193.02 & 234.43 &  318.13 & 240.17 & 247.56\\
				M4 Yearly &1487.58 &1009.06 &  890.51 & 960.45 &  920.66 & 1067.16 & 875.76 & 929.06 & - &-\\
				M4 Quarterly & 838.19 &622.57 &574.34& 570.26 &573.19 &604.51 &610.51 &609.55 &631.01 &597.16\\
				M4 Monthly & 835.69&625.24 &563.58 &589.52 &582.60 &575.36 &596.19 &611.69 &612.52 &615.22\\
				M4 Weekly & 480.94&336.82 &333.32 &296.15 &335.66 &321.61 &293.21 &364.65 &338.37 &351.78\\
				M4 Daily &255.42 &178.27 &178.86 &176.60 &193.26 &179.67 &181.92 &231.36 &177.91 &299.79\\
				M4 Hourly &399.84  &1218.06 &1220.97 &386.27 &3358.10 &1310.85 &257.39 &285.35 &385.49 &886.02\\
				Tourism Yearly & 117966.55 &95579.23 &90653.60 &94121.08 &94818.89 &95033.24 &82682.97 &79567.22 &79593.22 &71471.29\\
				Tourism Quarterly & 13988.39 &15014.19 &\textcolor{red}{\textbf{7656.49}}&9972.42 &8925.52 &10475.47 &9092.58 &10267.97& 8981.04 &9511.37\\
				Tourism Monthly& 3019.44 &5302.10 & 2069.96 & 2940.08& 2004.51 & 2536.77 & 2187.28 & 2537.04&  2022.21 & \textcolor{red}{\textbf{1871.69}}\\
				CIF 2016 &650535.53 &581875.97 &714818.58 &855578.40 &642421.42 &469059.49 &563205.57& 603551.30 &1495923.44 &3200418.00\\
				Aus. Electricity Demand & 241.77 &659.60 &665.04 &370.74 &1282.99 &1045.92 &247.18 &241.77& 258.76 &302.41\\
				Dominick & 5.86& 5.70 & 5.86 & 7.08 & 5.81 & 7.10 & 8.19 & 8.09 & 5.85 & 5.23\\
				Bitcoin &6.57$\times$10$^{17}$ &5.33$\times$10$^{18}$ & 5.33$\times$10$^{18}$ & 9.9$\times$10$^{17}$ & 1.1$\times$10$^{18}$ & 3.62$\times$10$^{18}$ & 6.66$\times$10$^{17}$ & 1.93$\times$10$^{18}$ & 1.45$\times$10$^{18}$ & 1.95$\times$10$^{18}$\\
				Pedestrian Counts & 65.59&170.87 &170.94 &222.38 &216.50&635.16 &44.18 &\textcolor{red}{\textbf{43.41} }&46.41 &44.78\\
				Vehicle Trips & 13.37&29.98 &30.76 &21.21 &30.95 &30.07 &27.24 &22.61 &22.93 &22.00\\
				KDD Cup & 72.17 &42.04 &42.06 &39.20 &44.88 &52.20 &36.85 &34.82 &37.16 &48.98\\
				Weather & 2.79  &2.24 &2.51 &2.30 &2.35 &2.45 &8.17 &2.51 &2.09 &2.02\\
				%		NN5 Daily &  6.63 &3.80 &\textcolor{red}{\textbf{3.70} }&3.72 &4.41 &5.47 &4.22 &4.06 &3.94\\
				%		NN5 Weekly & 15.66 &15.30& 14.98 &15.70 &15.38 &14.94 &15.29 &15.02 &14.69\\
				%		Solar 10 Minutes &  3.28 &  3.29 &  8.77 & 3.28 & 2.37 & 3.28 & 5.69 & 3.28 &3.28\\
				%		Solar Weekly&  1202.39 &  1210.83 &  908.65 &  1131.01 &  839.88&1044.98 &  1513.49 &  1050.84 &721.59\\
				%		Electricity Hourly &  845.97& 846.03 &574.30 &1344.61 &868.20 &537.38 &407.14 &354.39& 329.75\\
				%		Electricity Weekly & 74149.18 &74111.14 &24347.24 &67737.82 &28457.18 &44882.52 &34518.43 &27451.83 &50312.05\\
				%		Carparts & 0.55 & 0.53 & 0.58 & 0.56 & 0.56 & 0.41 & 0.53&  \textcolor{red}{\textbf{0.39} }& \textcolor{red}{\textbf{0.39}}\\
				%		Traffic Hourly & 0.03 &0.03 &0.04& 0.03 &0.04 &0.02 &0.02 &0.01&0.01\\
				%		Traffic Weekly &  1.12 &1.13 &1.17 &1.14& 1.22& 1.13& 1.17& 1.15& 1.18\\
				%		Rideshare&  6.29 &7.62 &6.45& 6.29 &3.37& 6.30& 6.07 &6.59 &6.28\\
				%		Hospital & 21.76 &18.54 &\textcolor{red}{\textbf{17.43}} &17.97& 19.60 &19.24& 19.17& 22.86 &18.25\\
				%		Temperature Rain&  8.18&  8.22 & 7.14 & 8.21 & 7.19 & 6.13 & 6.76 & 5.56 & 5.37\\
				Sunspot&0.14 &4.93&4.93 &2.57 &4.93 &2.57& 3.83 &2.27 &7.97 &0.77\\
				Saugeen River Flow&  12.49&  21.50 &21.49 &22.26 &30.69 &22.38 &25.24& 21.28 &22.98 &23.51\\
				US Births& 1497.36 &1192.20 &586.93 &\textcolor{red}{\textbf{399.00}}&419.73 &526.33 &574.93 &441.70 &557.87 &424.93\\
				
				\midrule
				Dataset  & N-BEATS & WaveNet & Transformer & MSS$^{*}$& FEDformer$^{*}$& NetAtt$^{*}$&  Pyraformer$^{*}$  & PFSD$^{*}$ &  Informer & Ours \\
				\midrule
				M1 Yearly &   173300.20 &   284953.90 & 164637.90 & 59228.64 & 124729.30 &  66409.64 & 127110.48 & \textcolor{red}{\textbf{51417.35}}&-& 66062.77\\
				M1 Quarterly &  1820.25 & 1855.89&  1864.08 & 1686.22&1683.57 & 1727.60 & 1721.32 & \textcolor{red}{\textbf{1231.13}}&-&1234.77  \\
				M1 Monthly & 1820.37 & 2184.42& 2723.88 &2063.19& 2394.66& 1720.12& 2421.01 & 1952.81&-& \textcolor{red}{\textbf{1620.47}} \\
				M3 Yearly &  962.33 &  987.28 & 924.47 & 933.80&873.74 & 906.63 & 891.88 & 858.70 &-&\textcolor{red}{\textbf{530.78}} \\
				M3 Quarterly & 494.85 & 523.04& 719.62 & 538.85&623.58 & 591.25 & 711.46 &473.84&-&\textcolor{red}{\textbf{307.63}}  \\
				M3 Monthly & 648.60 & 699.30 &  798.38 &1127.37& 728.60 & 1014.96 & 693.24 & 912.28&-&\textcolor{red}{\textbf{547.31}} \\
				M3 Other & 221.85&  245.29 &  239.24 & 229.01&217.03 & 297.44& 196.81 &210.80&-& \textcolor{red}{\textbf{92.47}} \\
				M4 Yearly & - & - & - & 792.87 & 730.24 & 967.37& 757.92 &528.36&-&\textcolor{red}{\textbf{415.63}}  \\
				M4 Quarterly & 580.44 &  596.78 &  637.60& 560.72 & 594.24 & 617.30 & 608.55 &445.09&-& \textcolor{red}{\textbf{382.57}} \\
				M4 Monthly & 578.48 &  655.51 & 780.47 & 644.51 & 688.95 & 781.42& 694.29 & 608.31&-&\textcolor{red}{\textbf{353.34}}\\
				M4 Weekly & 277.73 & 359.46 & 378.89 & 301.26 & 317.16 & 322.59& 295.60&250.68&-&\textcolor{red}{\textbf{222.42}}\\
				M4 Daily &  190.44 & 189.47 &  201.08 & 173.20 & 167.05& 207.44& 161.36 & 103.28&-& \textcolor{red}{\textbf{62.54}}\\
				M4 Hourly &  425.75 &  393.63 & 320.54 & 1355.21 & 246.33 & 1841.90& 228.87 & 999.83&-&\textcolor{red}{\textbf{94.06}}\\
				Tourism Yearly &  70951.80 & 69905.47 & 74316.52 & - & -& -& - &- &-& \textcolor{red}{\textbf{53029.08}}  \\
				Tourism Quarterly & 8640.56 & 9137.12 & 9521.67 & - & - & - & - & -&-&7799.49 \\
				Tourism Monthly& 2003.02 & 2095.13 & 2146.98 & -& - & - & -&  -&-&2227.20 \\
				CIF 2016 & 679034.80 & 5998224.62& 4057973.04 & - & - & - & - & - &-&\textcolor{red}{\textbf{226103.58}} \\
				Aus. Electricity Demand & 213.83 & 227.50 & 231.45 & - & - & - & - & - &-&\textcolor{red}{\textbf{42.48}} \\
				Dominick &  8.28 & 5.10 & 5.18& 5.39 & 5.10 & 6.02 & 5.16 & 4.80&-&\textcolor{red}{\textbf{4.43}}\\
				Bitcoin &  1.06$\times$10$^{18}$ & 2.46$\times$10$^{18}$ & 2.61$\times$10$^{18}$& - & - & - & - & - &-&\textcolor{red}{\textbf{3.77$\times$10$^{17}$}} \\
				Pedestrian Counts &  66.84 & 46.46 & 47.29 & - & - & - & - & - &-&66.66 \\
				Vehicle Trips & 28.16 & 24.15 & 28.01 & -& - & - & - & - &-&\textcolor{red}{\textbf{12.36}} \\
				KDD Cup &  49.10 & 37.08 &  44.46 & - & - & - & - & - &-&\textcolor{red}{\textbf{5.92}} \\
				Weather & 2.34 & 2.29 & 2.03 & - & - & -& - & - &-& \textcolor{red}{\textbf{1.87}}\\
				Sunspot &  14.47 &  0.17& \textcolor{red}{\textbf{0.13}} & -& - & - & - & -& 19.43& 0.14 \\
				Saugeen River Flow &  27.92 & 22.17 & 28.06& - & -& - & - & - &28.59&\textcolor{red}{\textbf{8.77}}\\
				US Births&  422.00&  504.40&  452.87 & - & - & -& -& -& 609.43&538.37\\
				\bottomrule
		\end{tabular}}
		%\begin{tablenotes}
		%	\footnotesize
		%	\item[1] The results produced by naive forecasting dose not participate in the comparison with experimental results of other models because of its particularity in forecasting strategy which is provided only for a simple reference.
		%\end{tablenotes}
	\end{threeparttable}
\end{table*}

\begin{table*}[htbp] \footnotesize
	\renewcommand{\arraystretch}{0.95}
	\caption{MEAN RMSE RESULTS OF UNIVARIATE DATASETS}
	\label{table2}
	\centering
	\begin{threeparttable}
		\setlength{\tabcolsep}{0.6mm}{
			\begin{tabular}{l| c c c c c c c c c c}
				\toprule
				Dataset &Naive &SES & Theta & TBATS & ETS & ARIMA & PR& CatBoost &  FFNN & DeepAR\\
				\midrule
				M1 Yearly &237288.10 &193829.49  &171458.07&116850.90&167739.02&175343.75&152038.68&237644.50&154309.80&173075.10\\
				M1 Quarterly & 3798.89&2545.73 &2282.65&2673.91&2408.47&2538.45&1909.31&2161.01&1871.85&2313.32 \\
				M1 Monthly &3533.38 &2725.83 &2564.88&2594.48&2263.96&2450.61&2478.88&2461.68&2527.03&2202.19\\
				M3 Yearly &1729.92 &1172.85 &1106.05&1386.33&1189.21&1662.17&1181.81&1341.70&1256.21&1157.88\\
				M3 Quarterly & 804.54&670.56 &567.70&653.61&598.73&650.76&605.50&697.96&621.73&606.56\\
				M3 Monthly & 1193.11&893.88 &753.99&765.20&755.26&790.76&830.04&874.20&833.15&873.71\\
				M3 Other & 479.26&309.68 &242.13&216.95&224.08&220.77&262.31&349.90&268.99&277.74\\
				M4 Yearly &1612.24 &1154.49 &1020.48&1099.95&1052.12&1230.35&1000.18&1065.02&-&-\\
				M4 Quarterly & 955.55&732.82  &673.15&672.74&674.27&709.99&711.93&714.21&735.84&700.32\\
				M4 Monthly & 1002.72& 755.45&683.72&743.41&705.70&702.06&720.46&734.79&743.47&740.26\\
				M4 Weekly & 553.29& 412.60&405.17&356.74&408.50&386.30&350.29&420.84&399.10&422.18\\
				M4 Daily &293.15 &209.75 &210.37&208.36&229.97&212.64&213.01&263.13&209.44&343.48\\
				M4 Hourly & 477.27 & 1476.81&1483.70&469.87&3830.44&1563.05&312.98&344.62&467.89&1095.10\\
				Tourism Yearly & 130104.42 &106665.20 &99914.21&105799.40&104700.51&106082.60&89645.61&87489.00&87931.79&78470.68\\
				Tourism Quarterly & 17050.68 & 17270.57&\textcolor{red}{\textbf{9254.63}}&12001.48&10812.34&12564.77&11746.85&12787.97&12182.57&11761.96\\
				Tourism Monthly& 3873.31 &7039.35 &2701.96&3661.51&\textcolor{red}{\textbf{2542.96}}&3132.40&2739.43&3102.76&2584.10&2359.87\\
				CIF 2016 &712332.30 &657112.42 &804654.19&940099.90&722397.37&526395.02&648890.31&705273.30&1629741.53&3532475.00\\
				Aus. Electricity Demand &  340.70& 766.27&771.51&446.59&1404.02&1234.76&319.98&300.55&330.91&357.00\\
				Dominick &8.31 & \textcolor{red}{\textbf{6.48}}&6.74&8.03&6.59&7.96&9.44&9.15&6.79&6.67\\
				Bitcoin& 8.27$\times$10$^{17}$& 5.35$\times$10$^{18}$&5.35$\times$10$^{18}$&1.16$\times$10$^{18}$&1.22$\times$10$^{18}$&3.96$\times$10$^{18}$&8.29$\times$10$^{18}$&2.02$\times$10$^{18}$&1.57$\times$10$^{18}$&2.02$\times$10$^{18}$\\
				Pedestrian Counts & 94.29& 228.14&228.20&261.25&278.26&820.28&61.84&\textcolor{red}{\textbf{60.78}}&67.17&65.77\\
				Vehicle Trips &18.13 &36.53 &37.44&25.69&37.61&34.95&31.69&27.28&27.88&26.46\\
				KDD Cup & 111.97 &73.81&73.83&71.21&76.71&82.66&68.20&65.71&68.43&80.19\\
				Weather &  3.80 &2.85&3.27&2.89&2.96&3.07&9.08&3.09&2.81&2.74\\
				
				Sunspot &0.53 &4.95&4.95&2.97&4.95&2.96&3.95&2.38&8.43&1.14\\
				Saugeen River Flow & 22.30 & 39.79&39.79&42.58&50.39&43.23&47.70&39.32&40.64&45.28\\
				US Births& 1921.21 & 1369.50&735.51&\textcolor{red}{\textbf{606.54}}&607.20&705.51&732.09&618.38&726.72&683.99\\
				\midrule
				Dataset  & N-BEATS & WaveNet & Transformer & MSS$^{*}$& FEDformer$^{*}$& NetAtt$^{*}$&  Pyraformer$^{*}$  & PFSD$^{*}$ &Informer &Ours\\
				\midrule
				M1 Yearly &  192489.80&312821.80&182850.60&68119.81&143607.73&81092.33&145991.89&\textcolor{red}{\textbf{59867.94}}&- &80553.54\\
				M1 Quarterly &  2267.27 &2271.68&2231.50&1977.00&1992.56&2057.60&2026.49&\textcolor{red}{\textbf{1458.75}}& -&1519.10\\
				M1 Monthly & 2183.37&2578.93&3129.84&2427.46&2918.05&\textcolor{red}{\textbf{2024.08}}&2957.84&2369.96& -&2085.94 \\
				M3 Yearly & 1117.37 &1147.62&1084.75&1079.09&1019.83&1061.72&1054.66&981.94&- &\textcolor{red}{\textbf{655.87}}\\
				M3 Quarterly &  582.83&606.75&819.18&636.68&735.21&693.52&810.20&568.22&- &\textcolor{red}{\textbf{384.34}}\\
				M3 Monthly & 796.91&845.30&948.40&1311.49&877.76&1193.29&836.84&1079.11&- &\textcolor{red}{\textbf{697.11}} \\
				M3 Other &  248.53&276.97&271.02&260.48&245.08&335.88&227.20&247.66&- &\textcolor{red}{\textbf{115.68}}\\
				M4 Yearly &  -&-&-&898.74&787.35&1173.95&816.41&606.06&- &\textcolor{red}{\textbf{516.66}}\\
				M4 Quarterly &  684.65&696.96&739.06&662.18&691.68&715.13&712.44&514.54&- &\textcolor{red}{\textbf{476.51}}\\
				M4 Monthly & 705.21&787.94&902.38&778.20&831.55&902.91&853.13&720.67&- &\textcolor{red}{\textbf{471.81}}\\
				M4 Weekly & 330.78&437.26&456.90&354.97&379.04&388.03&337.62&320.38&- &\textcolor{red}{\textbf{286.07}}\\
				M4 Daily   &221.69&220.45&233.63&205.22&192.67&249.70&183.30&118.88&- &\textcolor{red}{\textbf{84.87}}\\
				M4 Hourly &  501.19&468.09&391.22&1643.46&304.69&2124.99&284.09&1209.48&- &\textcolor{red}{\textbf{136.42}}\\
				Tourism Yearly &78241.67 &77581.31&80089.25&-&-&-&-&-&- &\textcolor{red}{\textbf{62680.50}}\\
				Tourism Quarterly & 11305.95&11546.58&11724.14&-&-&-&-&-& -&10014.33 \\
				Tourism Monthly&  2596.21&2694.22&2660.06&-&-&-&-&-&- &2932.16\\
				CIF 2016 & 772924.30 &6085242.41&4625974.00&-&-&-&-&-&-&\textcolor{red}{\textbf{288763.10}}\\
				Aus. Electricity Demand &  268.37&286.48&295.22&-&-&-&-&-&-&\textcolor{red}{\textbf{62.16}}\\
				Dominick &   9.78&6.81&6.63&7.39&6.97&7.02&6.89&6.56&-&6.60\\
				Bitcoin & 1.26$\times$10$^{18}$&2.55$\times$10$^{18}$&2.67$\times$10$^{18}$&&&&&&-&\textcolor{red}{\textbf{4.39$\times$10$^{17}$}} \\
				Pedestrian Counts &  99.33&67.99&70.17&-&-&-&-&-&-&98.13 \\
				Vehicle Trips &  33.56&28.99&32.98&-&-&-&-&-&-&\textcolor{red}{\textbf{23.58}}\\
				KDD Cup & 80.39&68.87&76.21&-&-&-&-&-&-&\textcolor{red}{\textbf{10.16}} \\
				Weather & 3.09&2.98&2.81&-&-&-&-&-&-&\textcolor{red}{\textbf{2.70}}\\
				Sunspot &  14.52 &0.66&\textcolor{red}{\textbf{0.52}}&-&-&-&-&-&20.31&0.53\\
				Saugeen River Flow &  48.91&42.99&49.12&-&-&-&-&-&44.42&\textcolor{red}{\textbf{13.36}} \\
				US Births& 627.74&768.81&686.51&-&-&-&-&-&734.44&679.99\\
				\bottomrule
		\end{tabular}}
		%\begin{tablenotes}
		%	\footnotesize
		%	\item[1] The results produced by naive forecasting dose not participate in the comparison with experimental results of other models because of its particularity in forecasting strategy which is provided only for a simple reference.
		%\end{tablenotes}
	\end{threeparttable}
\end{table*}

\begin{table*}[htbp] 
	\renewcommand{\arraystretch}{0.95}
	\caption{MEAN MAE RESULTS OF MULTIVARIATE DATASETS}
	\label{table3}
	\centering
	\begin{threeparttable}
		\setlength{\tabcolsep}{1.2mm}{
			\begin{tabular}{l| c c c c c c c c c c}
				\toprule%\footnotemark[1]
				Dataset&Naive& SES & Theta & TBATS & ETS & ARIMA & PR& CatBoost &  FFNN & DeepAR \\
				\midrule
				NN5 Daily&4.63 &6.63 &3.80 & \textcolor{red}{\textbf{3.70}}& 3.72& 4.41&5.47 & 4.22& 4.06&3.94\\
				NN5 Weekly &19.44 &15.66 & 15.30& 14.98& 15.70& 15.38&14.94 &15.29 &15.02 &14.69\\
				Web Traffic Daily& 484.67&363.43 & 358.73& 415.40& 403.23& 340.36&- &- &- &-\\
				Web Traffic Weekly & 2756.28 &2337.11&2373.98 &2241.84 & 2668.28& 3115.03& 4051.75&10715.36 & 2025.23&2272.58\\
				Solar 10 Minutes & \textcolor{blue}{\textbf{0.00}}&3.28& 3.29& 8.77&3.28 & 2.37& 3.28&5.69 &3.28 &3.28\\
				Solar Weekly & 998.99&1202.39& 1210.83& 908.65& 1131.01& 839.88&1044.98 & 1513.49& 1050.84&721.59\\
				Electricity Hourly & 279.78 &845.97& 846.03&574.30 &1344.61 & 868.20& 537.38& 407.14&354.39 &329.75\\
				Electricity Weekly &99675.88 &74149.18 & 74111.14& 24347.24& 67737.82& 28457.18&44882.52 &34518.43 &27451.83 &50312.05\\
				Carparts &0.66 &0.55 &0.53 & 0.58& 0.56& 0.56&0.41 &0.53 &\textcolor{red}{\textbf{0.39}} &\textcolor{red}{\textbf{0.39}}\\
				FRED-MD& 5607.17&2798.22&3492.84 & 1989.97& 2041.42& 2957.11& 8921.94&2475.68 &2339.57 &4263.36\\
				Traffic Hourly &0.01 &0.03 &0.03 &0.04 &0.03 &0.04 &0.02 & 0.02& 0.01&0.01\\
				Traffic Weekly &1.25 &1.12& 1.13& 1.17& 1.14&1.22 &1.13 &1.17 &1.15 &1.18\\
				Rideshare &1.61 &6.29 &7.62 &6.45 &6.29 &3.37 &6.30 &6.07 &6.59 &6.28\\
				Hospital &32.29 &21.76& 18.54& 17.43& 17.97& 19.60& 19.24& 19.17& 22.86&18.25\\
				COVID Deaths &310.84 &353.71 &321.32 &96.29 &85.59 &85.77 &347.98 & 475.15&144.14 &201.98\\
				Temperature Rain & 6.66&8.18&8.22 &7.14 & 8.21& 7.19& 6.13& 6.76&5.56 &5.37\\
				\midrule
				Dataset  & N-BEATS & WaveNet & Transformer & MSS$^{*}$& FEDformer$^{*}$& NetAtt$^{*}$&  Pyraformer$^{*}$  & PFSD$^{*}$ &Informer &Ours\\
				\midrule
				NN5 Daily & 4.92&3.97 &4.16 &- &- &- &- &- &4.07&4.92\\
				NN5 Weekly &\textcolor{red}{\textbf{14.19}} &19.34 & 20.34&- &- &- &- &- &19.45&15.09\\
				Web Traffic Daily & -& -& -& -& -& -& -& -&-&\textcolor{red}{\textbf{217.47}}\\
				Web Traffic Weekly & 2051.30& 2025.50& 3100.32&- &- &- &- &- &-&\textcolor{red}{\textbf{1437.49}}\\
				Solar 10 Minutes & 3.52&- &3.28 &3.36&3.18 &3.93&3.22 &2.17 &3.67&\textcolor{red}{\textbf{1.60}} \\
				Solar Weekly & 1172.64& 1996.89& 576.35& 841.69& \textcolor{red}{\textbf{479.30}}& 1247.77& 513.24&649.22&2360.71&700.31\\
				Electricity Hourly & 350.37&286.56 &398.80 & -& -& -& -& -&441.77&\textcolor{red}{\textbf{203.39}}\\
				Electricity Weekly & 32991.72& 61429.32& 76382.47& -& -& -& -&- &47773.67&\textcolor{red}{\textbf{15699.48}}\\
				Carparts & 0.98&0.40 &\textcolor{red}{\textbf{0.39}} & -&- &- &- &- &-&0.53\\
				FRED-MD& 2557.80&2508.40 & 4666.04& -& -& -& -&-&32700.73& \textcolor{red}{\textbf{596.54}}\\
				Traffic Hourly & 0.02& 0.02& 0.01&- &- &-& -&- &0.02&\textcolor{red}{\textbf{0.008 }}\\
				Traffic Weekly & 1.11& 1.20&1.42 & -&- &-& -&- &1.42&\textcolor{red}{\textbf{1.10}} \\
				Rideshare & 5.55& 2.75& 6.29& -& -& -& -&-&-&\textcolor{red}{\textbf{0.79}} \\
				Hospital & 20.18& 19.35& 36.19& -&- &- & -&- &38.82&\textcolor{red}{\textbf{16.40}}\\
				COVID Deaths & 158.81&1049.48 &408.66 & -& -&-&- &- &-&\textcolor{red}{\textbf{8.84}}\\
				Temperature Rain &7.28 & 5.81 & 5.24& -&-& -& -&- &-&\textcolor{red}{\textbf{4.56}}\\
				\bottomrule
		\end{tabular}}
		%\begin{tablenotes}
		%	\footnotesize
		%	\item[1] The results produced by naive forecasting dose not participate in the comparison with experimental results of other models because of its particularity in forecasting strategy which is provided only for a simple reference. For example, in Solar 10 Minutes dataset, naive forecasting achieve surprising results whose error is $\textcolor{blue}{\textbf{0.00}}$, which is unintuitive and unreasonable.
		%\end{tablenotes}
	\end{threeparttable}
\end{table*}

\begin{table*}[htbp] 
	\renewcommand{\arraystretch}{0.95}
	\caption{MEAN RMSE RESULTS OF MULTIVARIATE DATASETS}
	\label{table4}
	\centering
	\begin{threeparttable}
		\setlength{\tabcolsep}{1.2mm}{
			\begin{tabular}{l| c c c c c c c c c c}
				\toprule%\tnote{1}
				Dataset & Naive&SES & Theta & TBATS & ETS & ARIMA & PR& CatBoost &  FFNN & DeepAR \\
				\midrule
				NN5 Daily & 6.68&8.23& 5.28& \textcolor{red}{\textbf{5.20}}& 5.22& 6.05& 7.26& 5.73& 5.79&5.50\\
				NN5 Weekly &24.27 &18.82& 18.65& 18.53& 18.82& 18.55& 18.62& 18.67& 18.29&18.53\\
				Web Traffic Daily & 911.51&590.11&583.32 & 740.74& 650.43& 595.43& -& -& -&-\\
				Web Traffic Weekly &4020.90 &2970.78& 3012.39& 2951.87& 3369.64& 3777.28& 4750.26& 14040.64&2719.65 &2981.91\\
				Solar 10 Minutes&\textcolor{blue}{\textbf{0.00}} &7.23&7.23 &10.71 & 7.23& 5.55& 7.23& 8.73& 7.21&7.22\\
				Solar Weekly & 1350.79 &1331.26& 1341.55&1049.01 &1264.43 &967.87 &1168.18 &1754.22 &1231.54 &873.62\\
				Electricity Hourly & 414.29&1026.29&1026.36 &743.35 &1524.87 & 1082.44&689.85 &582.66 &519.06 &477.99\\
				Electricity Weekly & 104510.94&77067.87 &76935.58 &28039.73 &70368.97 &32594.81 &47802.08 &37289.74 &30594.15 &53100.26\\
				Carparts &1.17 &0.78 &0.78 &0.84 & 0.80& 0.81& \textcolor{red}{\textbf{0.73}}& 0.79&0.74 &0.74\\
				FRED-MD&6333.09 &3103.00& 3898.72& 2295.74& 2341.72&3312.46 &9736.93 &2679.38 &2631.4 &4638.71\\
				Traffic Hourly & 0.02 &0.04& 0.04& 0.05& 0.04& 0.04& 0.03&0.03 &0.02 &0.02\\
				Traffic Weekly &1.63 &1.51 &1.53 &1.53 &1.53 &1.54 &1.50 &1.50 &1.55 &1.51\\
				Rideshare & 1.98 &7.17& 8.60& 7.35&7.17 &4.80 & 7.18& 6.95& 7.14&7.15\\
				Hospital &39.54 &26.55& 22.59& 21.28& 22.02& 23.68&23.48 &23.45 &27.77 &22.01\\
				COVID Deaths &313.04 &403.41& 370.14& 113.00& 102.08& 100.46& 394.07& 607.92& 173.14&230.47\\
				Temperature Rain & 10.15&10.34& 10.36& 9.20&10.38 &9.22 &9.83 &8.71 &8.89 &9.11\\
				\midrule
				Dataset  & N-BEATS & WaveNet & Transformer & MSS$^{*}$& FEDformer$^{*}$& NetAtt$^{*}$&  Pyraformer$^{*}$  & PFSD$^{*}$ & Informer&Ours\\
				\midrule
				NN5 Daily &6.47 &5.75 &5.92 & -&- & -& -& -& 5.52&6.52\\
				NN5 Weekly & \textcolor{red}{\textbf{17.35}}& 24.16& 24.02& -& -& -& -& -&23.03&18.98\\
				Web Traffic Daily & -& -& -& -& -& -& -&- & -&\textcolor{red}{\textbf{465.84}}\\
				Web Traffic Weekly &2820.62 &2719.37 &3815.38 &- &- &- & -&- &-&\textcolor{red}{\textbf{2197.40}}\\
				Solar 10 Minutes & 6.62& -& 7.23& 6.94& 6.91&7.97& 7.18&5.28 &6.41&\textcolor{red}{\textbf{1.60}}\\
				Solar Weekly & 1307.78& 2569.26& 693.84& 972.45&\textcolor{red}{\textbf{609.94}} & 2493.06&672.54 &776.15 &2623.95&863.46\\
				Electricity Hourly & 510.91& 489.91&514.68 & -& -& -&- & -&629.88&\textcolor{red}{\textbf{302.56}}\\
				Electricity Weekly & 35576.83& 63916.89& 78894.67&- & -& -& -&- &54022.60&\textcolor{red}{\textbf{22540.95}}\\
				Carparts & 1.11&0.74 &0.74 & -& -&- & -& -&-&0.78\\
				FRED-MD& 2812.97& 2779.48&5098.91 & -& -&- &- &- &32867.61&\textcolor{red}{\textbf{708.38}}\\
				Traffic Hourly &0.02 &0.03 &0.02 & -& -&- & -&- &0.04&\textcolor{red}{\textbf{0.01}}\\
				Traffic Weekly & \textcolor{red}{\textbf{1.44}}& 1.61& 1.94& -&- &- &- &- &1.76&1.50\\
				Rideshare &6.23 &3.51 &7.17 & -& -& -& -&- &-&\textcolor{red}{\textbf{1.04}}\\
				Hospital & 24.18& 23.38& 40.48& -&- &- &- &- &44.25&\textcolor{red}{\textbf{20.45}}\\
				COVID Deaths & 186.54& 1135.41& 479.96& -& -& -& -& -&-&\textcolor{red}{\textbf{13.27}}\\
				Temperature Rain &11.03 &9.07 &9.01 & -&- &- &- &- &-&\textcolor{red}{\textbf{7.05}}\\
				\bottomrule
		\end{tabular}}
		%\begin{tablenotes}
		%	\footnotesize
		%	\item[1] The results produced by naive forecasting dose not participate in the comparison with experimental results of other models because of its particularity in forecasting strategy which is provided only for a simple reference. For example, in Solar 10 Minutes dataset, naive forecasting achieve surprising results whose error is $\textcolor{blue}{\textbf{0.00}}$, which is unintuitive and unreasonable.
		%\end{tablenotes}
	\end{threeparttable}
\end{table*}

\section{Experiments}
In this section, multiple experiments are conducted to evaluate the effectiveness and validity of the proposed method.
\subsection{Datasets Description}
In order to fully illustrate the performance of the proposed model, the comparison experiments are conducted on 43 datasets which are provided by monash time series forecasting archive (MTSFA) \cite{DBLP:conf/nips/GodahewaBWHM21}. Specifically, among them, there are 27 univariate and 16 multivariate datasets and they cover multiple domains, such as Tourism, Banking, Energy, Sales, Economic, Transport, Nature, Web and Health. Moreover, the datasets have different sampling rates such as yearly, quarterly and monthly, which also correspond disparate expected forecast horizons. 
\subsection{Baseline Methods for Comparison}
To demonstrate the performance improvement gained by the proposed model, we compare it with baseline methods, such as Naive (Forecasting)\footnote{The results produced by naive forecasting dose not participate in the comparison with experimental results of other models because of its particularity in forecasting strategy which is provided only for a simple reference. For example, on Solar 10 Minutes dataset, naive forecasting achieve surprising results whose error is $\textcolor{blue}{\textbf{0.00}}$, which is unintuitive and unreasonable.}, Simple Exponential Smoothing (SES) \cite{articleHolt}, Theta \cite{articleAssimakopoulos}, Trigonometric Box-Cox ARMA Trend Seasonal Model (TBATS) \cite{articleLivera}, Exponential Smoothing (ETS) \cite{inbookHyndman},  (Dynamic
Harmonic Regression-)ARIMA \cite{bookBox,hyndman2018forecasting}, Pooled Regression Model (PR) \cite{DBLP:journals/jors/TraperoKF15}, CatBoost \cite{DBLP:conf/nips/ProkhorenkovaGV18}, Feed-Forward Neural Network (FFNN) \cite{Goodfellow-et-al-2016}, DeepAR \cite{articleFlunkert}, N-BEATS \cite{DBLP:conf/iclr/OreshkinCCB20}, WaveNet \cite{borovykh2017conditional}, Transformer \cite{DBLP:conf/nips/VaswaniSPUJGKP17}, MSS$^{*}$ \cite{articleYuntong}, FEDformer$^{*}$ \cite{zhou2022fedformer}, NetAtt$^{*}$ \cite{DBLP:journals/asc/HuX22}, Pyraformer$^{*}$ \cite{DBLP:conf/iclr/LiuYLLLLD22}, PFSD$^{*}$ \cite{hu2022time} and Informer \cite{zhou2021informer}. The experimental results of these methods except naive forecasting are acquired from MTSFA and PFSD. Besides, the results of experiments of Naive (Forecasting) are generated by following the experimental rules given by MTSFA strictly.
\subsection{Evaluation Metrics}
Measurement of model performance is an important objective of the experiments. Mean Absolute Error (MAE) and Root Mean Square Error (RMSE) are selected to evaluate the accuracy of forecasting of chosen comparative models whose definitions are defined as:
\begin{equation}
	MAE = \frac{\sum_{i=1}^{N}\vert \hat{y}_i-y_i\vert}{N}
\end{equation}
\begin{equation}
	RMSE = \sqrt{\frac{\sum_{i=1}^{N}\vert \hat{y}_i-y_i\vert^{2}}{N}}
\end{equation}
where $\hat{y}_i$ represents the value of forecasting.
\begin{figure*}
	\centering
	\subfigure[\qquad \qquad MAE Comparison Among Models Without $^*$]{
		\begin{minipage}[t]{0.485\linewidth}
			\centering
			\includegraphics[width=9cm,height=9.8cm]{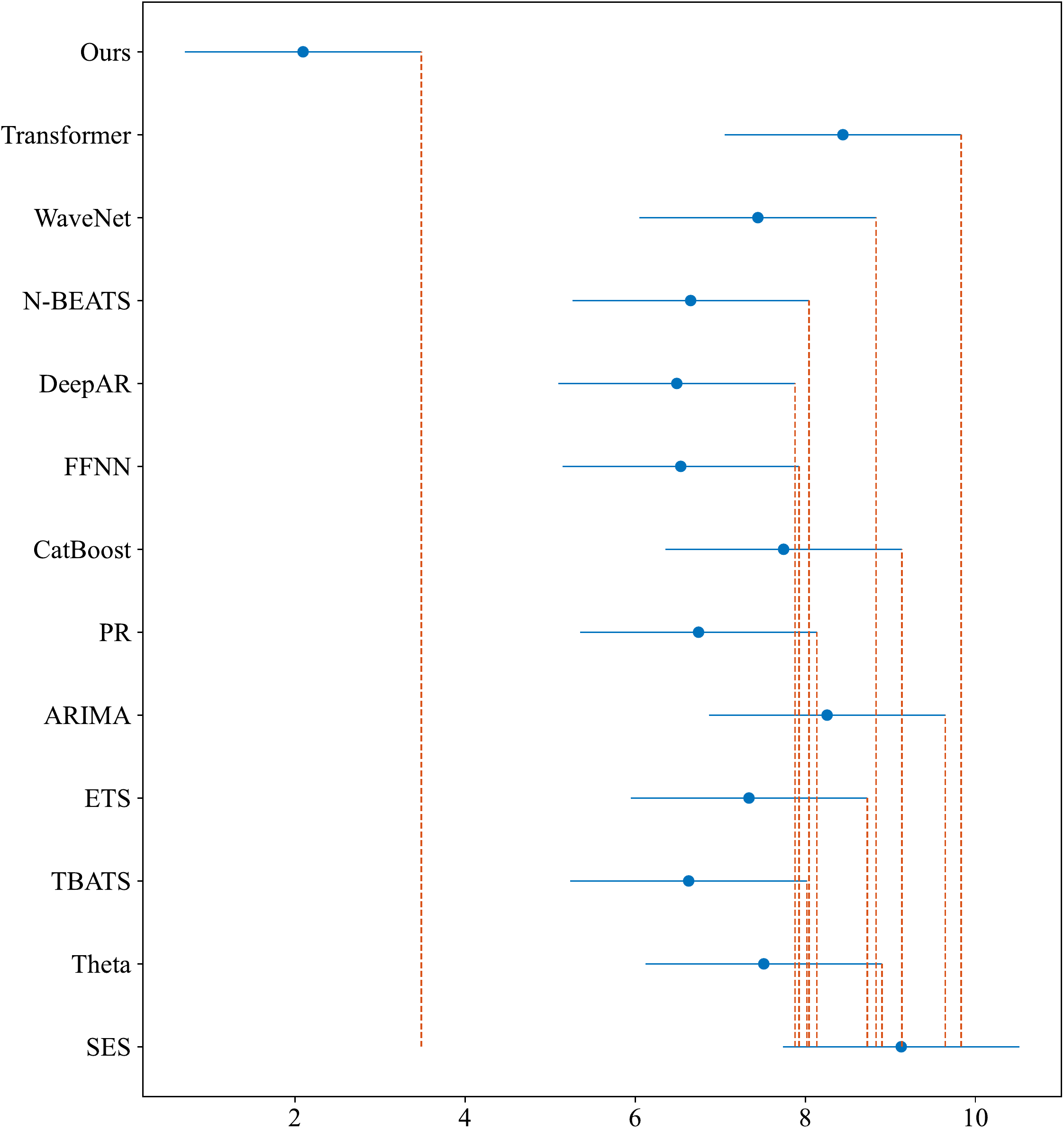}\\
		\end{minipage}%
	}%
	\subfigure[\qquad \qquad RMSE Comparison Among Models Without $^*$]{
		\begin{minipage}[t]{0.51\linewidth}
			\centering
			\includegraphics[width=9cm,height=9.812cm]{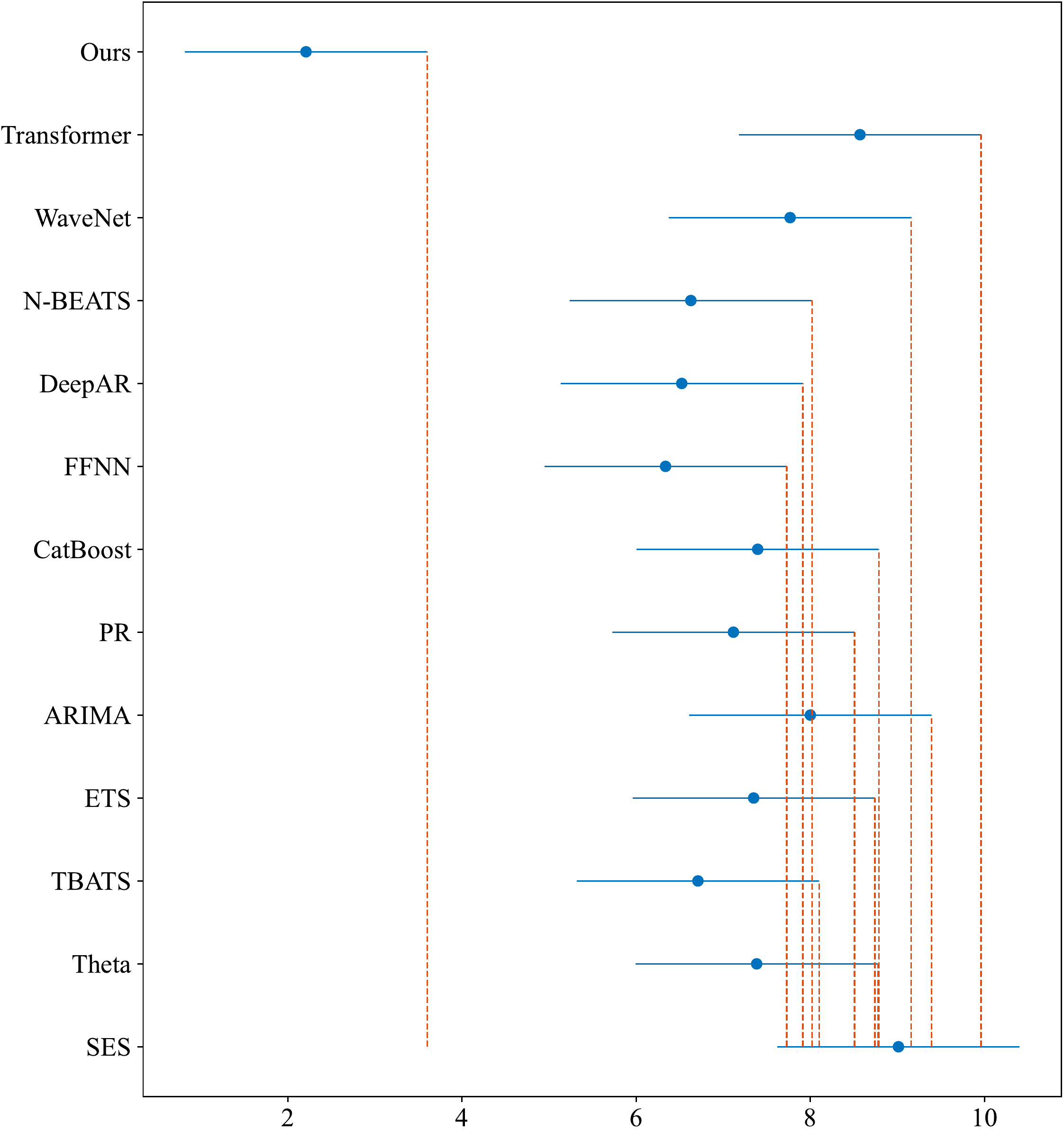}\\
		\end{minipage}%
	}%
	
	\subfigure[\qquad \qquad MAE Comparison Among Models With $^*$]{
		\begin{minipage}[t]{0.495\linewidth}
			\centering
			\includegraphics[width=8.9cm,height=4.65cm]{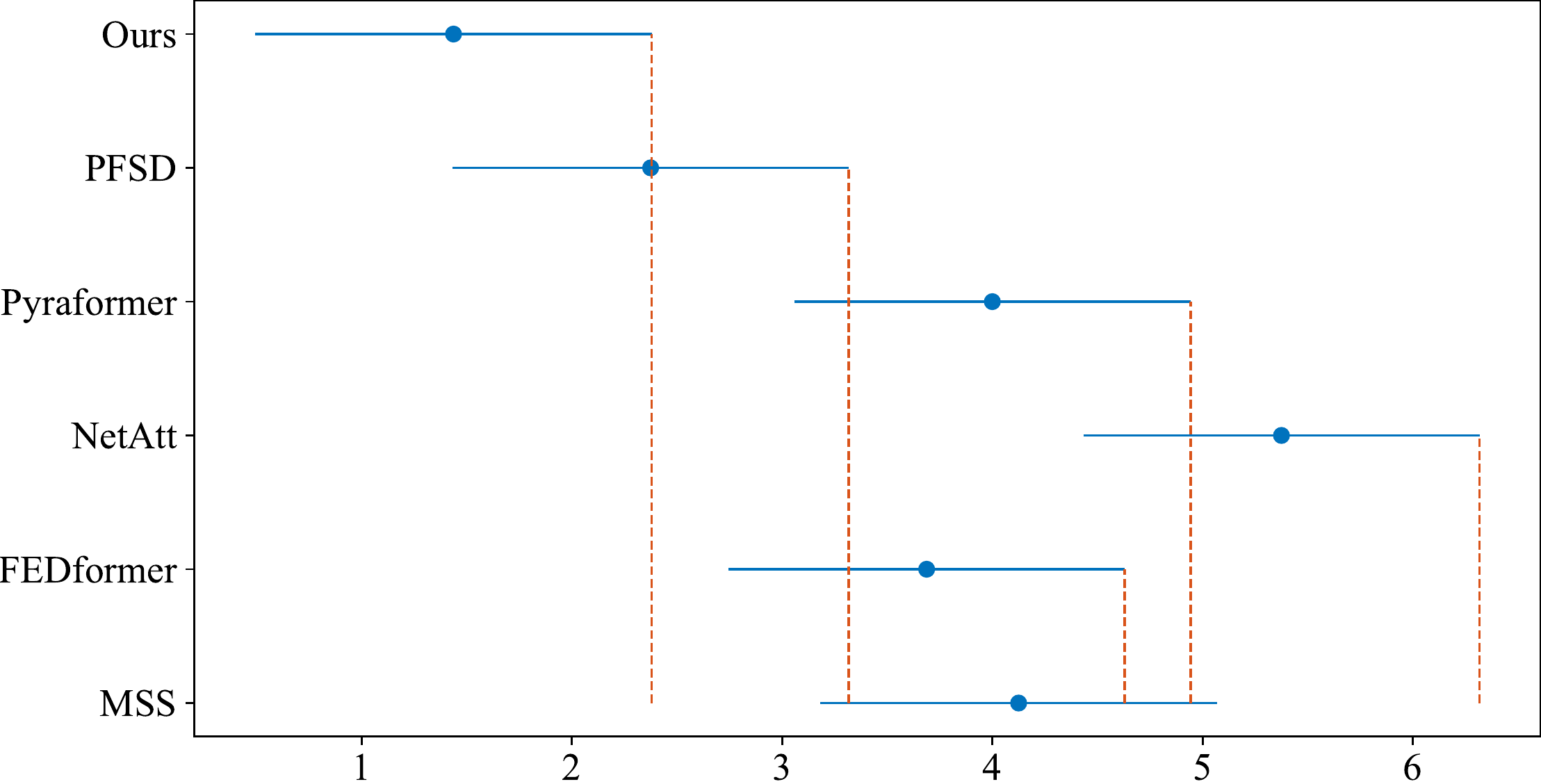}\\
		\end{minipage}%
	}%
	\subfigure[\qquad \qquad RMSE Comparison Among Models With $^*$]{
		\begin{minipage}[t]{0.49\linewidth}
			\centering
			\includegraphics[width=8.9cm,height=4.65cm]{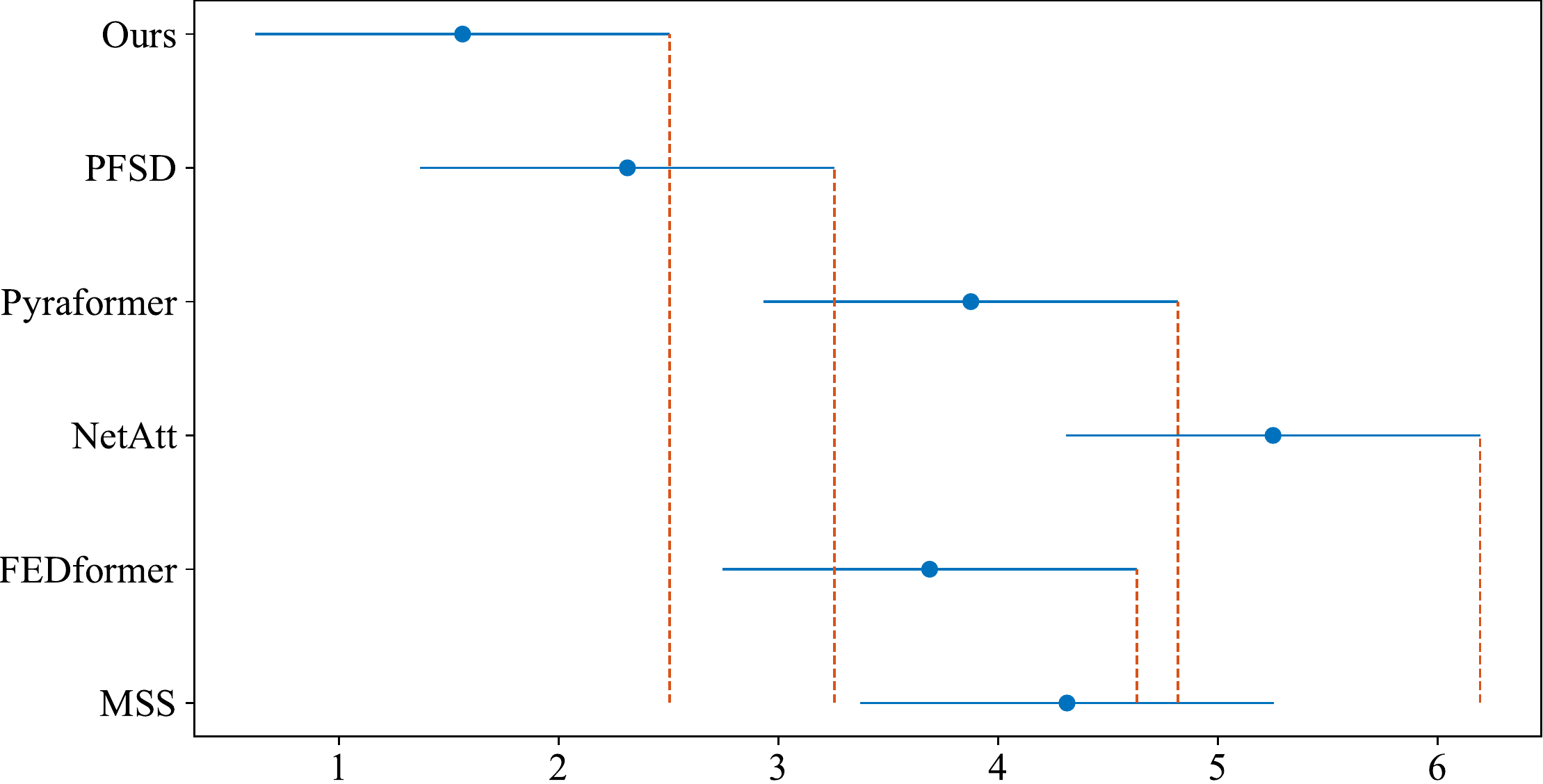}\\
		\end{minipage}%
	}%
	
	\caption{Friedman Test Figure: the Performance Comparison Based MAE and RMSE Among Models From the Perspective of Nemenyi Test.}
	\label{fig6}
\end{figure*}
\subsection{Implementation Details}
The proposed model is realized using the code framework provided by Pytorch 1.13.0. The experimental is conducted with CPU AMD 5900X, GPU NVIDIA RTX 3090, 64GB memory and SSD 2TB. The model is trained for 500 epochs using optimizer NAdam, scheduler ReduceLROnPlateau with factor 0.5, eps 1e-5, threshold 1e-5 and patience 5 and loss function L1Loss without any data augmentation.
\begin{figure*}
	\centering
	\subfigure[Saugeenday Dataset]{
		\begin{minipage}[t]{0.23\linewidth}
			\centering
			\includegraphics[width=1.63in]{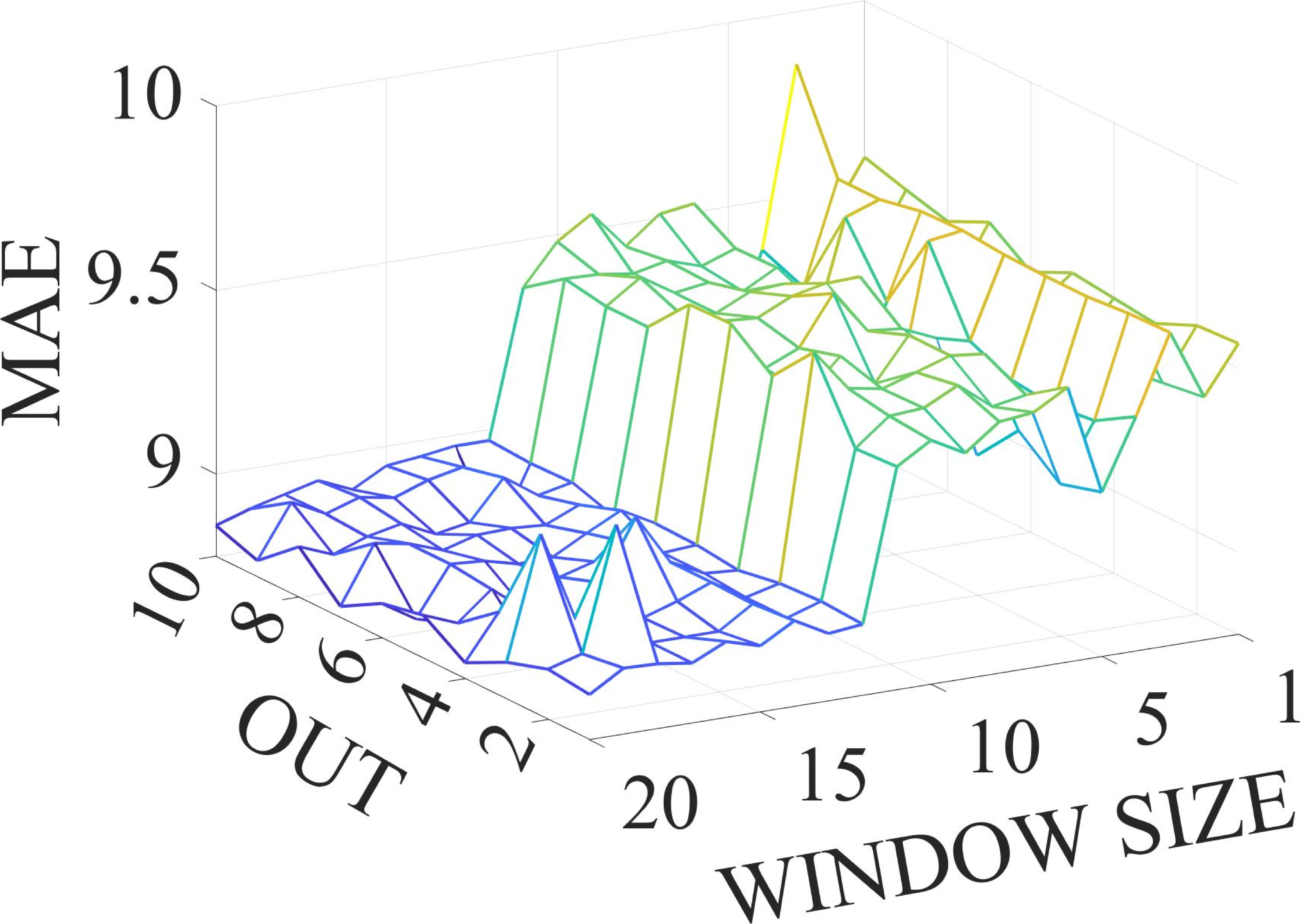}\\
			\vspace{1cm}
			\includegraphics[width=1.64in]{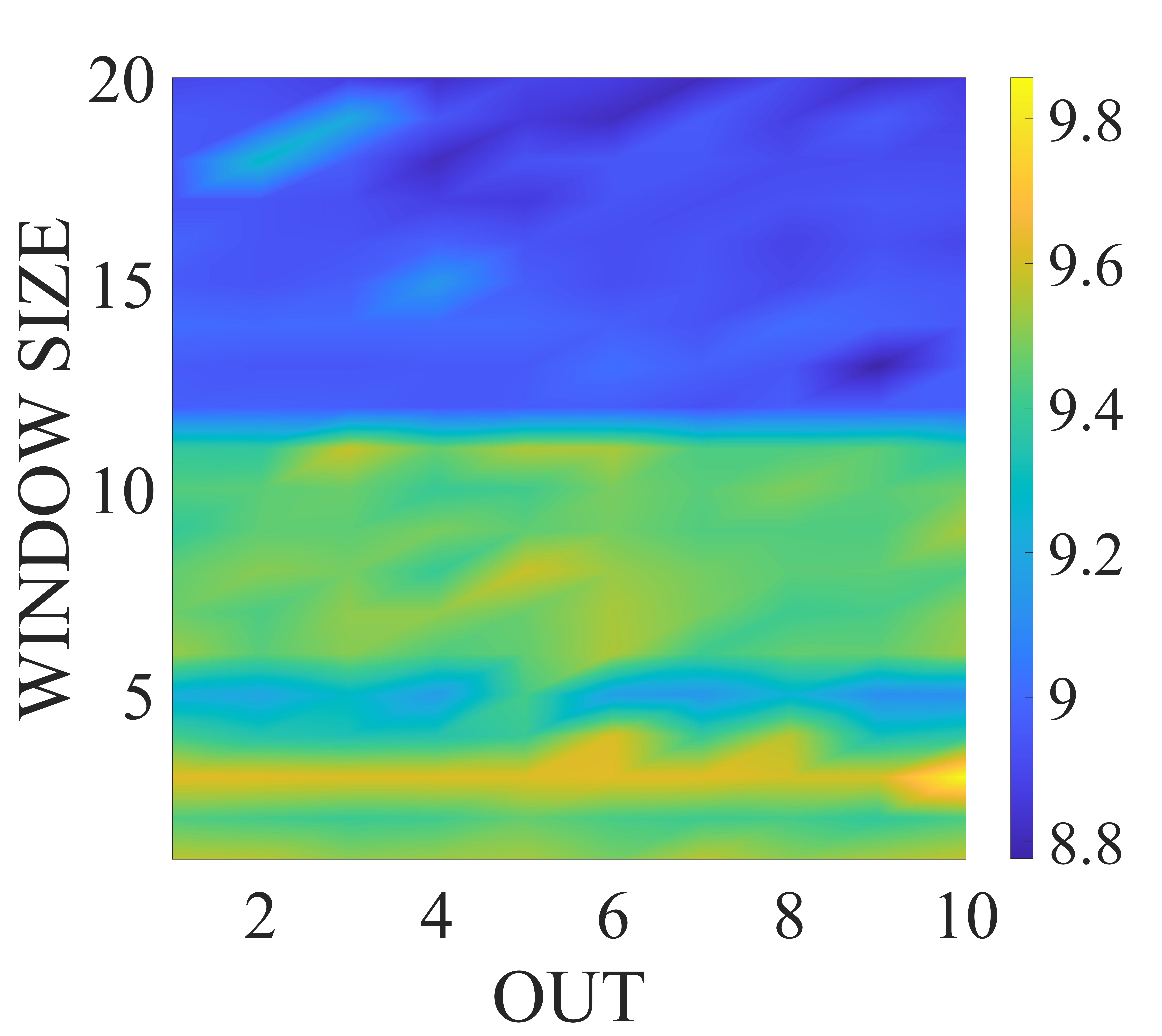}\\
			\vspace{0.3cm}
		\end{minipage}%
	}%
	\subfigure[Tourism Monthly Dataset]{
		\begin{minipage}[t]{0.23\linewidth}
			\centering
			\includegraphics[width=1.63in]{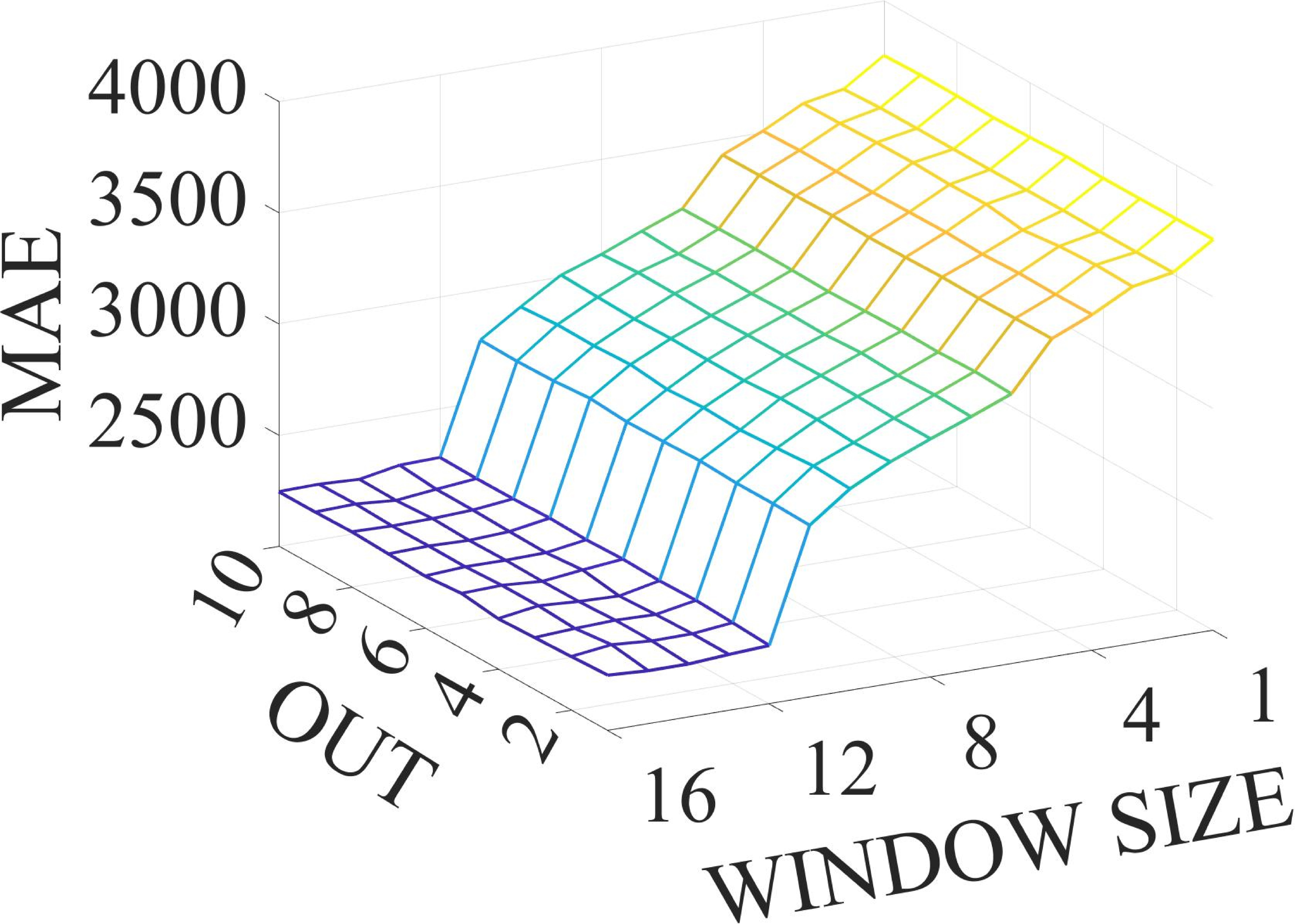}\\
			\vspace{1.07cm}
			\includegraphics[width=1.67in,height=1.4in]{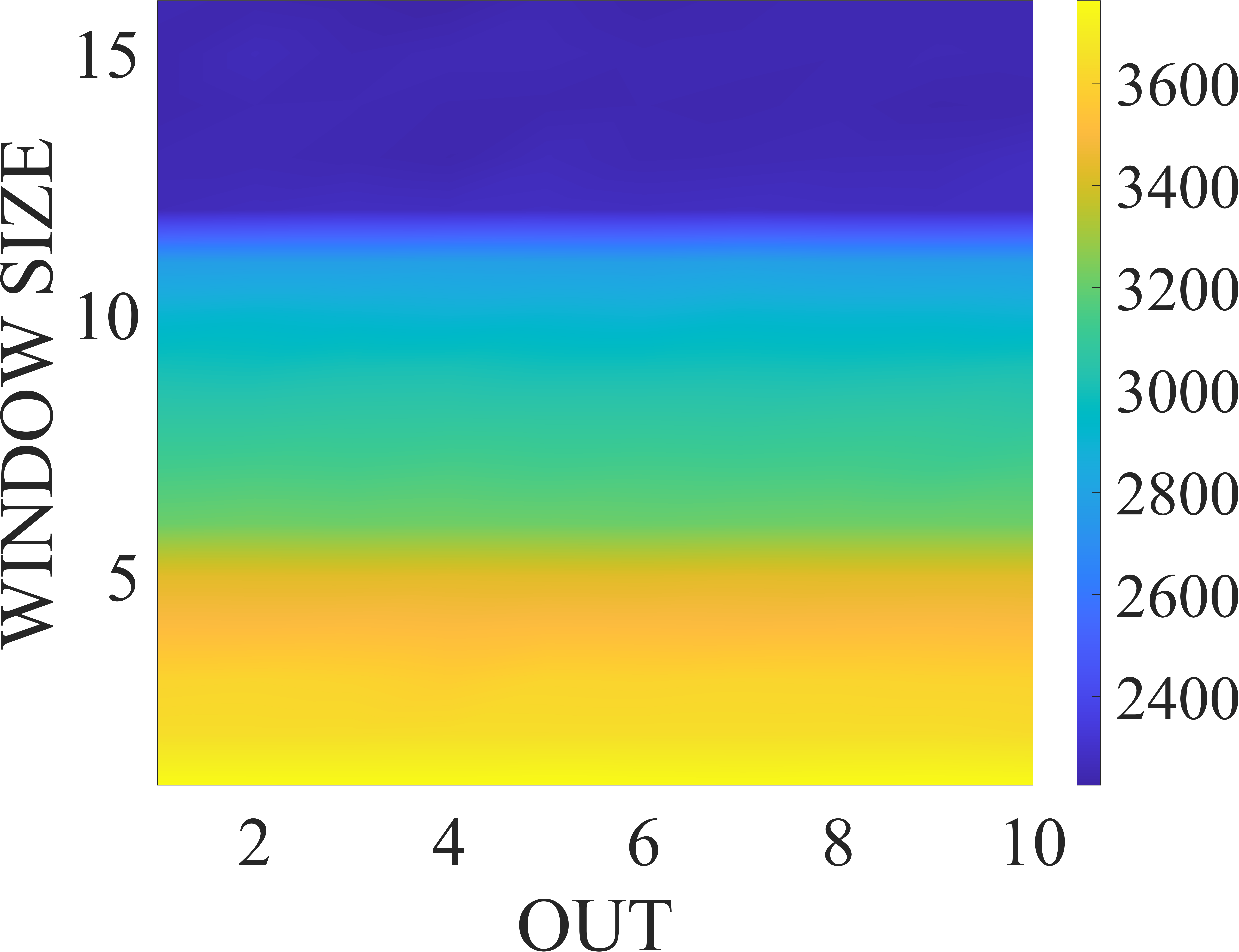}\\
			\vspace{0.3cm}
		\end{minipage}%
	}%
	\subfigure[Tourism Quarterly Dataset]{
		\begin{minipage}[t]{0.23\linewidth}
			\centering
			\includegraphics[width=1.55in]{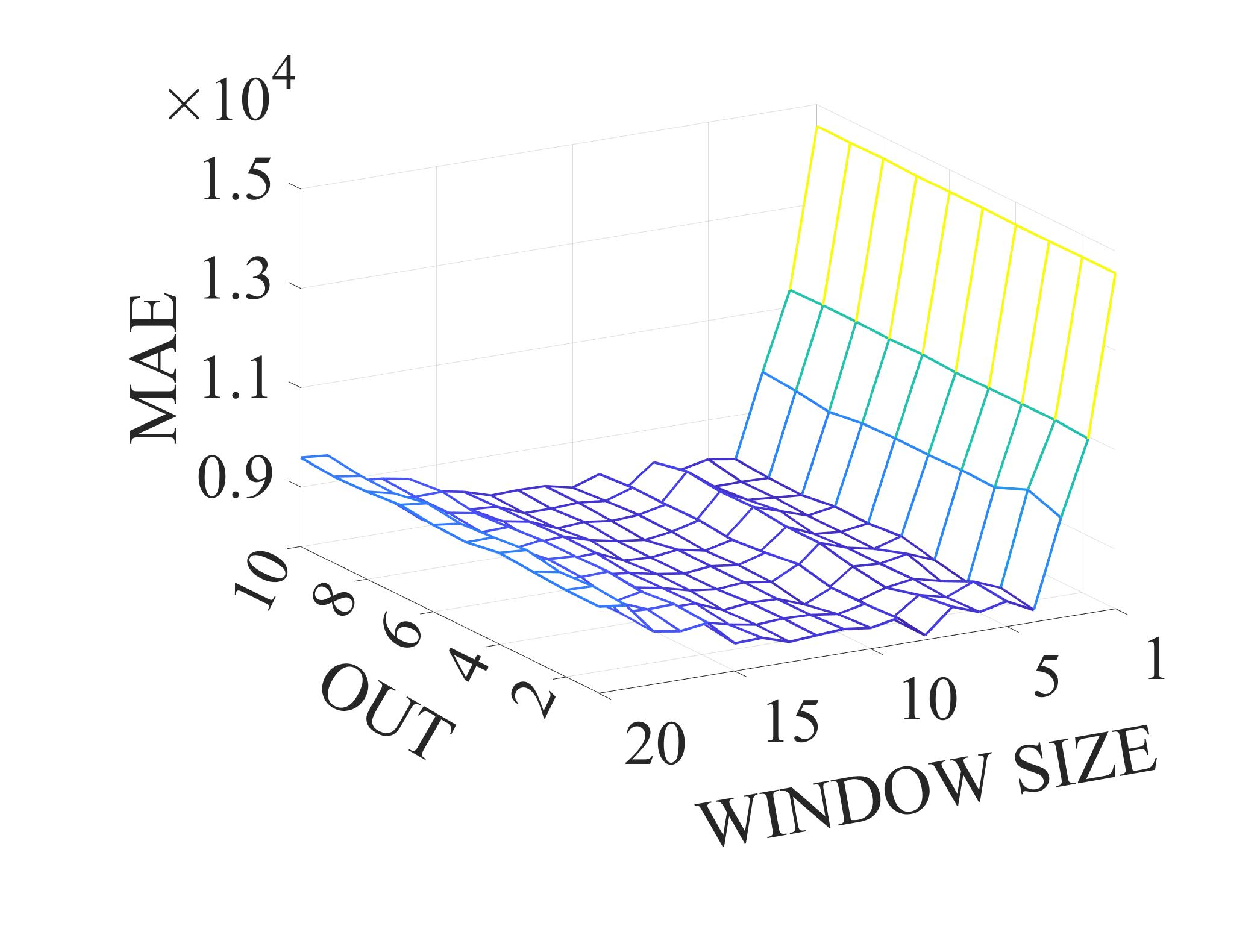}\\
			\vspace{0.79cm}
			\includegraphics[width=1.64in,height=1.51in]{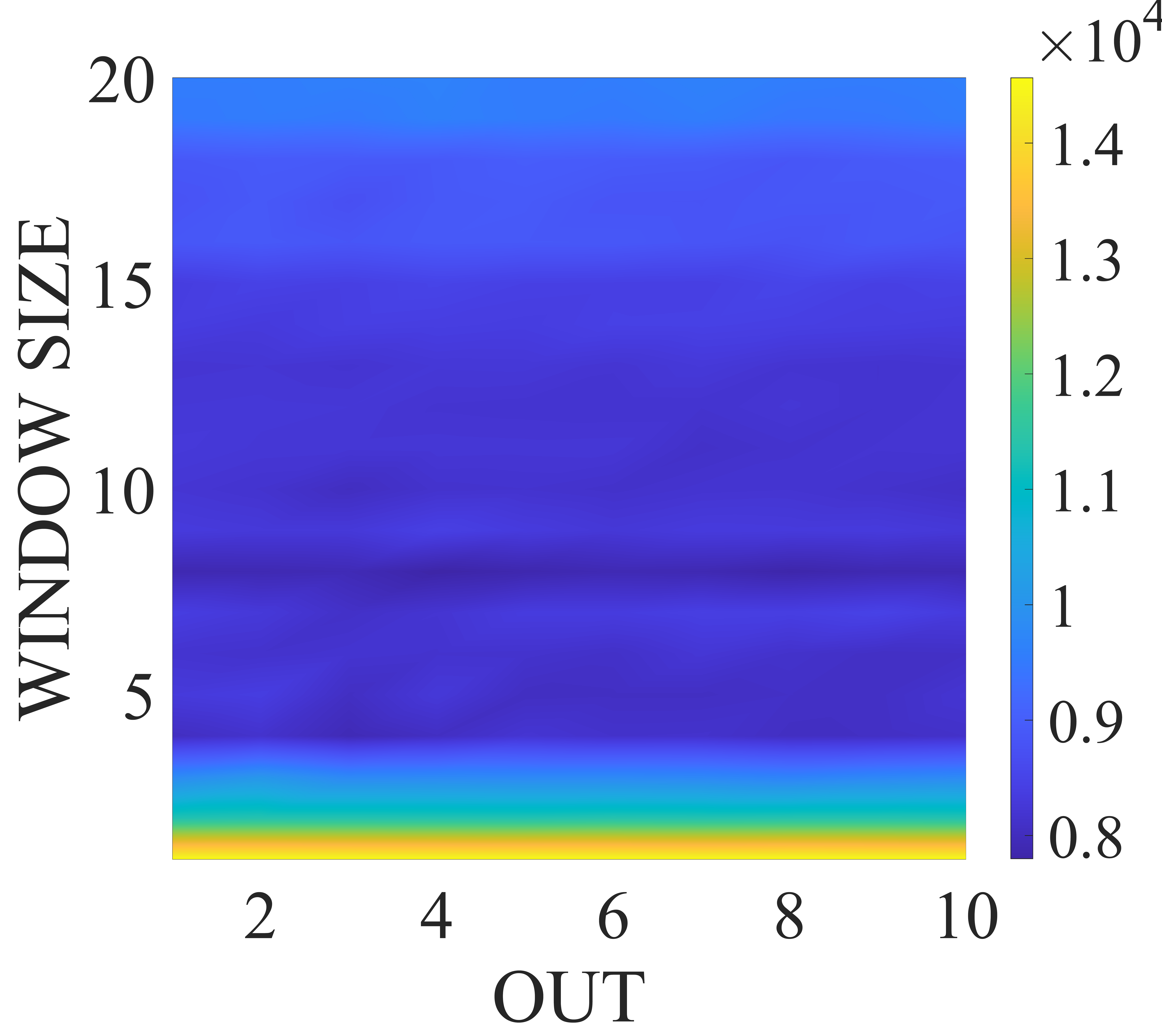}\\
			\vspace{0.3cm}
		\end{minipage}%
	}%
	\subfigure[US Births Dataset]{
		\begin{minipage}[t]{0.23\linewidth}
			\centering
			\includegraphics[width=1.63in]{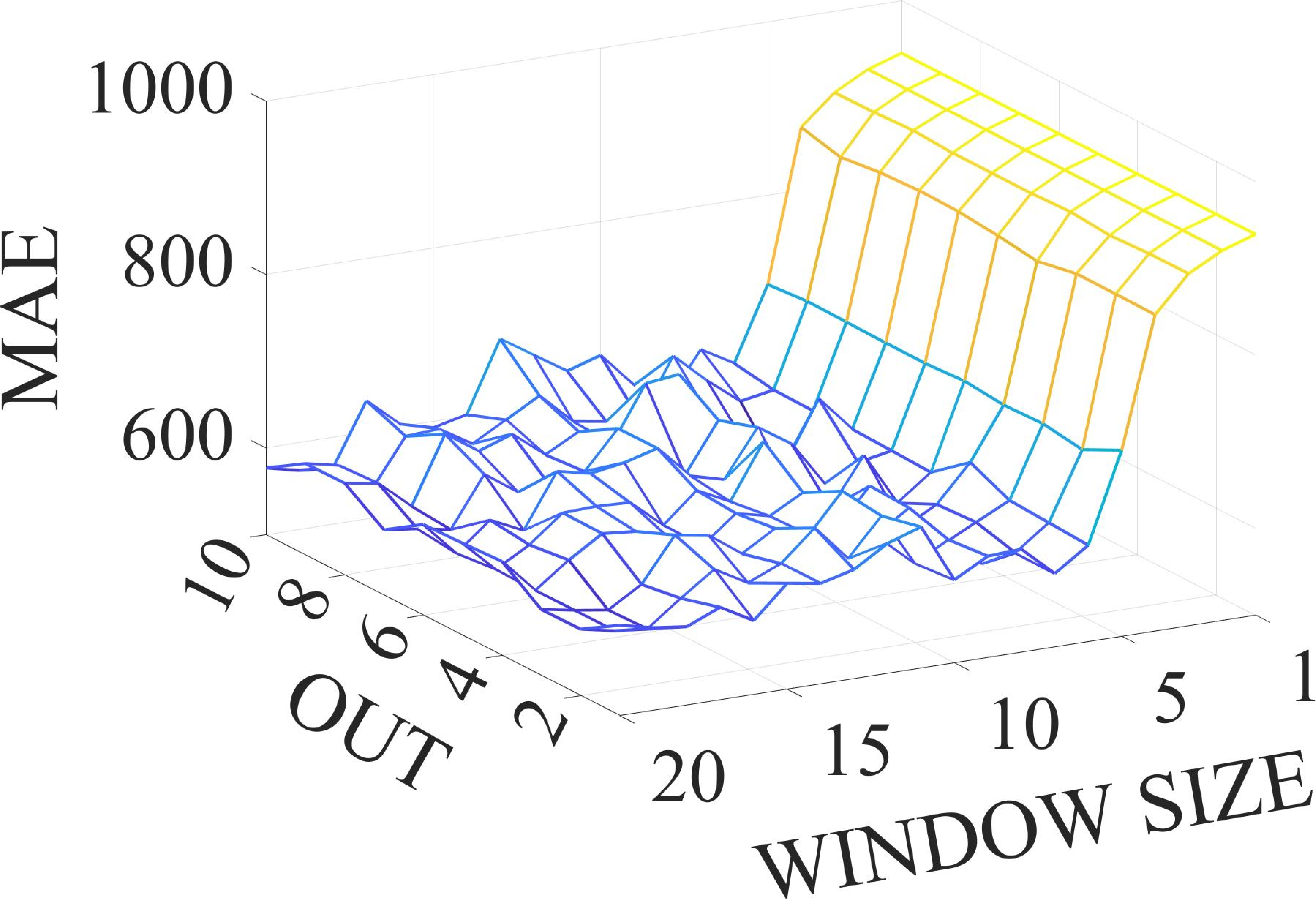}\\
			\vspace{0.98cm}
			\includegraphics[width=1.61in,height=1.44in]{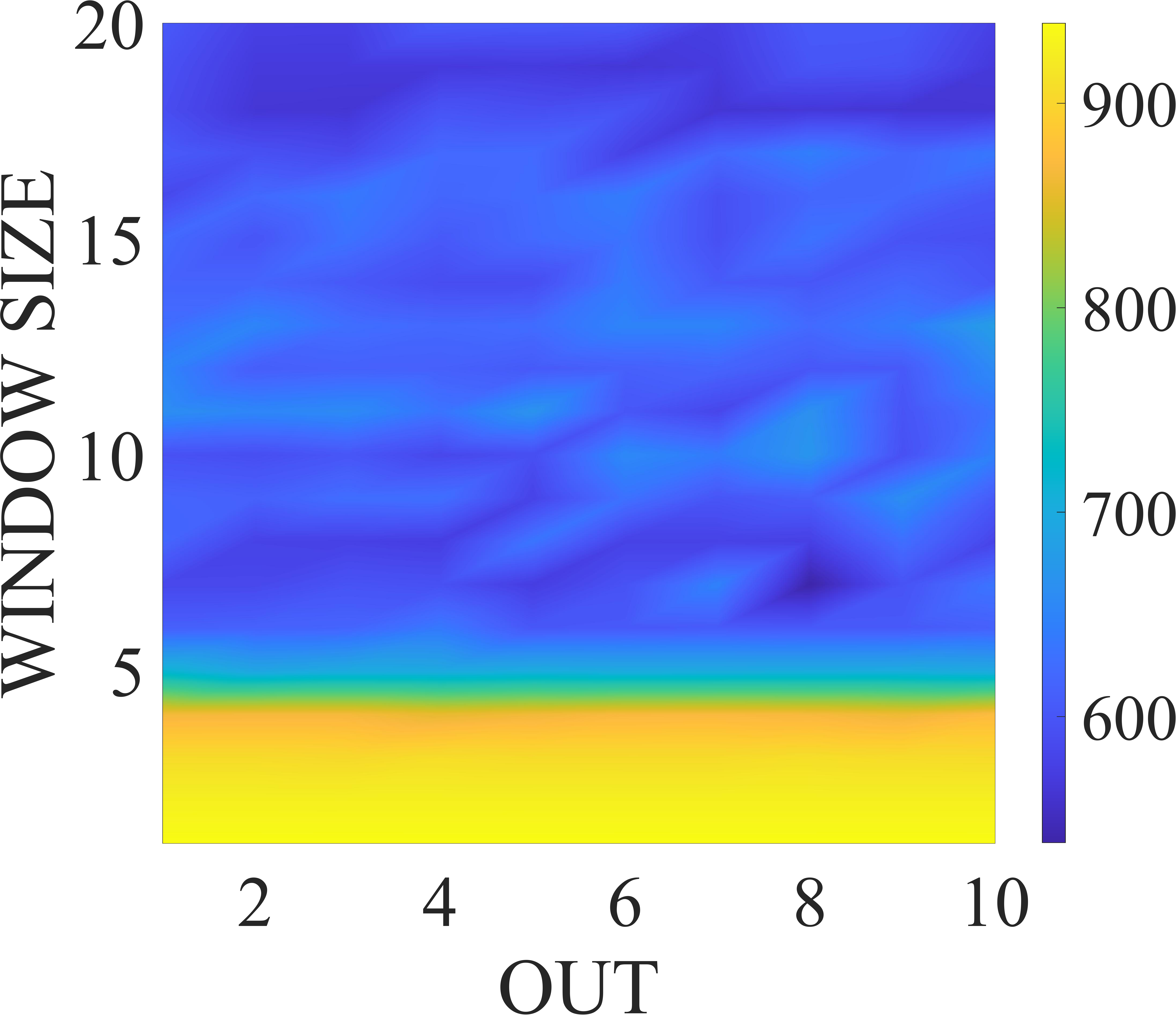}\\
			\vspace{0.3cm}
		\end{minipage}%
	}%
	
	\caption{MAE Variations When Parameter OUT and Window Size Vary on Univariate Datasets}
	\label{fig1}
\end{figure*}

\begin{figure*}
	\centering
	\subfigure[Electricity Weekly Dataset]{
		\begin{minipage}[t]{0.23\linewidth}
			\centering
			\includegraphics[width=1.68in,height=1.35in]{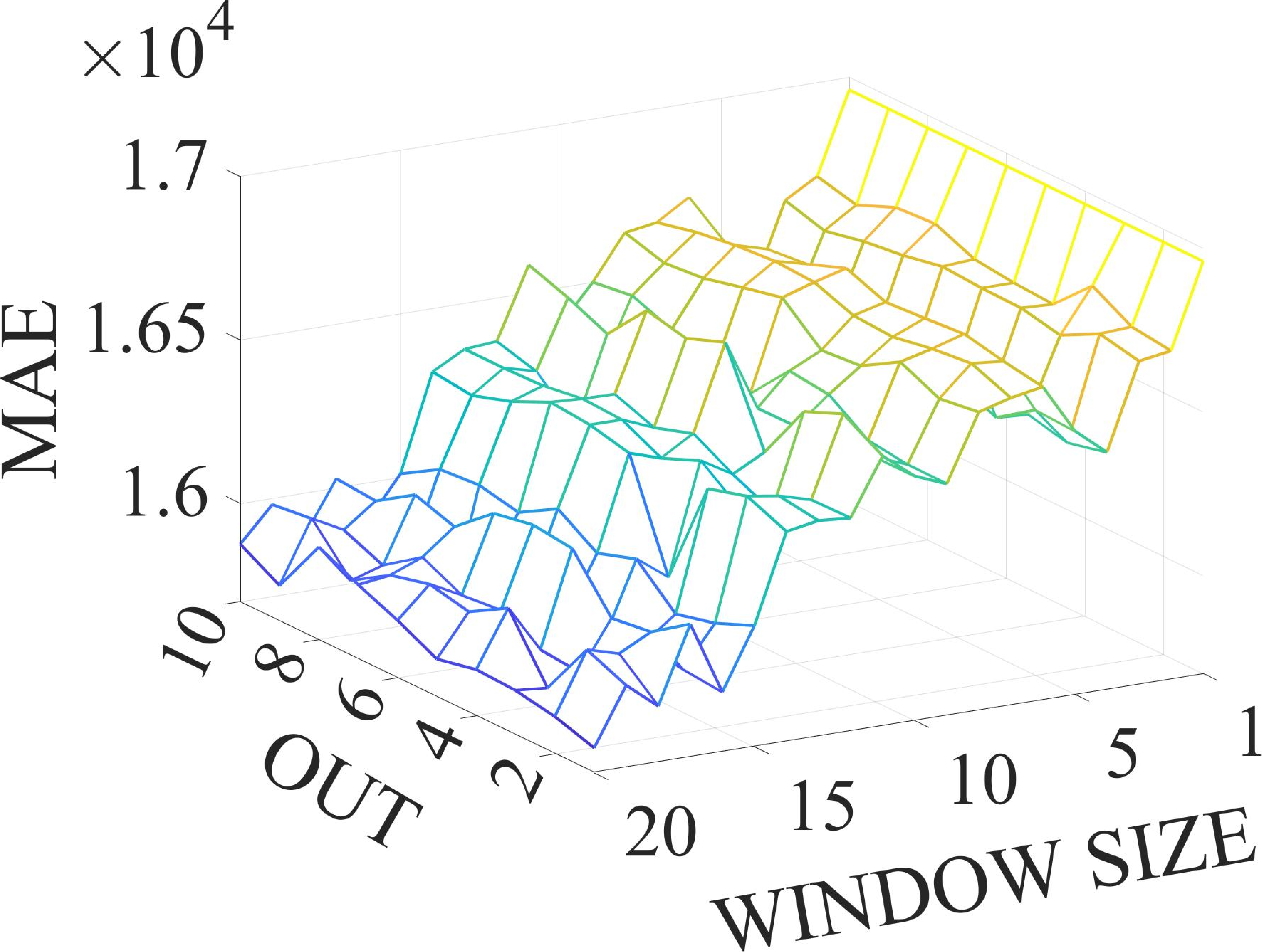}\\
			\vspace{0.8cm}
			\includegraphics[width=1.68in,height = 1.53in]{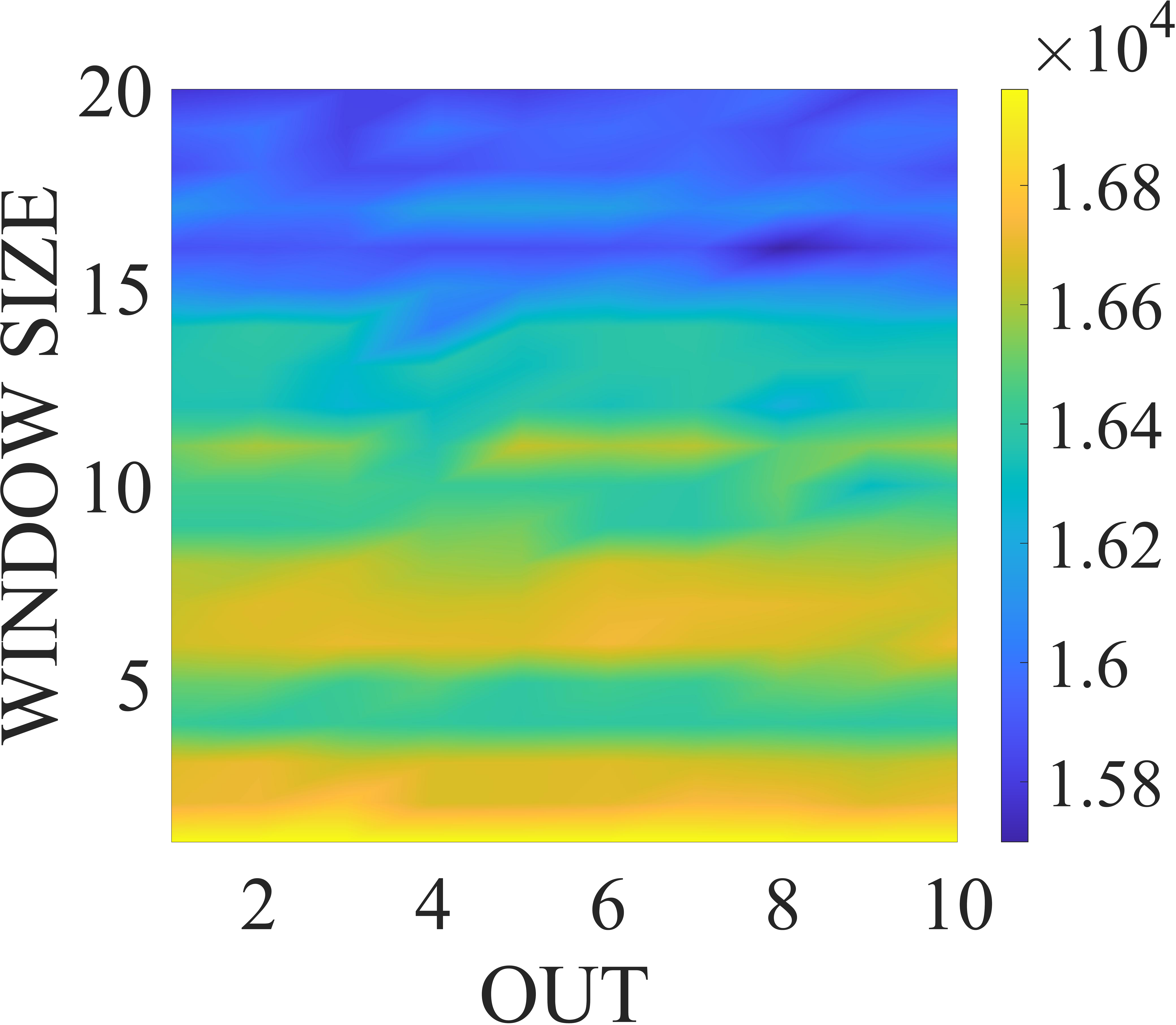}\\
			\vspace{0.3cm}
		\end{minipage}%
	}%
	\subfigure[Hospital Dataset]{
		\begin{minipage}[t]{0.23\linewidth}
			\centering
			\includegraphics[width=1.63in]{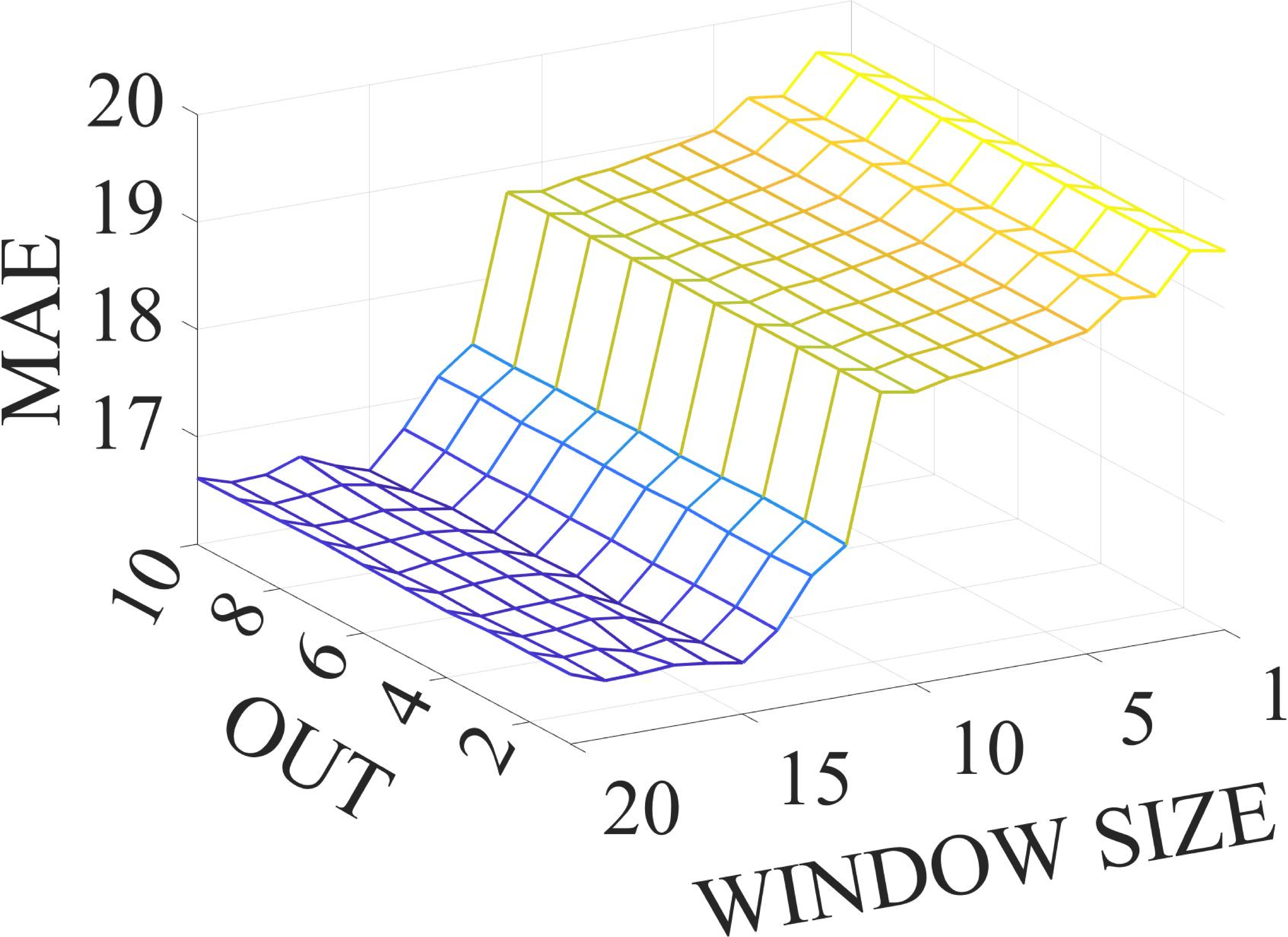}\\
			\vspace{1.04cm}
			\includegraphics[width=1.64in]{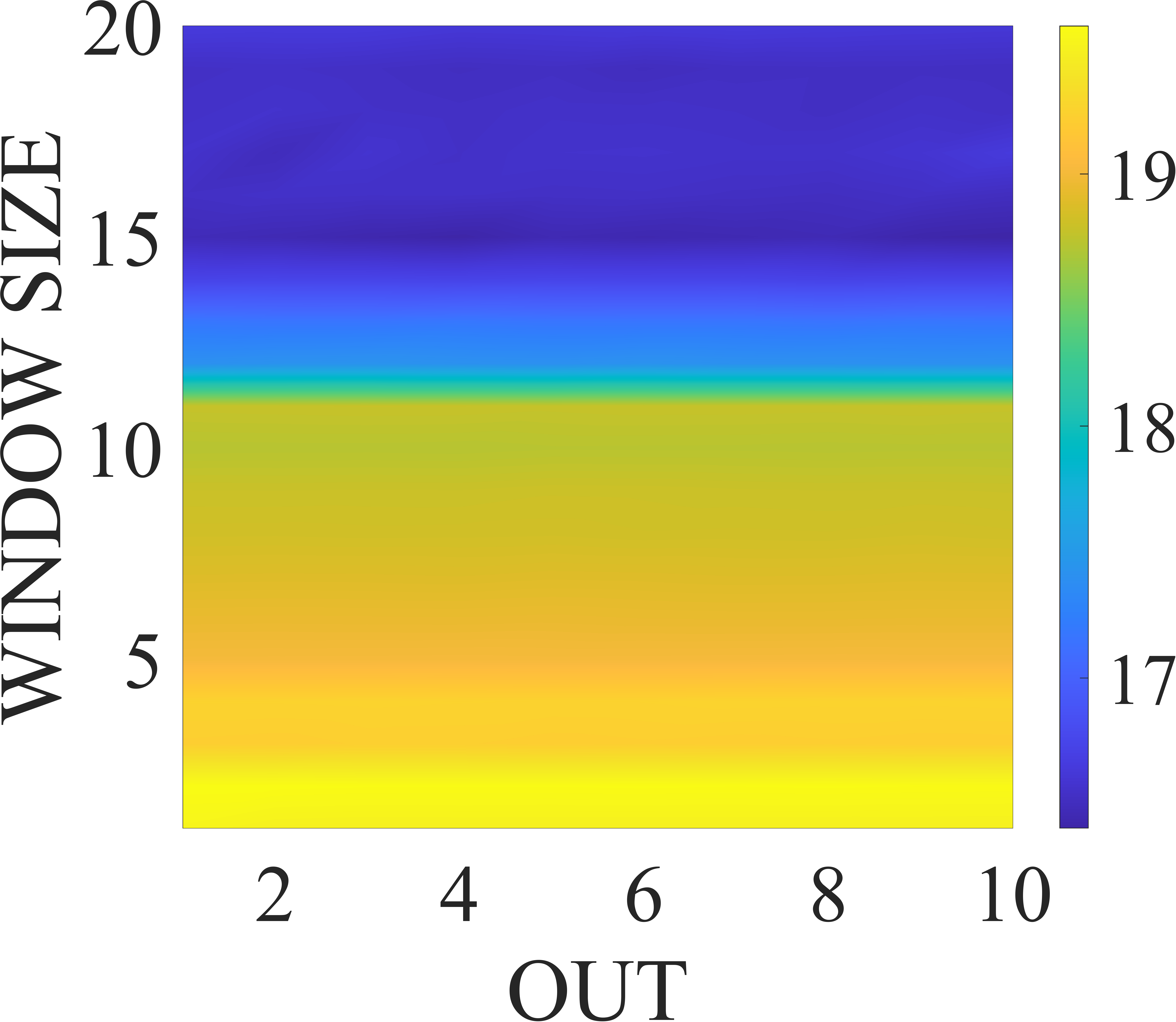}\\
			\vspace{0.3cm}
		\end{minipage}%
	}%
	\subfigure[Traffic Weekly Dataset]{
		\begin{minipage}[t]{0.24\linewidth}
			\centering
			\includegraphics[width=1.68in]{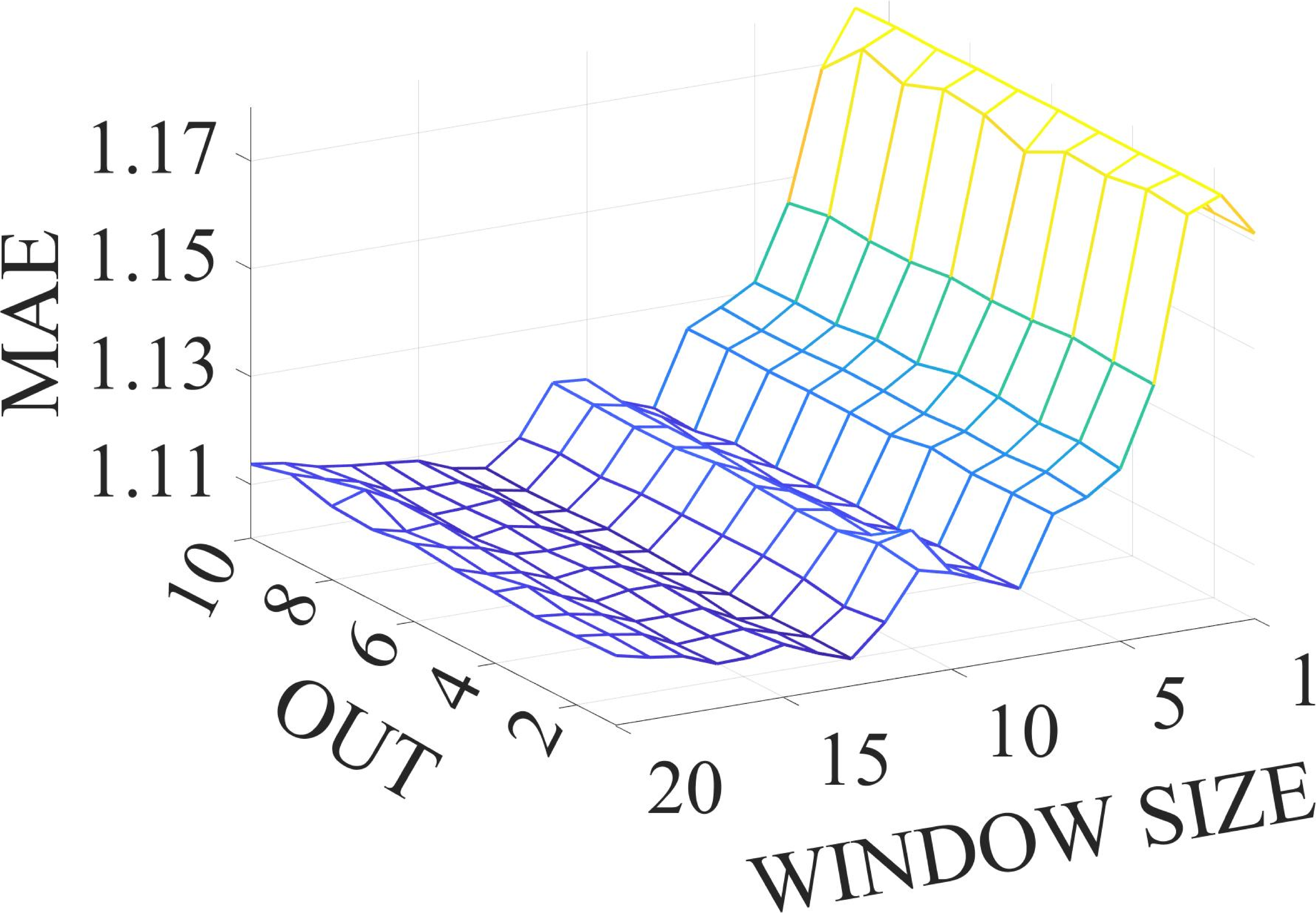}\\
			\vspace{1.03cm}
			\includegraphics[width=1.68in,height=1.43in]{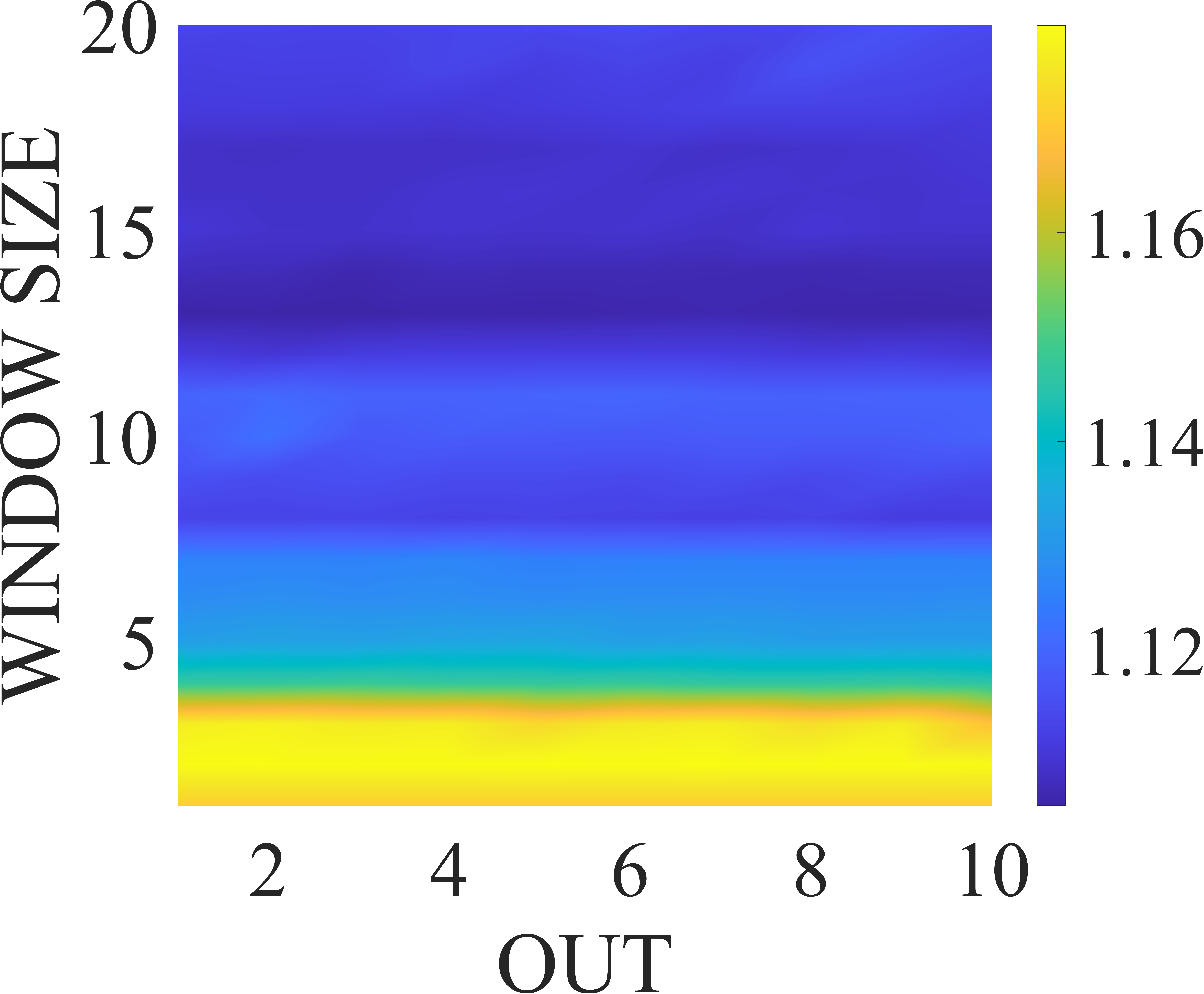}\\
			\vspace{0.3cm}
		\end{minipage}%
	}%
	\subfigure[Solar Weekly Dataset]{
		\begin{minipage}[t]{0.23\linewidth}
			\centering
			\includegraphics[width=1.69in,height=1.25in]{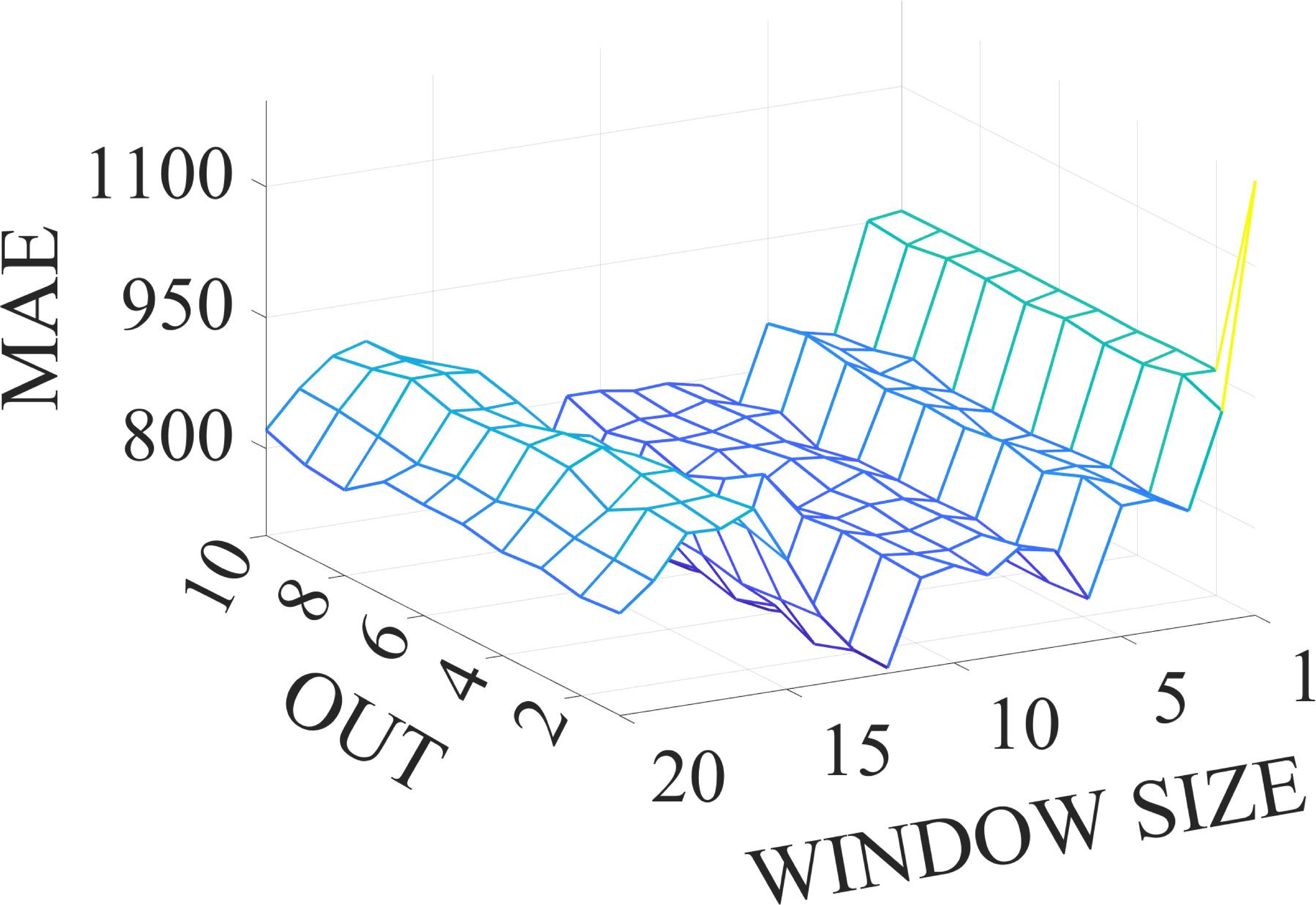}\\
			\vspace{1.04cm}
			\includegraphics[width=1.69in,height=1.41in]{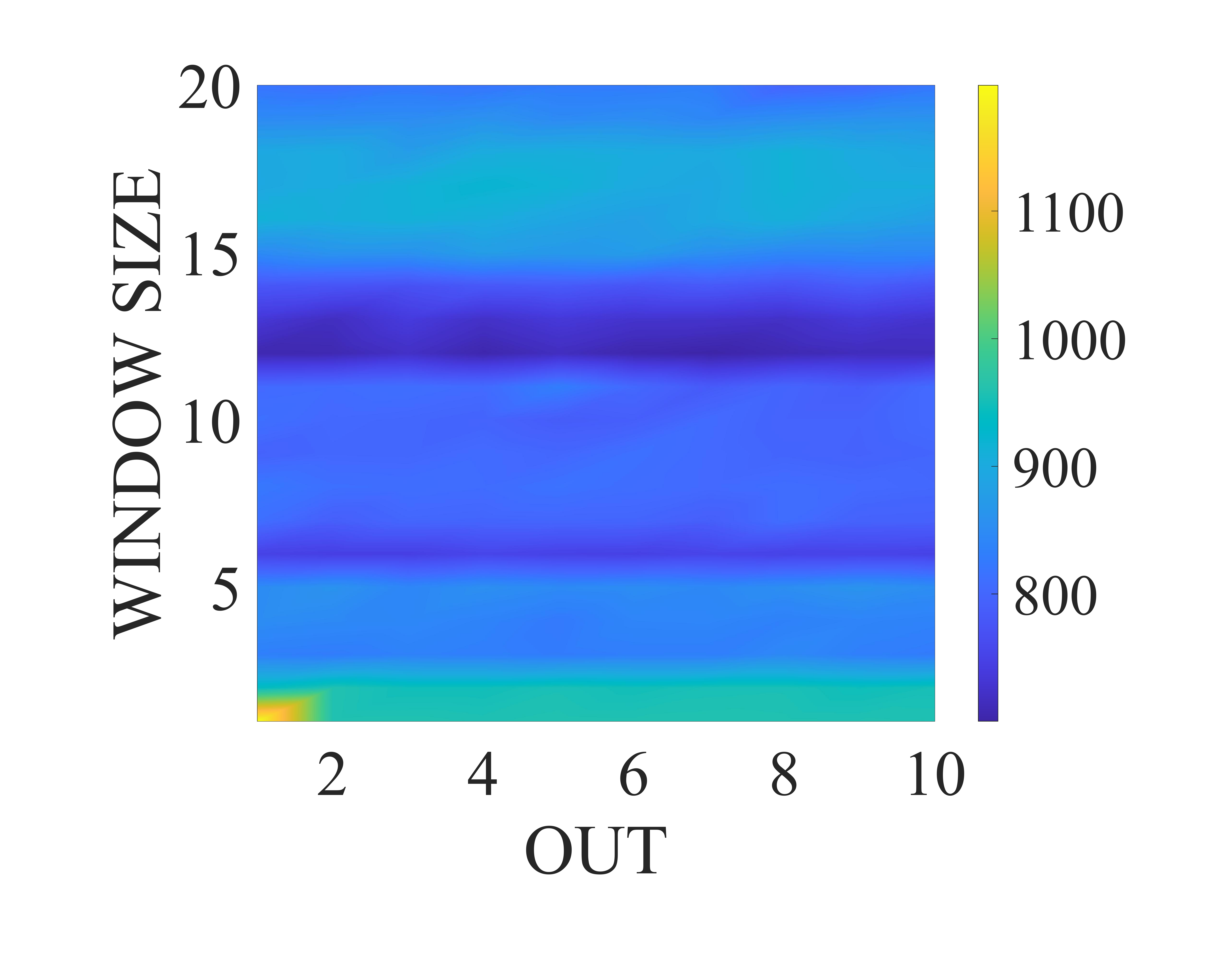}\\
			\vspace{0.3cm}
		\end{minipage}%
	}%
	
	\caption{MAE Variations When Parameter OUT and Window Size Vary on Multivariate Datasets}
	\label{figg}
\end{figure*}

\begin{figure*}
	\centering
	\subfigure[M1 Monthly Dataset]{
		\begin{minipage}[t]{0.23\linewidth}
			\centering
			\includegraphics[width=1.63in]{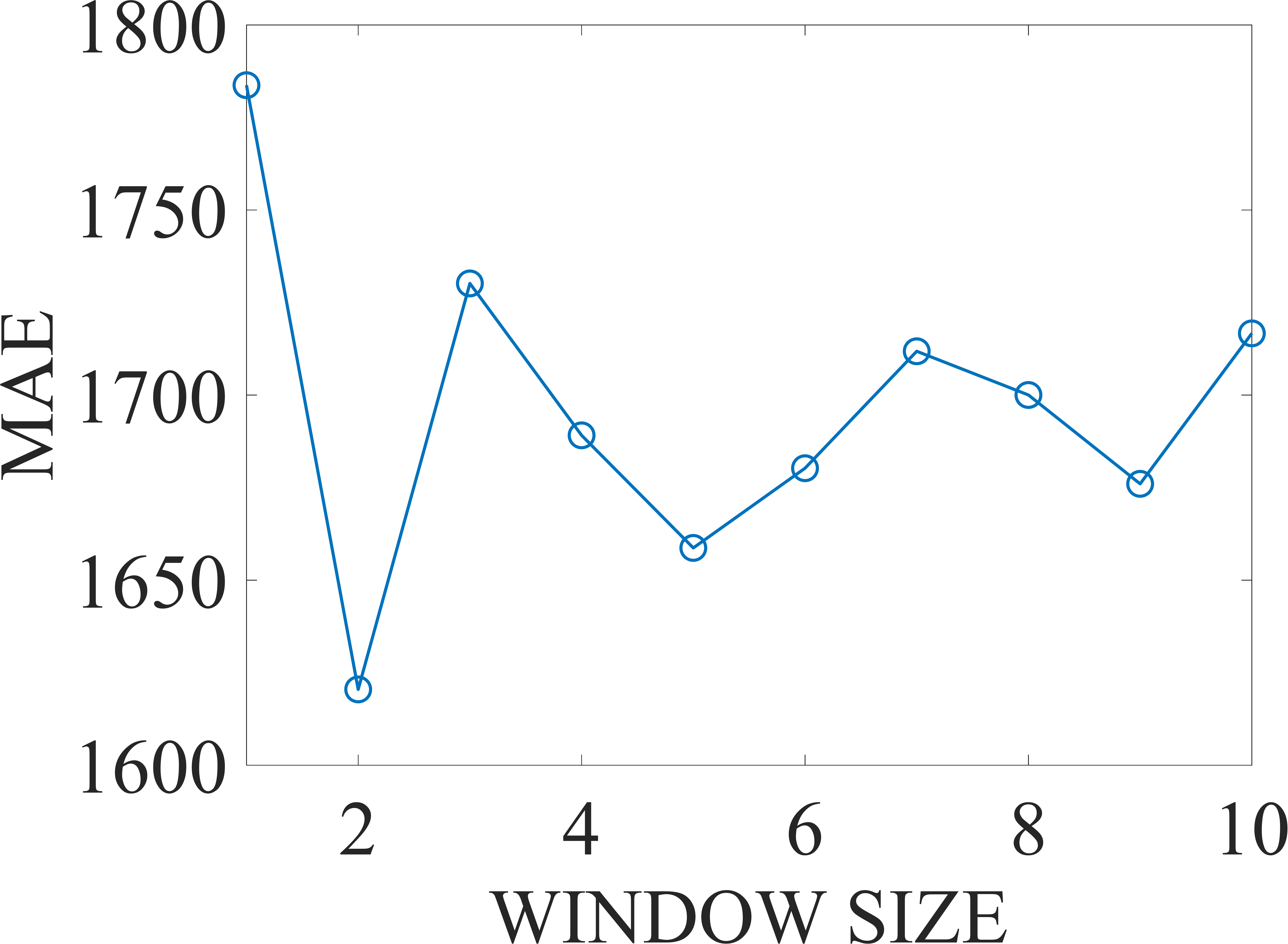}\\
		\end{minipage}%
	}%
	\subfigure[M1 Quarterly Dataset]{
		\begin{minipage}[t]{0.23\linewidth}
			\centering
			\includegraphics[width=1.63in]{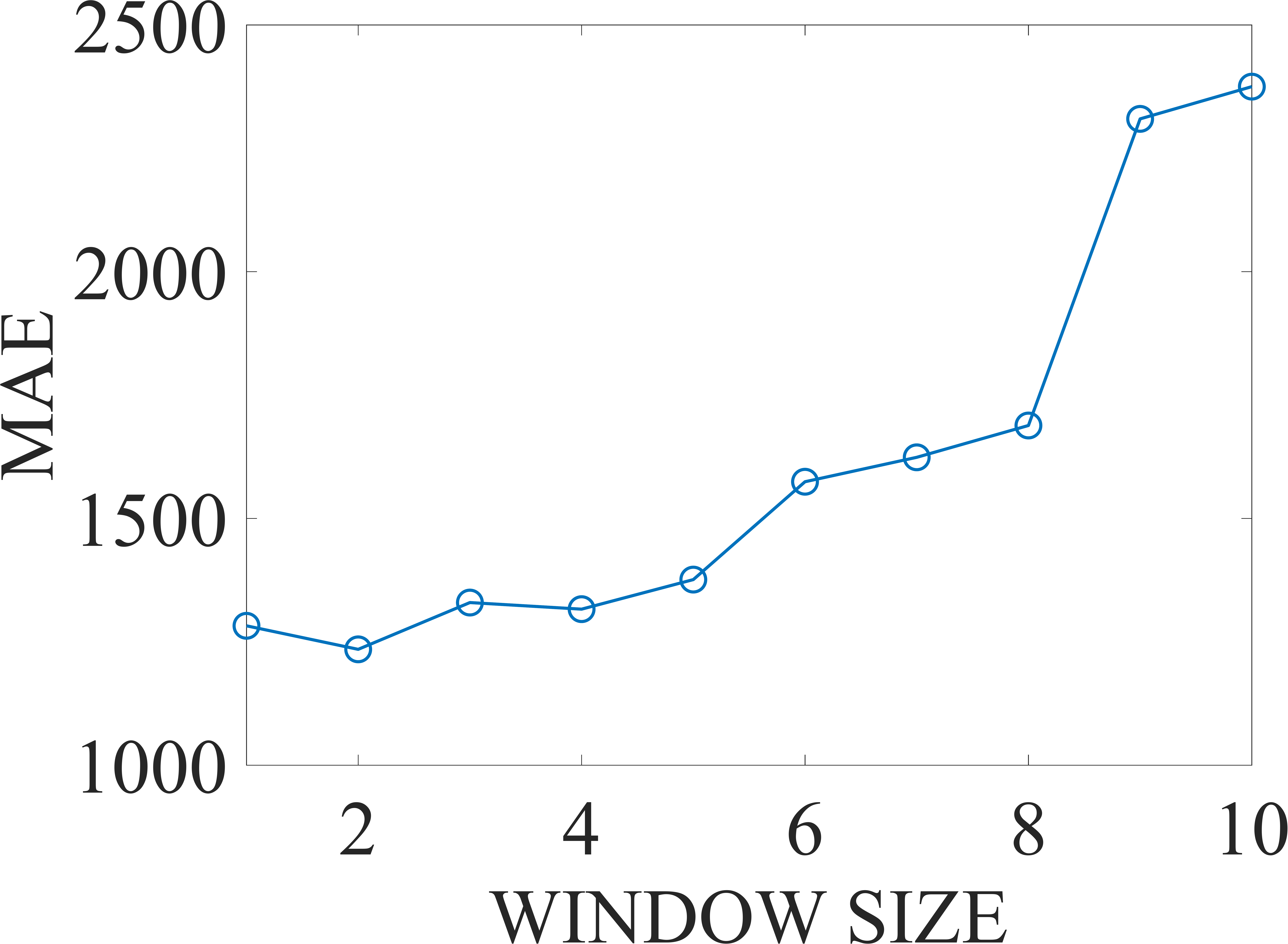}\\
		\end{minipage}%
	}%
	\subfigure[M1 Yearly Dataset]{
		\begin{minipage}[t]{0.23\linewidth}
			\centering
			\includegraphics[width=1.63in]{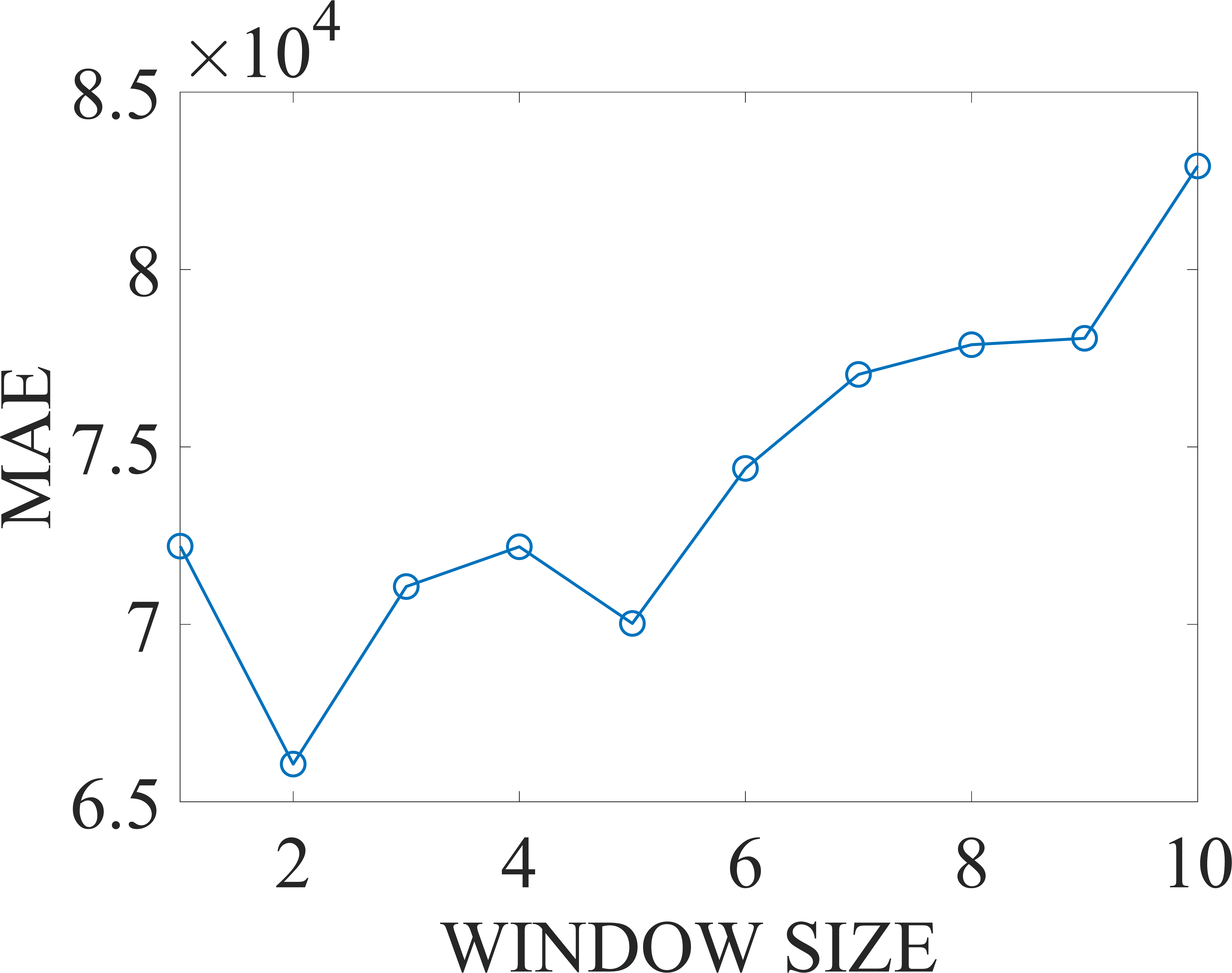}\\
		\end{minipage}%
	}%
	\subfigure[M3 Monthly Dataset]{
		\begin{minipage}[t]{0.23\linewidth}
			\centering
			\includegraphics[width=1.63in]{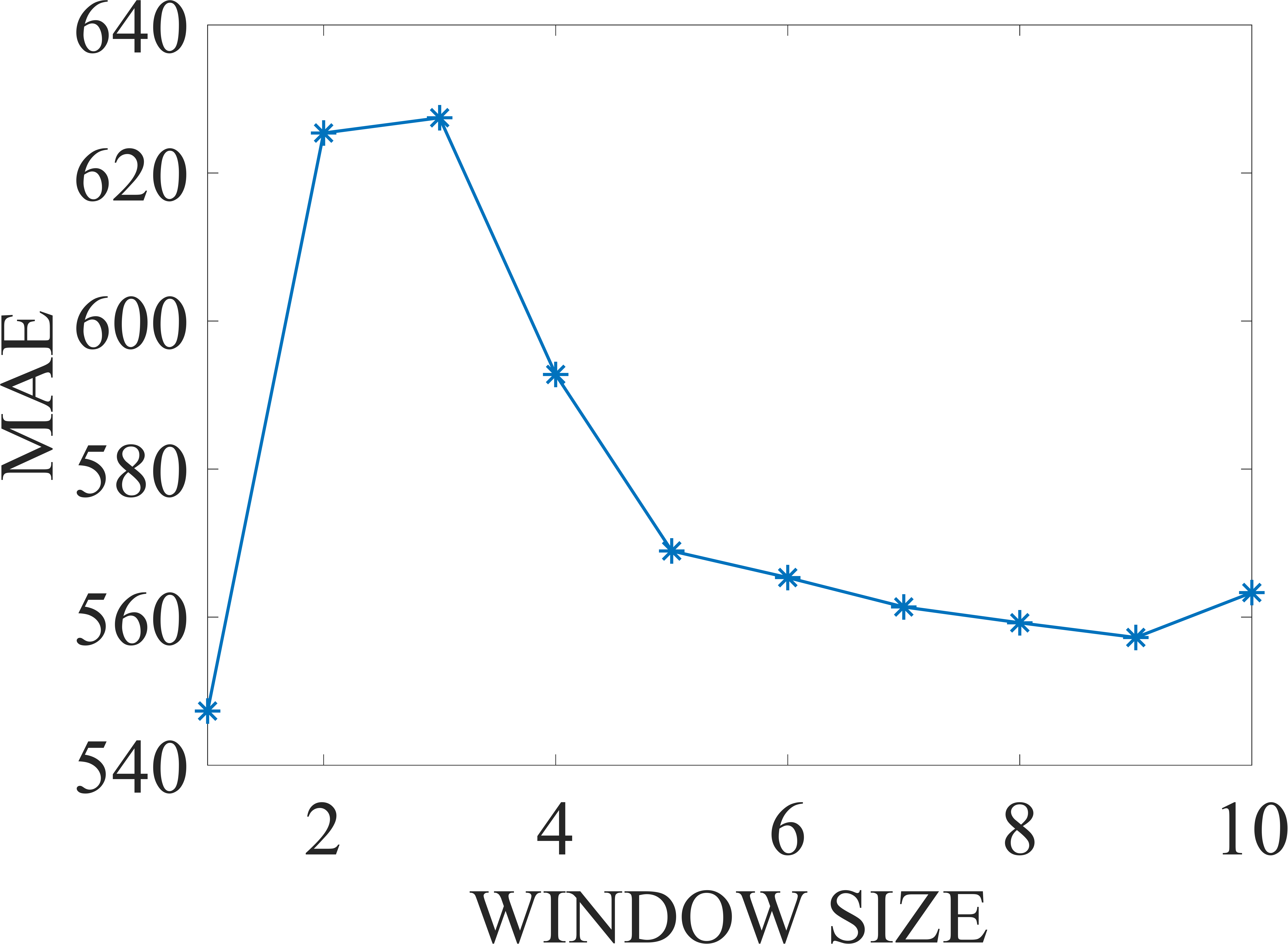}\\
		\end{minipage}%
	}%
	
	\subfigure[M3 Other Dataset]{
		\begin{minipage}[t]{0.23\linewidth}
			\centering
			\includegraphics[width=1.63in]{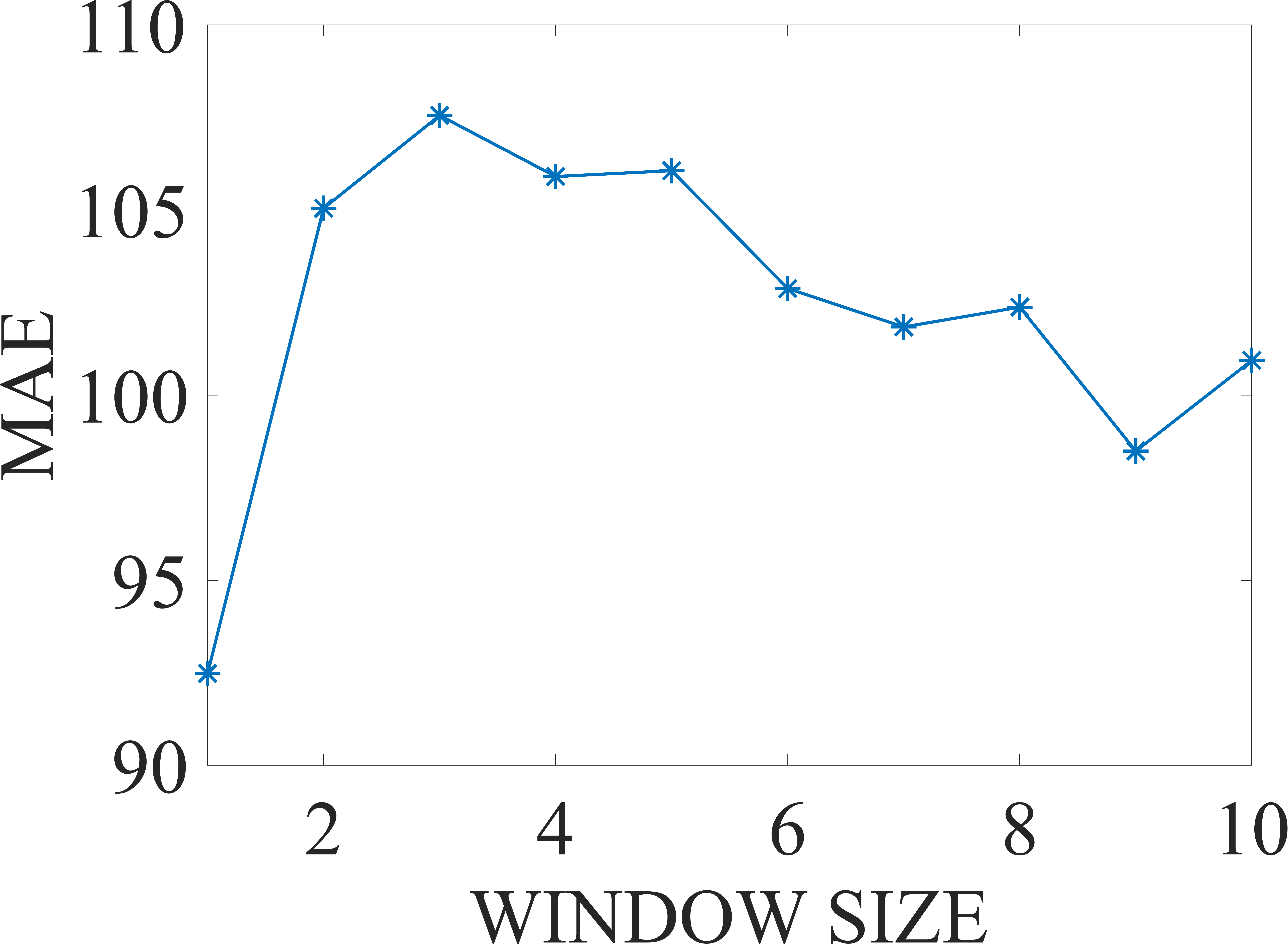}\\
		\end{minipage}%
	}%
	\subfigure[M3 Quarterly Dataset]{
		\begin{minipage}[t]{0.23\linewidth}
			\centering
			\includegraphics[width=1.63in]{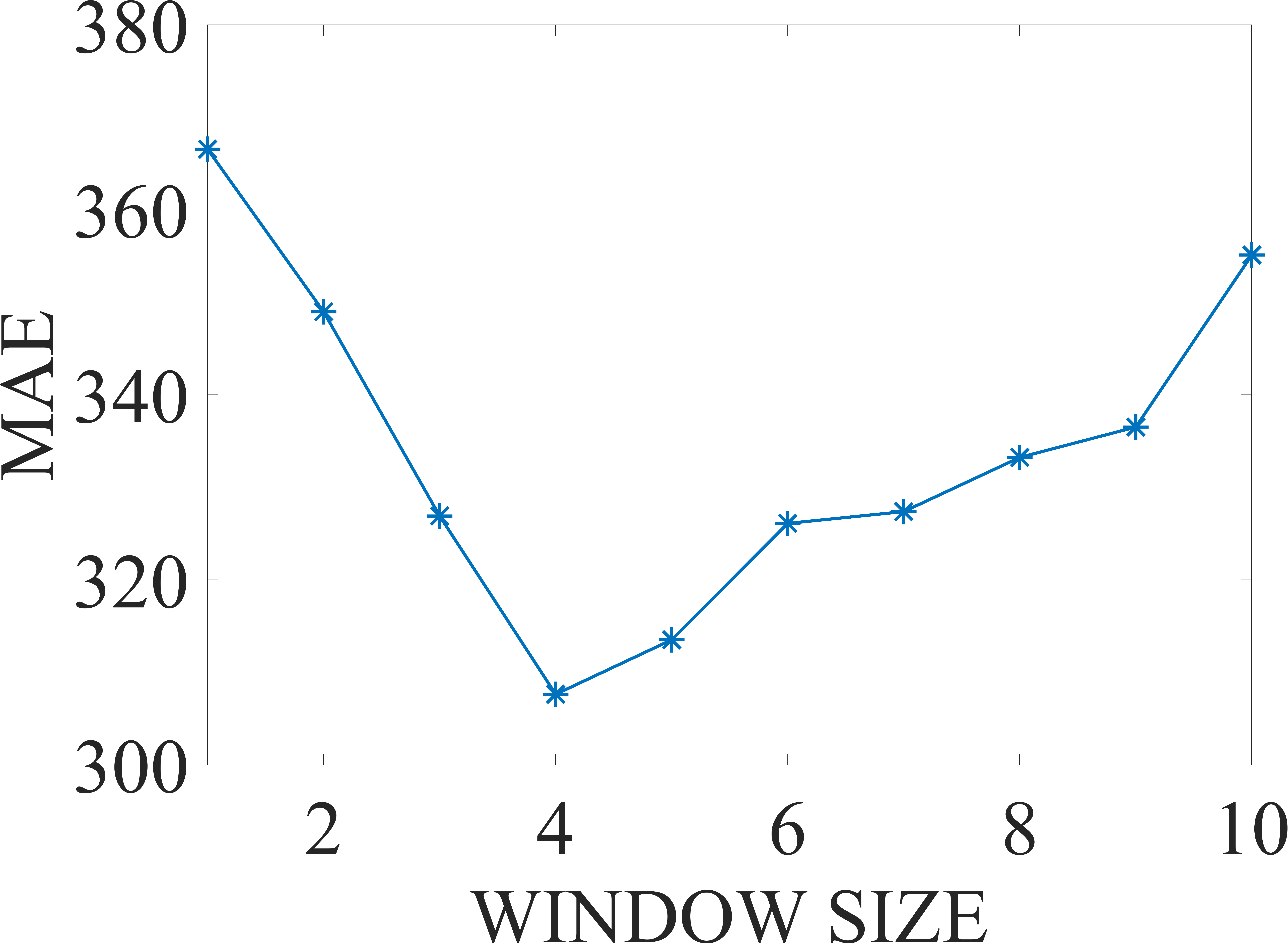}\\
		\end{minipage}%
	}%
	\subfigure[M3 Yearly Dataset]{
		\begin{minipage}[t]{0.23\linewidth}
			\centering
			\includegraphics[width=1.63in]{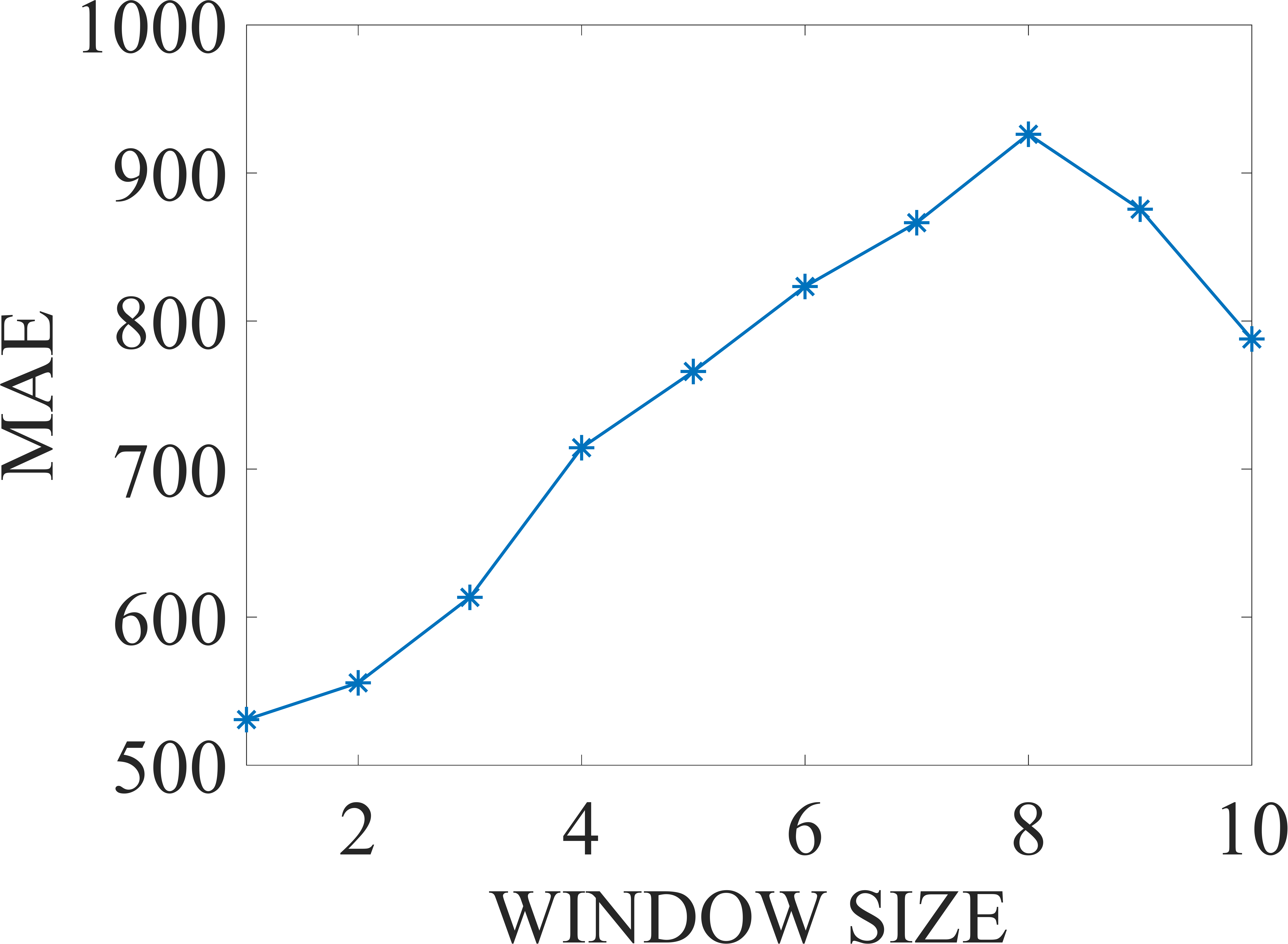}\\
		\end{minipage}%
	}%
	\subfigure[M4 Daily Dataset]{
		\begin{minipage}[t]{0.23\linewidth}
			\centering
			\includegraphics[width=1.63in]{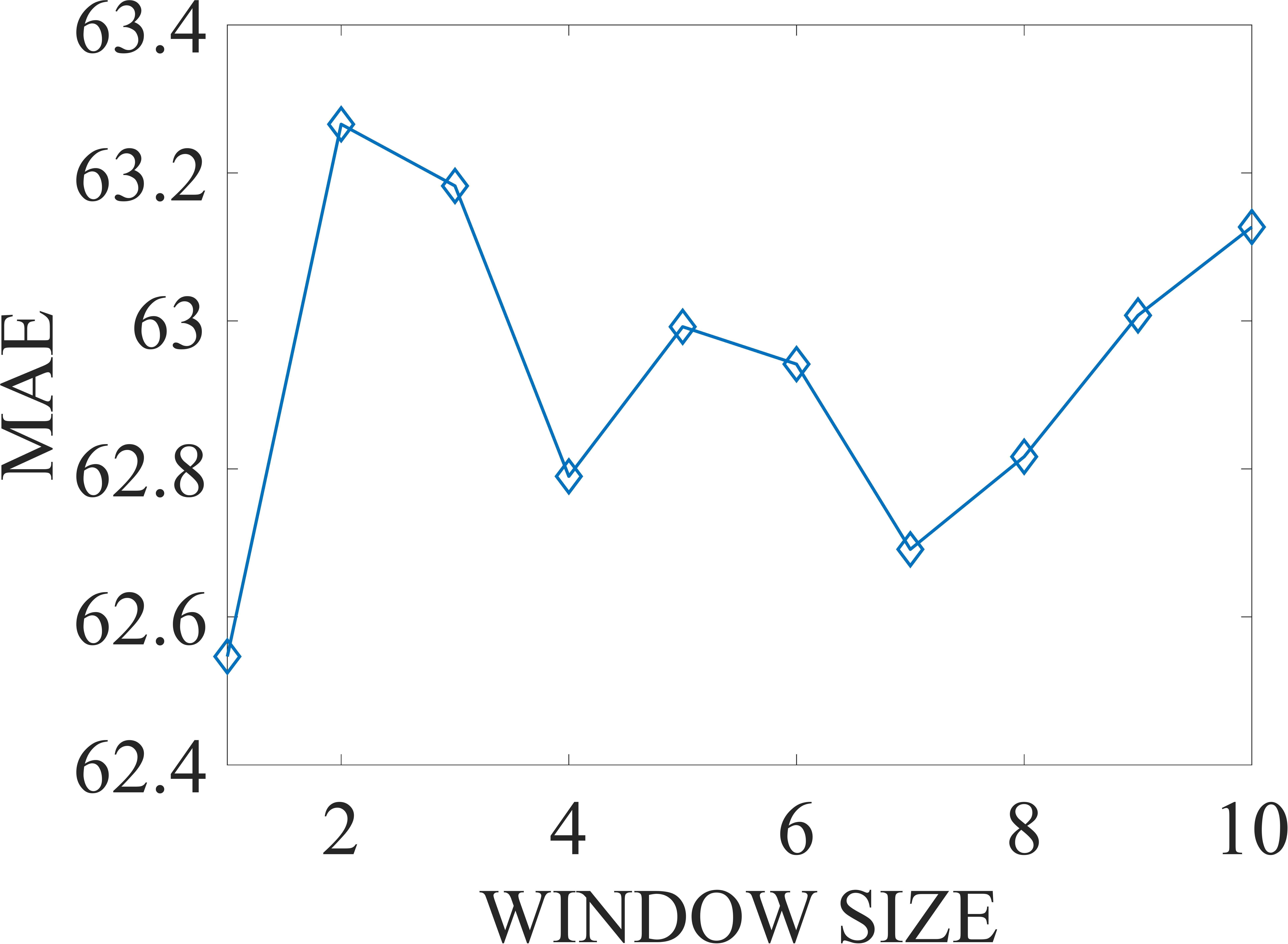}\\
		\end{minipage}%
	}%
	
	\subfigure[M4 Hourly Dataset]{
		\begin{minipage}[t]{0.23\linewidth}
			\centering
			\includegraphics[width=1.63in]{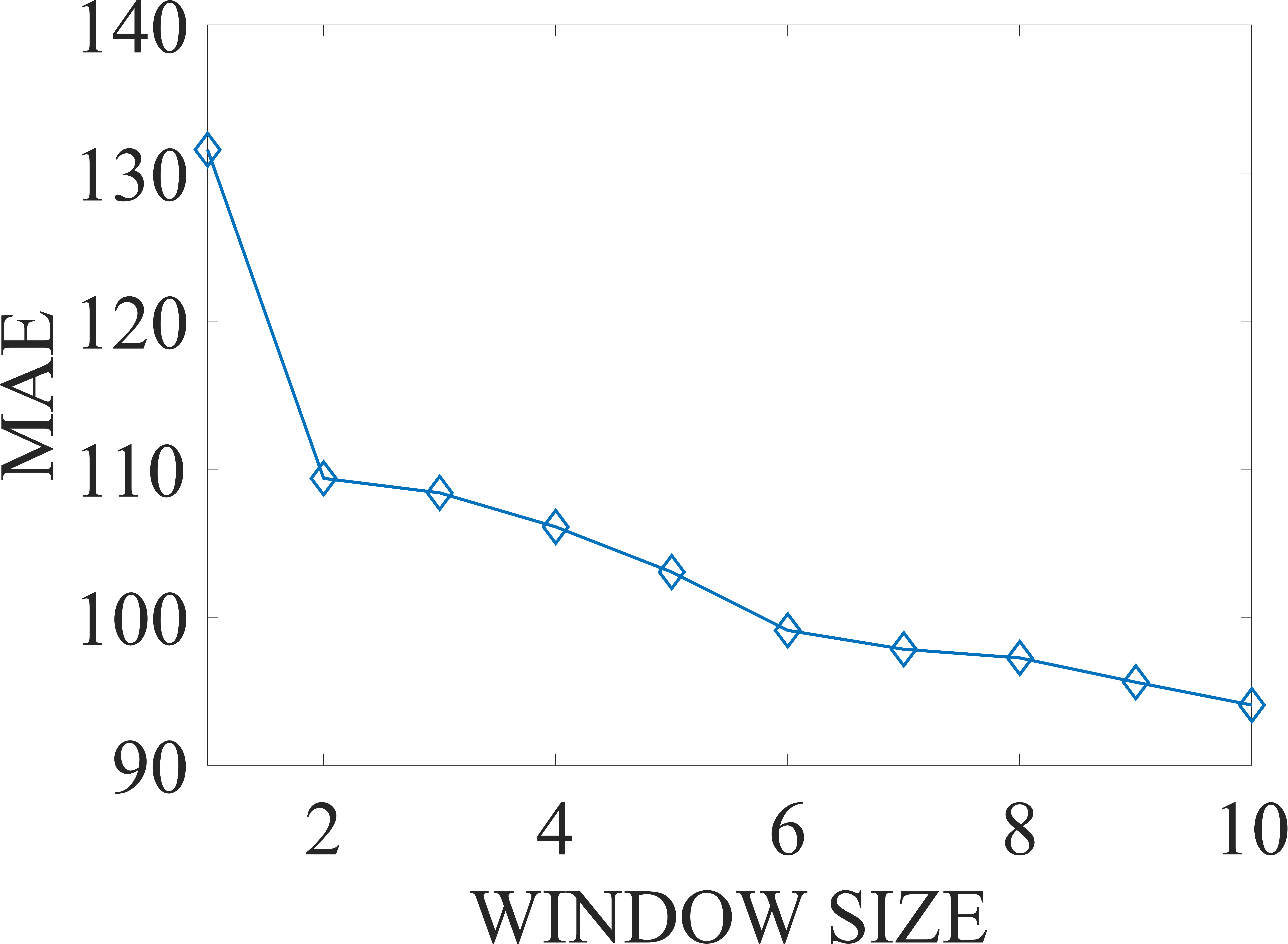}\\
		\end{minipage}%
	}%
%	\subfigure[M4 Monthly Dataset]{
%		\begin{minipage}[t]{0.23\linewidth}
%			\centering
%			\includegraphics[width=1.63in]{m4_monthly_dataset.tsf.pdf}\\
%		\end{minipage}%
%	}%
	\subfigure[M4 Quarterly Dataset]{
		\begin{minipage}[t]{0.23\linewidth}
			\centering
			\includegraphics[width=1.63in]{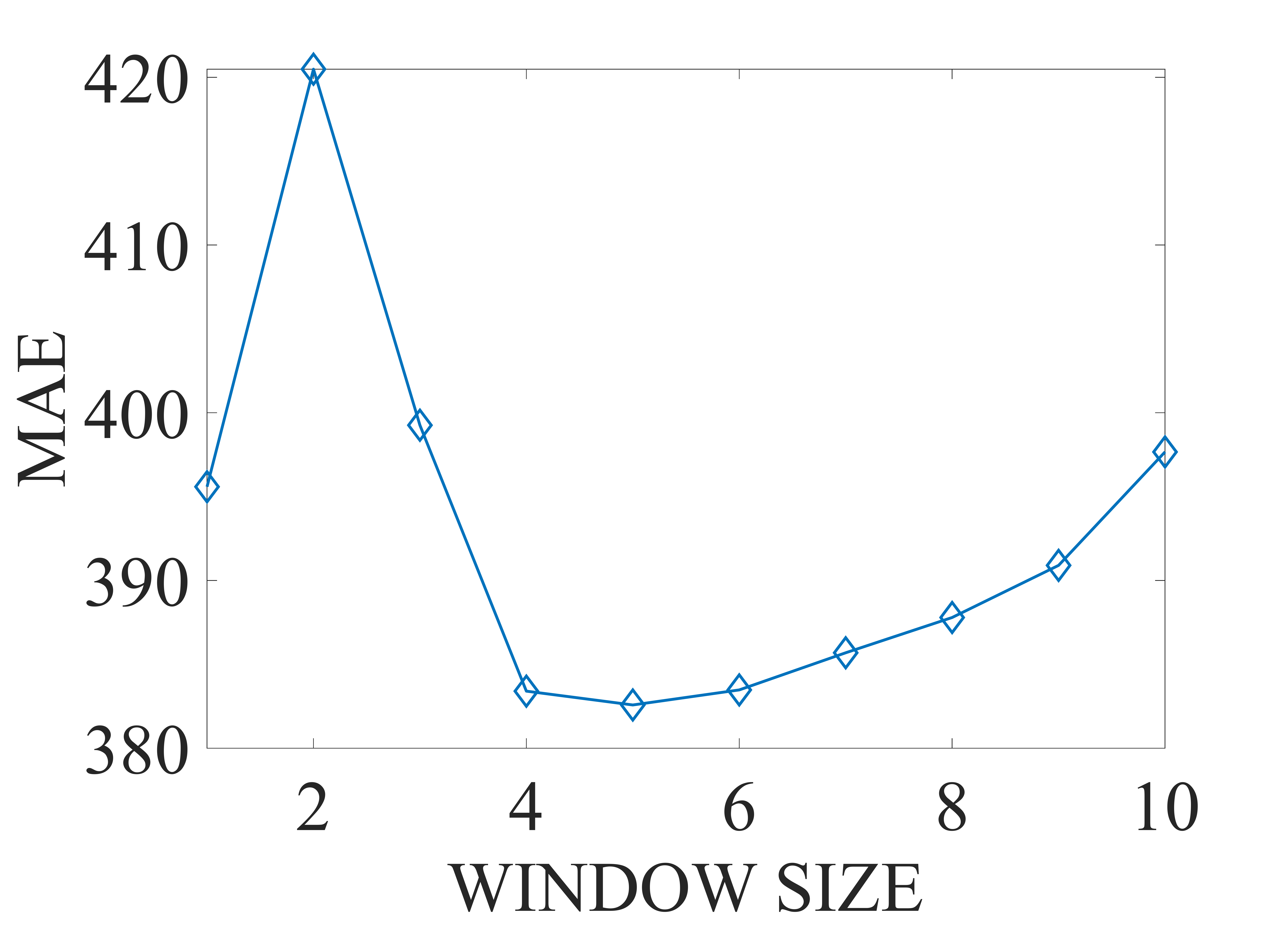}\\
		\end{minipage}%
	}%
	\subfigure[M4 Weekly Dataset]{
		\begin{minipage}[t]{0.23\linewidth}
			\centering
			\includegraphics[width=1.63in]{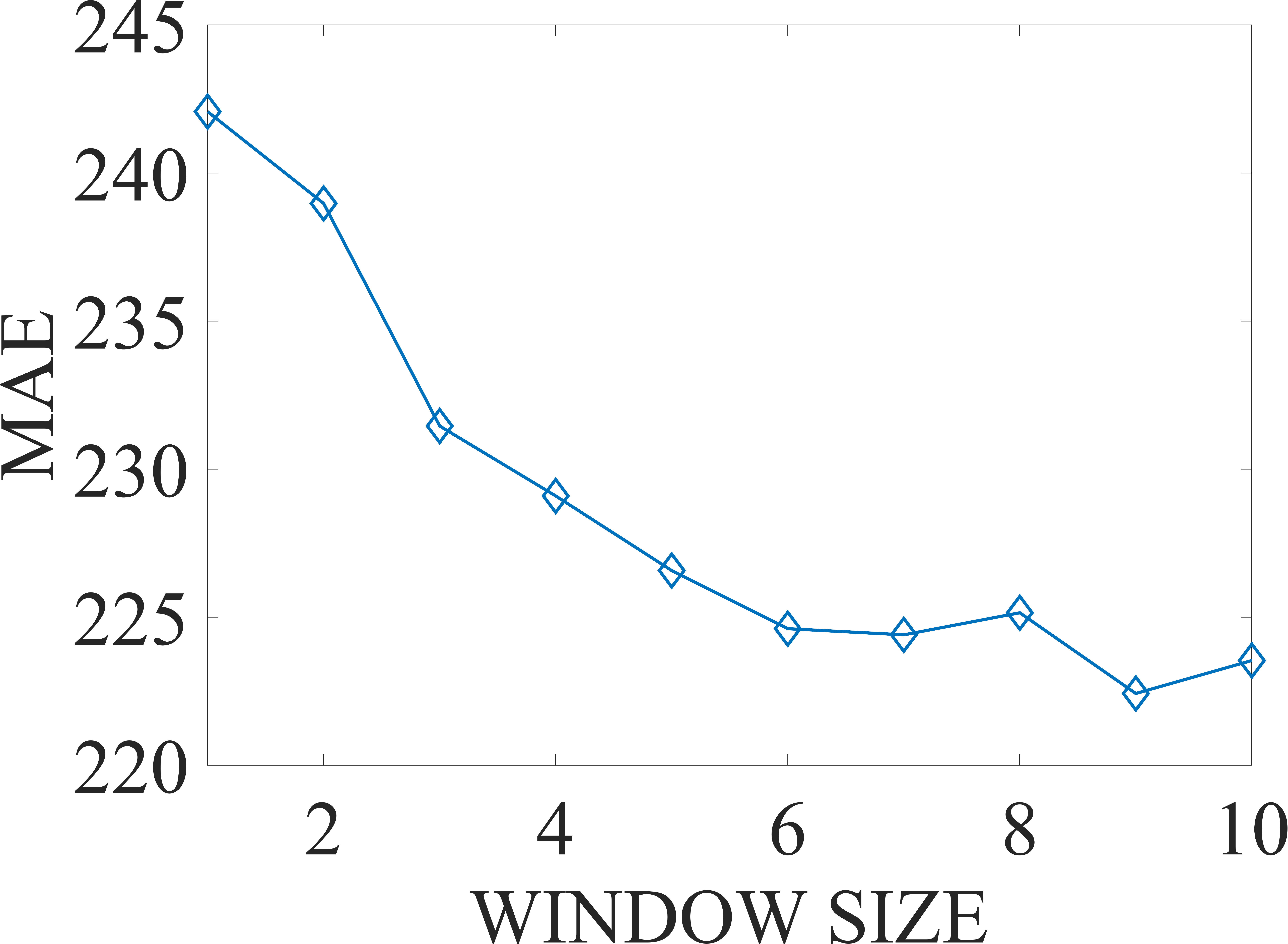}\\
		\end{minipage}%
	}%
	\subfigure[M4 Yearly Dataset]{
		\begin{minipage}[t]{0.23\linewidth}
			\centering
			\includegraphics[width=1.63in]{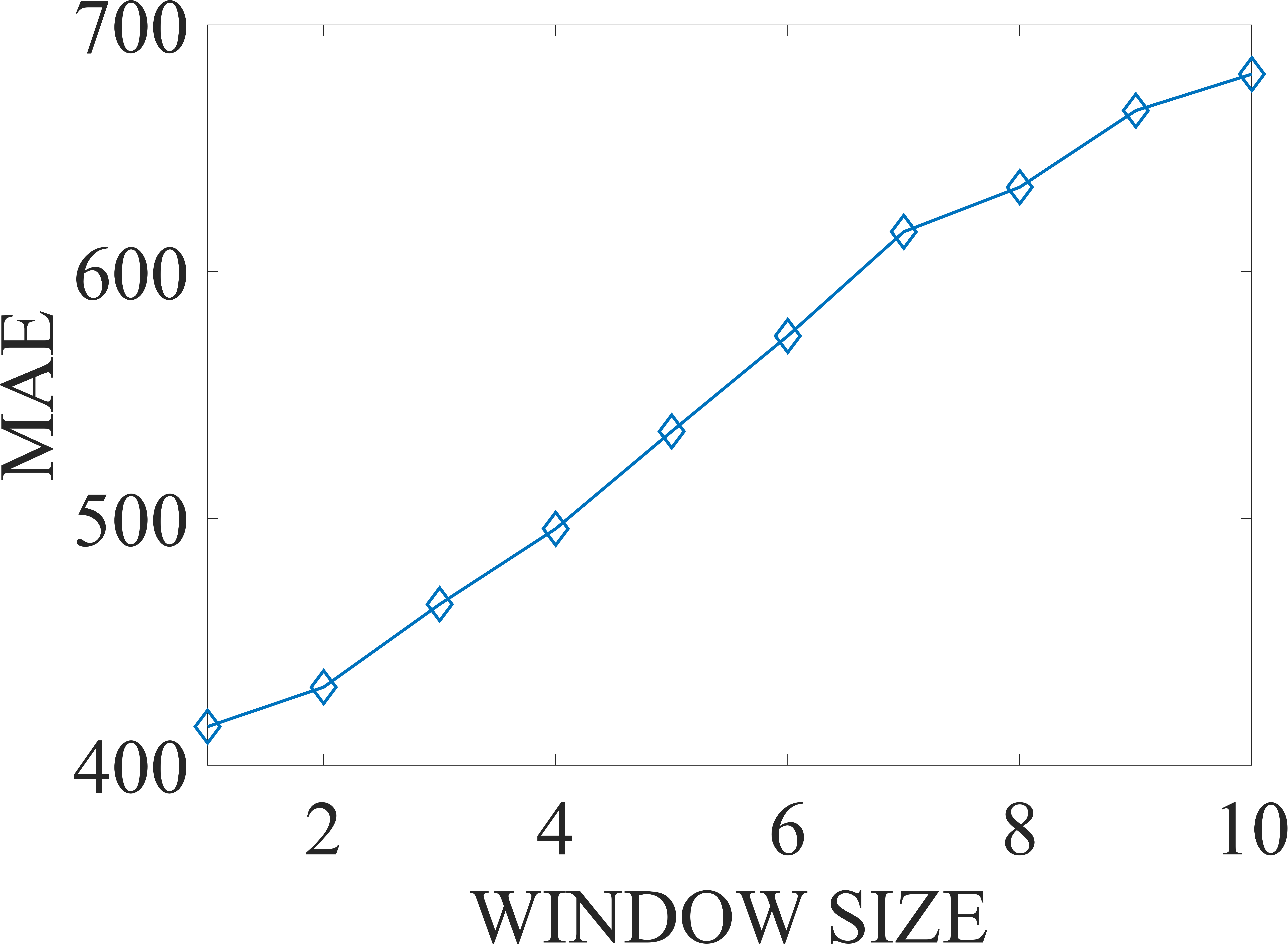}\\
		\end{minipage}%
	}%
%	\subfigure[CIF 2016 Dataset]{
%		\begin{minipage}[t]{0.23\linewidth}
%			\centering
%			\includegraphics[width=1.63in]{cif_2016_dataset.tsf.pdf}\\
%		\end{minipage}%
%	}%
%	\subfigure[Vehicle Trips Dataset]{
%		\begin{minipage}[t]{0.23\linewidth}
%			\centering
%			\includegraphics[width=1.63in]{vehicle_trips_dataset_without_missing_values.tsf.pdf}\\
%		\end{minipage}%
%	}%
%	\subfigure[KDD Cup 2018 Dataset]{
%		\begin{minipage}[t]{0.23\linewidth}
%			\centering
%			\includegraphics[width=1.63in]{kdd_cup_2018_dataset_without_missing_values.tsf.pdf}\\
%		\end{minipage}%
%	}%
	\caption{MAE Variations on Different Window Sizes on Univariate Datasets}
	\label{fig2}
\end{figure*}

\begin{figure*}
	\centering
	\subfigure[Car Parts Dataset]{
		\begin{minipage}[t]{0.23\linewidth}
			\centering
			\includegraphics[width=1.63in]{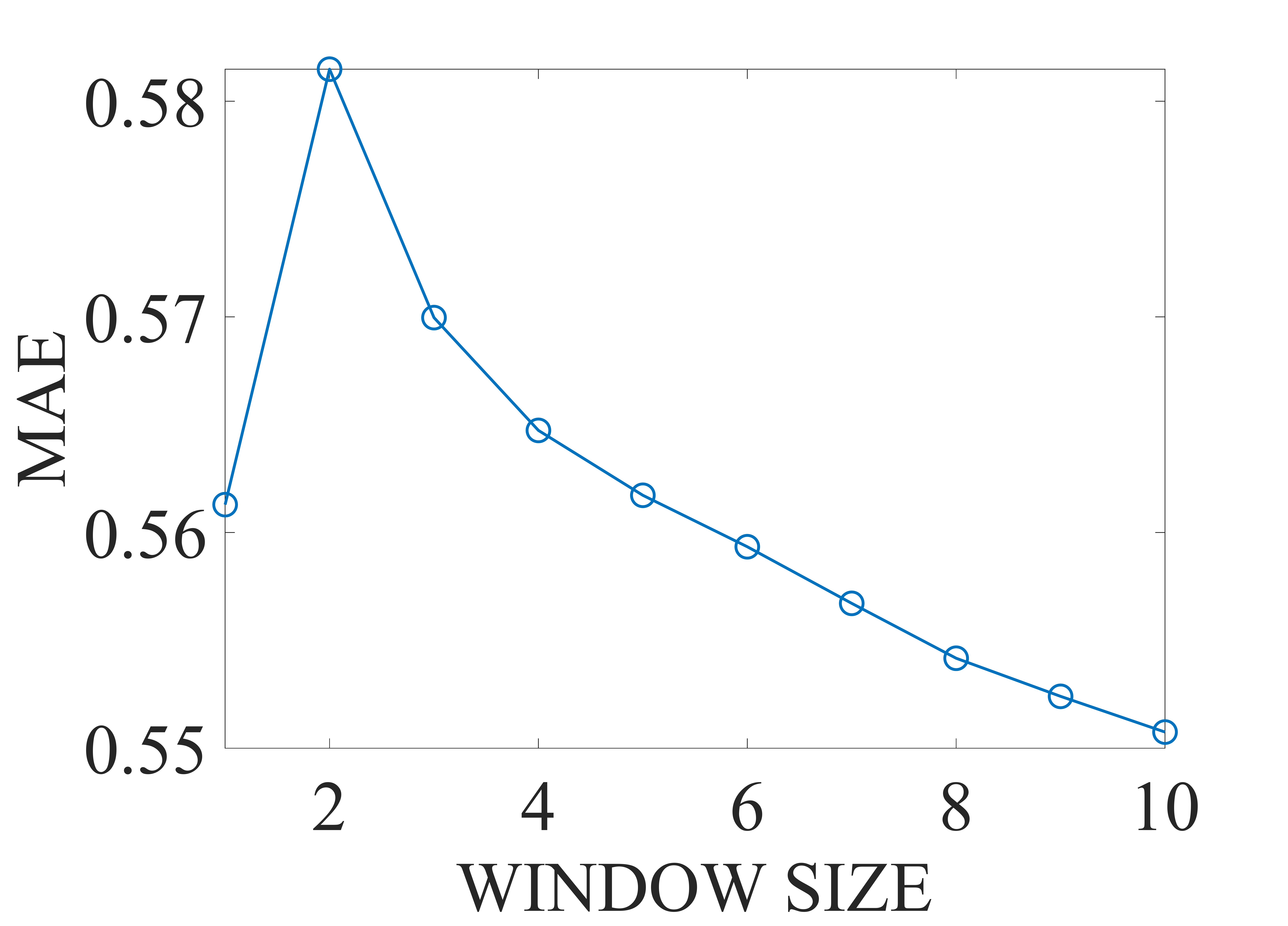}\\
		\end{minipage}%
	}%
	\subfigure[Covid Deaths Dataset]{
		\begin{minipage}[t]{0.23\linewidth}
			\centering
			\includegraphics[width=1.63in]{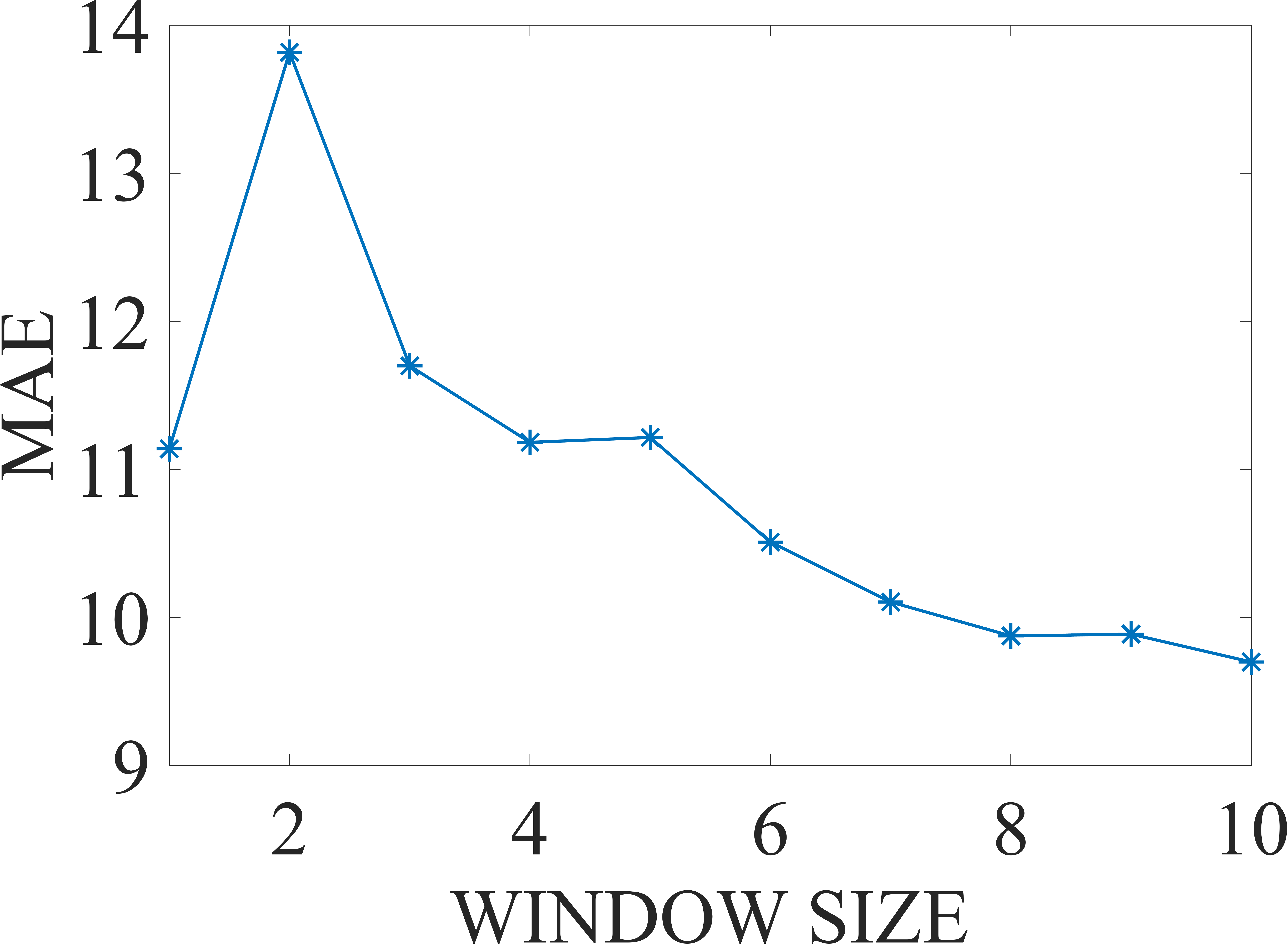}\\
		\end{minipage}%
	}%
	\subfigure[NN5 Daily Dataset]{
		\begin{minipage}[t]{0.23\linewidth}
			\centering
			\includegraphics[width=1.63in]{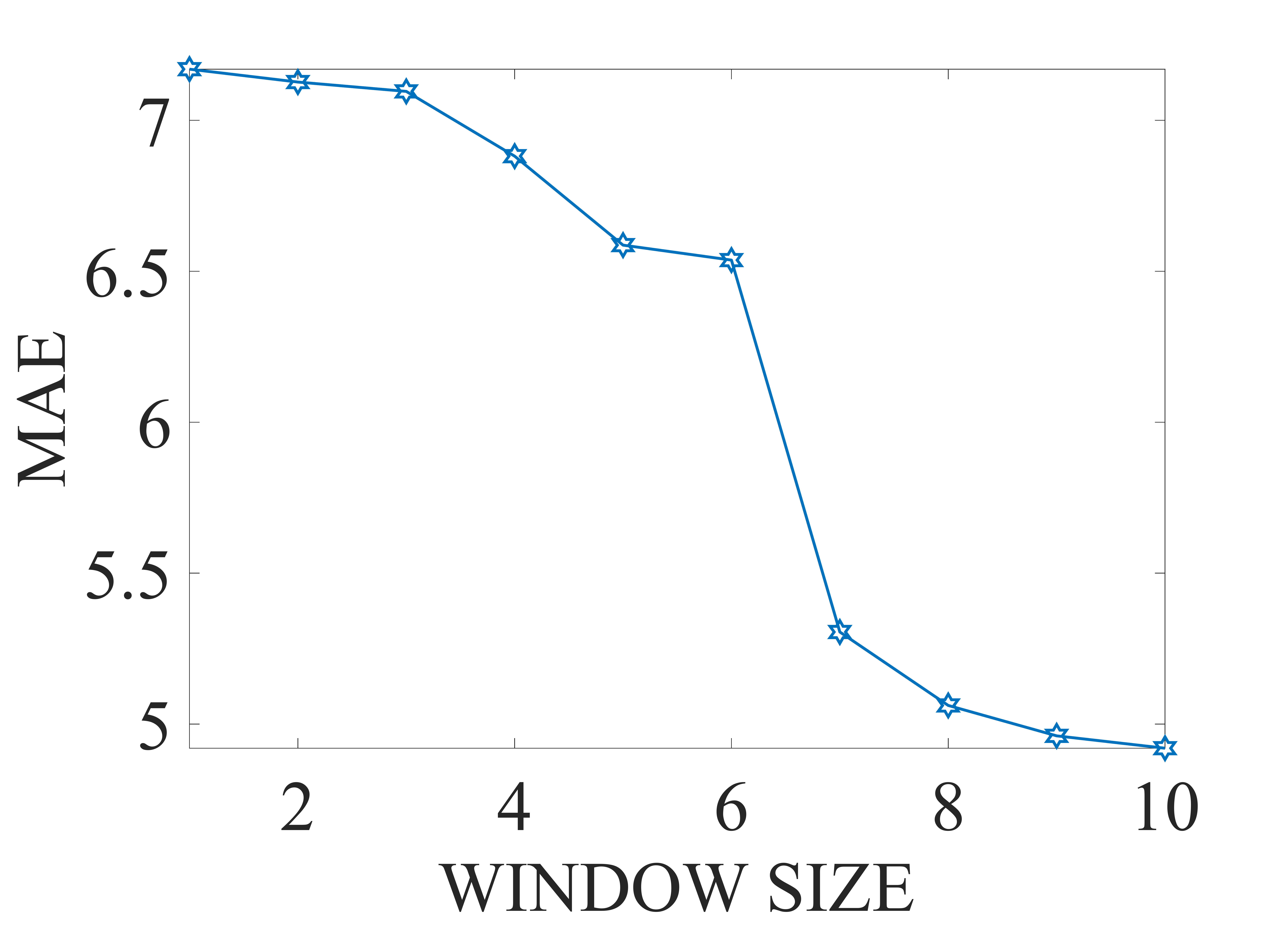}\\
		\end{minipage}%
	}%
	\subfigure[Rideshare Dataset]{
		\begin{minipage}[t]{0.23\linewidth}
			\centering
			\includegraphics[width=1.63in]{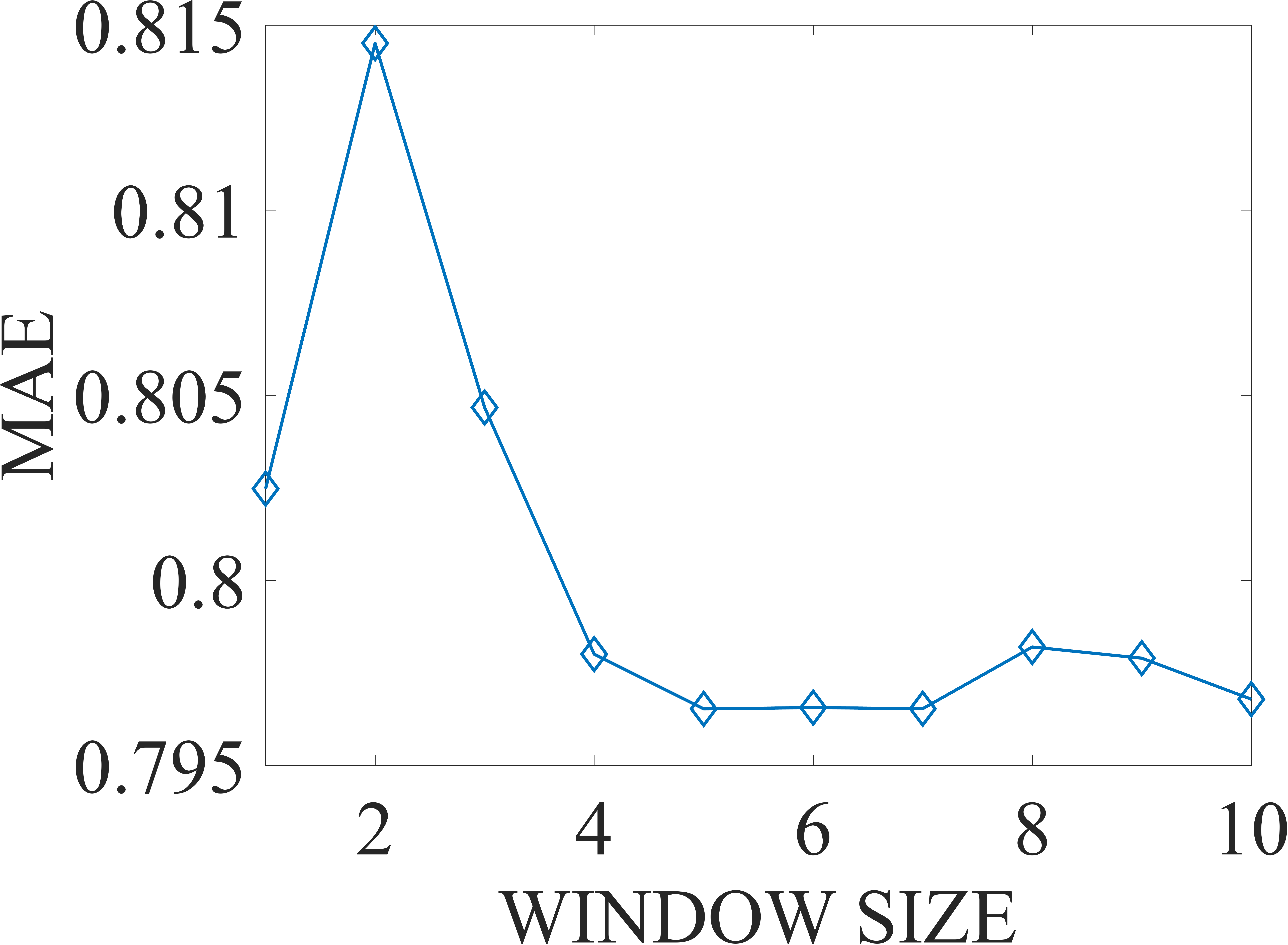}\\
		\end{minipage}%
	}%

	\caption{MAE Variations on Different Window Sizes on Multivariate Datasets}
	\label{fig4}
\end{figure*}

\subsection{Discussion on Experimental Results}
The experimental results of univariate and multivariate datasets are provided in Table \ref{table1}, \ref{table2} and Table \ref{table3}, \ref{table4} respectively. Generally, the proposed model obtains state-of-the-art results on most of the experimental time series datasets. However, the results of RMSE fail to remain consistent with MAE, it demonstrates that the proposed model's ability in handling abnormal prediction values is relatively lacking. We argue that the main reason for this phenomenon is that the proposed model pays much more attention to the global information distributed to the elements captured by the sliding window and ignores the influence of the original values on the future trend to a certain extent due to the strategies of data encoding and utilization of information processed of the proposed model. Especially, our proposed model outperforms transformer-based methods which attract lots of researchers' attention recently on almost all of the datasets, we consider that temporal data is not similar to images and videos in which there are enormous amount of semantic information needed to be extracted .

\subsection{Overall Performance Comparison Between Proposed and Comparative Models}
In order to comprehensively demonstrate superiority of the proposed model, we utilize Nemenyi test with $CD=q_{0.05}\sqrt{\frac{k(k+1)}{6N_d}}$ in which $k$ is the number of algorithms participating in the comparison and $N $is the number of datasets. Due to lack of some results of models with superscript $*$ on certain datasets, the Nemenyi test is divided into two groups to ensure fairness of comparison and the evaluation results are shown in Friedman test figure at Fig.\ref{fig6}. It can be easily concluded that the proposed model acquire the most excellent integrated performance on experimental datasets provided.

\subsection{Parameter Study}
Different datasets have their corresponding optimal parameter setting for the proposed model. We selected four univariate and four multivariate data sets for a brief analysis. In Fig.\ref{fig1} and \ref{figg}, it can be obtained that the performance of the proposed model benefits from a larger window size. And lifting factor $OUT$ has limited influence on the model capability and can reduce the error in some cases. Besides, synthesizing conditions of Fig.\ref{fig2} and \ref{fig4}, increasing the window size dose not necessarily improve model's performance, but larger window sizes can help capture more information and establish the foundation of precise predictions in general. Specifically, on multiple datasets such as M4 Monthly, Quarterly, KDD Cup 2018 and Covid Deaths datasets, error increases considerably when window size equals 2. The main reason probably is that the size sacrifices timeliness of data to some extent and is not capable of providing sufficient semantic information to the model so that the proposed model encounters difficulty in producing accurate predictions.

\begin{figure*}
	\centering
	\subfigure{
		\begin{minipage}[t]{0.25\linewidth}
			\centering
			\includegraphics[width=4.5cm,height=2.6cm]{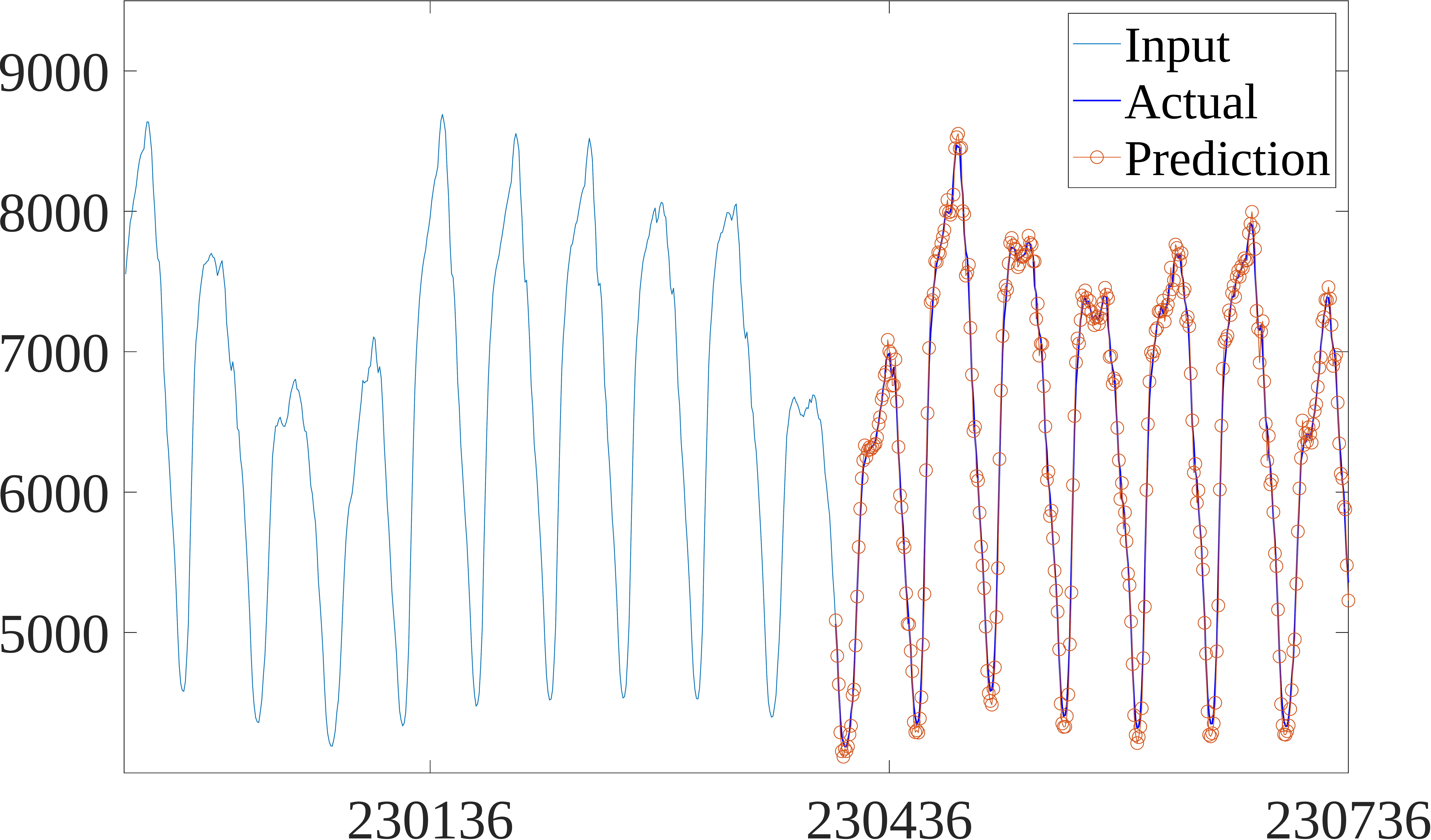}\\
		\end{minipage}%
		\begin{minipage}[t]{0.25\linewidth}
			\centering
			\includegraphics[width=4.5cm,height=2.6cm]{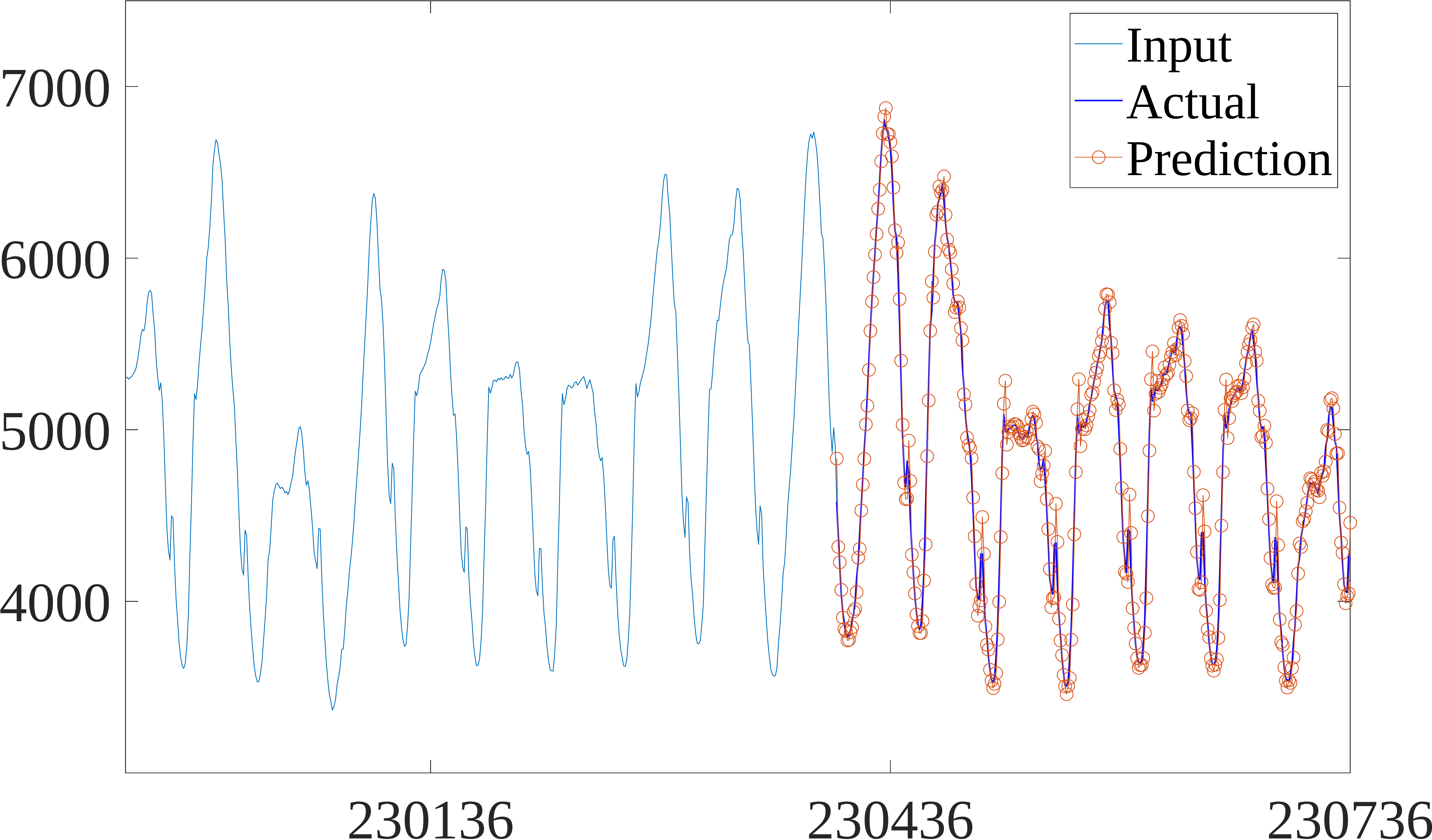}\\
		\end{minipage}%
	\begin{minipage}[t]{0.25\linewidth}
		\centering
		\includegraphics[width=4.5cm,height=2.6cm]{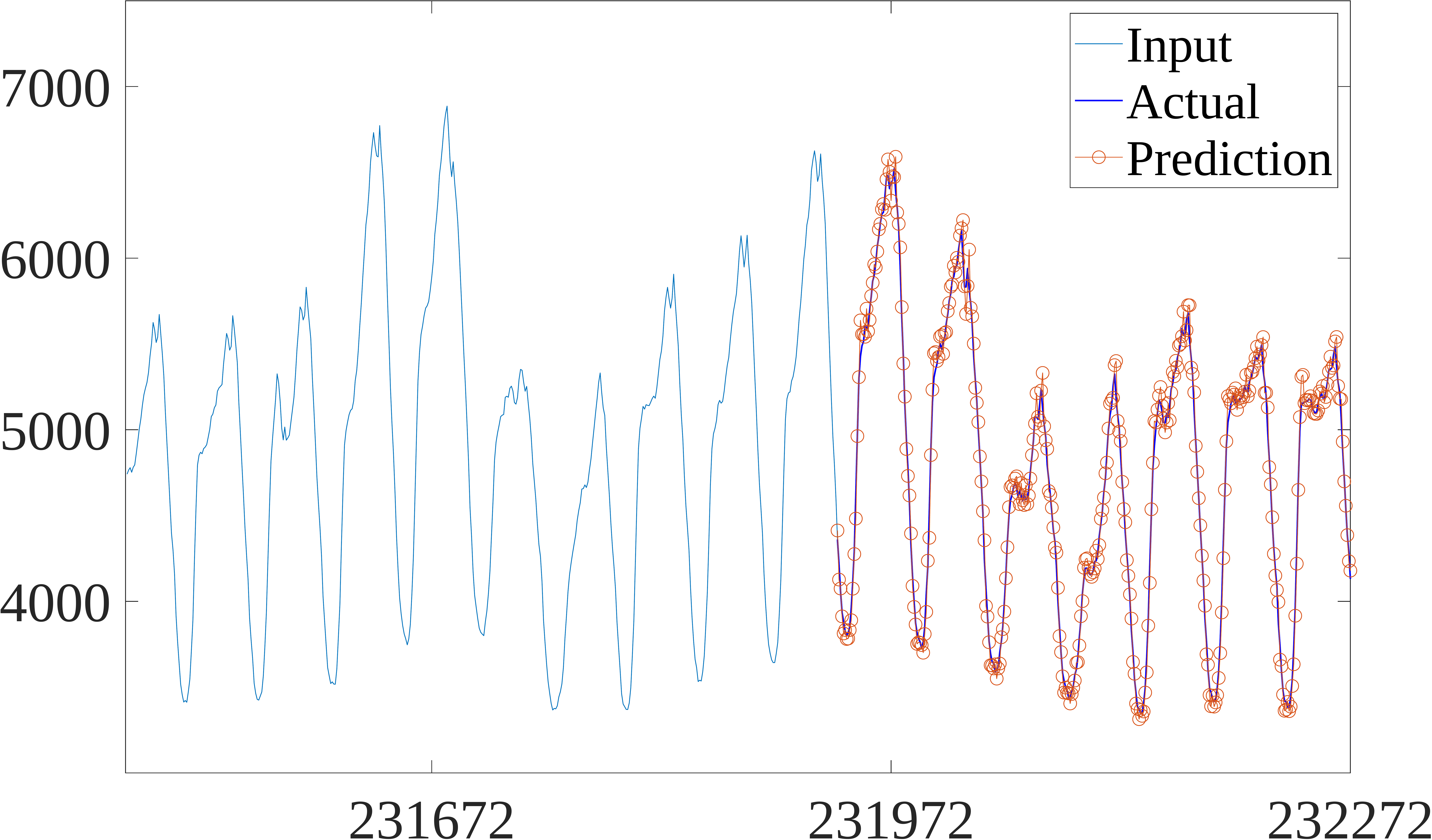}\\
	\end{minipage}%
	\begin{minipage}[t]{0.25\linewidth}
		\centering
		\includegraphics[width=4.5cm,height=2.6cm]{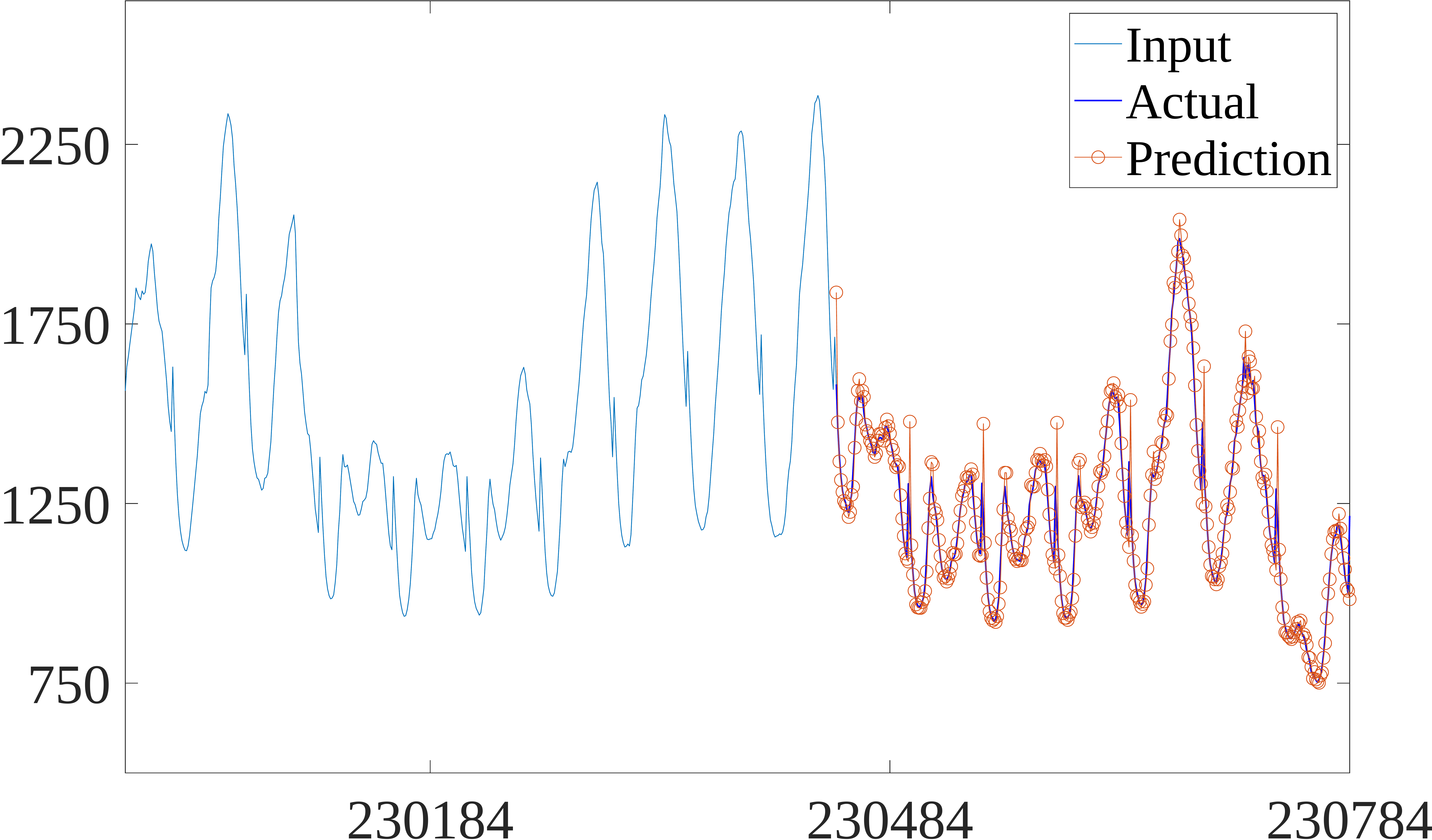}\\
	\end{minipage}%
	}%
	\caption{Qualitative Prediction Results by the Proposed Method on Aus. Electricity Demand Dataset}
	\label{fig7}
\end{figure*}
%\IEEEpubidadjcol

\section{Conclusion}
In this paper, a novel time series forecasting model is proposed which consists of encoder part and semi-asymmetric convolutional architecture. The main role of devised data encoder is assigning elements in original observation time series with positional information so that the model could possess a global view on observation sequences. Based on processed data, a novel architecture is designed referencing asymmetric convolution and considering variability of time series which enables the model to obtain information at flexible scales on different time series. Capturing features with separate range helps model to learn underlying relationship among elements with effective understanding of associated positional information. Both of them contributes to the outstanding performance of the proposed model.

To comprehensively demonstrate the performance of the model proposed in this paper, we conduct experiments on 27 univariate and 16 multivariate datasets. The experimental results illustrate that the proposed model outperforms comparative methods on most of forecasting tasks. Specifically, the proposed model achieves the highest rank on all competition datasets such as M series, KDD Cup and Web Traffic. In addition to these intuitive results, Nemenyi test also strongly demonstrates the excellent performance of the proposed model. Besides, we also investigate the influences of the two main parameters on 24 datasets to further explain settings of the proposed model. 

Nevertheless, the proposed model achieves relatively satisfying performance in most of forecasting experiments, there are still some potentials in it which can be further explored. We think we may be able to improve the model in two possible directions: 1): The attention mechanism can be introduced into the model to help the model better understand semantic information in time series, 2): A recurrent architecture of convolutional neural network is expected to be developed to better memory past information.

% or


%\appendices

%\section{}

%\ifCLASSOPTIONcompsoc
%\else
\section*{Acknowledgment}
%This work is supported by National Science and Technology Major Project (Granted No. 2020AAA0109401) and Natural Science Foundation of China (granted No. 62192731). And the authors greatly appreciate the anonymous reviewers’ suggestions and the editor’s encouragement.

\scriptsize
\bibliography{cite}
\bibliographystyle{IEEEtran}

\end{document}